%% file: main.tex
\documentclass{article}
\usepackage[main, final]{neurips_2026}

\usepackage[utf8]{inputenc} 
\usepackage[T1]{fontenc}    
\usepackage{hyperref}       
\usepackage{url}            
\usepackage{booktabs}       
\usepackage{amsfonts}       
\usepackage{nicefrac}       
\usepackage{microtype}      
\usepackage{xcolor}         
\usepackage{graphicx} 
\usepackage{tikz} 
\usepackage{hyperref}
\usepackage{color}

\usepackage{amsmath}

\usepackage{theorem}
\usepackage{fancybox,amssymb,amstext,amsmath,stmaryrd,wasysym,proof,mathtools,multirow,stackrel}
\SetSymbolFont{stmry}{bold}{U}{stmry}{m}{n}
\SetSymbolFont{wasy}{bold}{U}{wasy}{m}{n}
\usepackage{algorithm}
\usepackage{algpseudocode}
\makeatletter
\algrenewcommand\ALG@beginalgorithmic{\footnotesize}
\makeatother

\usepackage{booktabs}
\usepackage{array}
\newcolumntype{H}{>{\setbox0=\hbox\bgroup}c<{\egroup}@{}}
\usepackage{enumitem}
\usepackage{xcolor}         
\usepackage[table]{xcolor} 
\usepackage{subcaption} 

\usepackage{tcolorbox}
\tcbuselibrary{breakable,skins}

\newtcolorbox[auto counter,number within=section]{llmprompt}[2][]{%
  enhanced,
  breakable,
  colback=black!2,
  colframe=black!45,
  colbacktitle=black!8,
  coltitle=black,
  boxrule=0.4pt,
  arc=0pt,
  left=6pt,right=6pt,top=6pt,bottom=6pt,
  before skip=10pt,after skip=10pt,
  fonttitle=\normalfont\normalsize\bfseries,
  fontupper=\normalfont\normalsize\ttfamily,
  before upper={\raggedright\setlength{\parindent}{0pt}\setlength{\parskip}{2pt}},
  lines before break=4,
  title={Prompt~\thetcbcounter: #2},
  title after break={Prompt~\thetcbcounter\ (continued)},
  #1
}

\input{figures/retrieval_qualitative_style}

\input{macros.tex}

\title{LogSTOP: Temporal Scores over Prediction Sequences for Matching and Retrieval}

\author{
Avishree Khare\textsuperscript{1},
Hideki Okamoto\textsuperscript{2},
Bardh Hoxha\textsuperscript{2},
Georgios Fainekos\textsuperscript{2},
Rajeev Alur\textsuperscript{1} \\
\textsuperscript{1}University of Pennsylvania \\
\textsuperscript{2}Toyota Motor North America (R\&D)
}

\begin{document}

\maketitle

\begin{abstract}
Neural models such as YOLO and HuBERT detect local properties such as objects ("car") and emotions ("angry") in video frames and audio clips. Lifting frame-level detection scores to sequence-level scores for temporal properties enables applications such as query matching (e.g., ``does the speaker \textit{eventually} sound happy in this audio clip?") and ranked retrieval (e.g., "retrieve top 5 videos with a 10 second scene where a car appears \textit{until} a pedestrian appears").
We formalize this problem of assigning Scores for TempOral Properties (STOPs) over sequences, given noisy score predictors for local properties.
We then propose LogSTOP, a scoring function that efficiently computes scores for temporal properties expressed in Linear Temporal Logic.
Empirically, LogSTOP outperforms Large Multimodal Models by at least 16\% on query matching with temporal properties over objects-in-videos and emotions-in-speech, while matching or improving upon other Temporal Logic-based baselines. On ranked retrieval with temporal properties, LogSTOP improves mean average precision over zero-shot text-to-video retrieval baselines by up to 28\%.
Code is available at \texttt{https://github.com/avishreekh/logstop}.
\end{abstract}

\input{src/introduction}
\input{src/setup}
\input{src/method}
\input{src/applications}
\input{src/experiments}
\input{src/discussion}
\input{src/related}
\input{src/conclusion}

\section{Acknowledgements and Disclosure of Funding}

We would like to thank the NeurIPS reviewers and Area Chair for their feedback. This research was partially supported by NSF award SLES 2331783.

\bibliographystyle{plain}
\bibliography{sources}

\appendix
\input{appendix/main}



\end{document}

%% file: figures/retrieval_qualitative_style.tex
\definecolor{rqTeal}{HTML}{2D7D46}
\definecolor{rqBlue}{HTML}{235ABC}
\definecolor{rqAbsent}{HTML}{EBEEF1}
\definecolor{rqInk}{HTML}{24363B}
\definecolor{rqLine}{HTML}{D3DEDF}
\definecolor{rqLight}{HTML}{EFF5F5}
\definecolor{rqGreenLight}{HTML}{EFF7F0}
\definecolor{rqGreenStrip}{HTML}{D6E8D9}

%% file: macros.tex
\theorembodyfont{\itshape}
\newtheorem{mytheorem}{Theorem}[section]
\newtheorem{mylemma}[mytheorem]{Lemma}
\newtheorem{myproposition}[mytheorem]{Proposition}

\newtheorem{mycorollary}[mytheorem]{Corollary}
\theorembodyfont{\rmfamily}

\newtheorem{myremark}[mytheorem]{Remark}
\newtheorem{myexample}[mytheorem]{Example}
\newtheorem{mydefinition}[mytheorem]{Definition}

\newenvironment{proposition}{\begin{myproposition}}{\end{myproposition}}

\newenvironment{proof}{\par\noindent\textit{Proof. }\ignorespaces}{\hfill$\square$\par}
\providecommand{\qed}{\hfill$\square$}

\newcommand{\ttrue}{\mathrm{t{\kern-1.5pt}t}}
\newcommand{\ffalse}{\mathrm{f{\kern-1.5pt}f}}

\newcommand{\UntilOpSTL}[1]{\mathbin{\mathcal{U}_{#1}}}
\newcommand{\UntilOp}[1]{\mathbin{\mathcal{U}}}
\newcommand{\Release}[1]{\mathbin{\mathcal{R}}}

\newcommand{\DiaOpSTL}[1]{\Diamond_{#1}}
\newcommand{\BoxOpSTL}[1]{\square_{#1}}
\newcommand{\DiaOp}[1]{\Diamond}
\newcommand{\BoxOp}[1]{\square}

\newcommand{\parhead}[1]{\smallskip \noindent {\bfseries\boldmath\ignorespaces {#1}}\hskip 0.45em plus 0.15em minus 0.15em }

%% file: src/introduction.tex
\section{Introduction}

Detecting complex temporal events in unstructured data sequences, such as videos and audio clips, is important in several domains.
For instance, monitoring systems 
need to
\textit{match} scenes perceived by autonomous vehicles 
against critical \textit{temporal} properties, such as 
"the car \textit{always} remains in a given lane"~\citep{strem}. 
Beyond matching, video retrieval systems must rank videos by relevance to temporal queries, such as "a 10 second scene where a person \textit{eventually} stands up". 
Neural detection models such as YOLO~\citep{yolo} and SlowR50~\citep{feichtenhofer2019slowfast}
can provide detection scores for local properties such as objects (\(\mathsf{car}\)) and actions (\(\mathsf{standing}\)) in individual video frames and short video clips, respectively.
How can these scores be lifted to the temporal properties discussed above? Concretely,

\begin{center}
\textit{Given a temporal property and (potentially noisy) predictors for local properties, how can we score how well a sequence satisfies the temporal property?}    
\end{center}

The sequences and local properties can span multiple modalities and classes of interest, including objects and actions in videos, and speakers in audio clips. 
Scores for TempOral Properties (STOPs) can support the two downstream applications discussed above:  \textbf{query matching} by comparing against a suitable threshold, and \textbf{ranked retrieval} of the top-$k$ sequences sorted by scene-level scores. 

We adopt Linear Temporal Logic, with temporal operators such as "Always" (\(\BoxOp{I}\)) and "Until" (\(\UntilOp{I}\)), as a suitable language
for expressing diverse
temporal properties of interest.
For example, 
the property ``the car always remains in lane" can be written as \(\BoxOp{I}\,\mathsf{car\_in\_lane}\) 
where the local property \(\mathsf{car\_in\_lane}\) can be detected using a neural model. 
Similarly, 
the property ``speakers A and B sound happy until A sounds sad" can be expressed
as \((\mathsf{happy}_A \land \mathsf{happy}_B)
\UntilOp{I}\,\mathsf{sad}_A\).

\input{graphics/charts/example}

Existing methods for query matching include Large Multimodal Models prompted directly with temporal queries in natural language. Neuro-symbolic approaches to video query matching~\cite{yang2023specification,Choi2024TowardsNV} show that Vision Language Models struggle with logical queries and use Probabilistic Model Checking~\cite{storm} over queries in LTL or Probabilistic Computation Tree Logic (PCTL). 
These approaches, however, do not explicitly account for transient prediction errors, such as detector failures due to occlusion, and model checking requires exponential time in the number of local properties. Quantitative semantics for LTL assign real-valued degrees of satisfaction~\citep{FAINEKOS20094262,donze2010robust,Akazaki2015TimeRI,gmrob} and can therefore provide an efficient solution to the STOP problem. However, standard quantitative STL semantics, such as STL Spatial Robustness~\cite{FAINEKOS20094262}, aggregate with \(\min\)/\(\max\) operators and are designed for falsification; their usefulness for ranking noisy prediction sequences by temporal query relevance has not been studied.

We propose a scoring function called LogSTOP, inspired by quantitative semantics for LTL, for assigning scores to temporal properties over noisy local prediction sequences. First, LogSTOP includes a lightweight class-specific windowed smoothing step to reduce the effect of transient errors in local property predictions. Second, LogSTOP recursively composes the resulting local property scores according to the logical and temporal structure of the query, allowing a length-\(T\) sequence and temporal property \(\varphi\) to be scored efficiently in \(\mathcal{O}(T\cdot|\varphi|)\) time and space. Third, this temporal aggregation is monotonic in the post-processed local property scores: improving the score for a local property in a direction that supports the temporal property cannot decrease the LogSTOP. This monotonicity makes the scores useful for both ranked retrieval and threshold-based query matching. We further propose a length- and query-adaptive threshold for query matching. 
Figure~\ref{fig:method_overview} illustrates the use of LogSTOP for retrieving scenes relevant to a temporal query, with detections from Grounding DINO~\citep{liu2024grounding}. Windowed smoothing improves resilience to transient prediction errors due to occlusion and the adaptive threshold is better for matching than the constant \(\log 0.5\) threshold.

We evaluate LogSTOP on query matching and ranked retrieval with 15 temporal property templates spanning videos and speech, and local properties including objects, actions, and emotions. Since no existing benchmarks cover this breadth of temporal properties and modalities, we introduce the QMTP (Query Matching for Temporal Properties) and TP2VR (Temporal Property to Video Retrieval) benchmarks over existing source datasets. On query matching, LogSTOP with YOLO and HuBERT~\citep{hsu2021hubert,superb} outperforms Large Vision / Audio Language Models (LVLMs / LALMs) by at least \(16\%\) in balanced accuracy, while matching or improving upon Temporal-Logic baselines. On ranked retrieval, LogSTOP with OWLv2~\citep{minderer2023scaling} and SlowR50~\citep{feichtenhofer2019slowfast} outperforms zero-shot text-to-video retrieval methods, such as mPLUG~\citep{li2022mplug} and text-text similarity with video captions, by \(28\%\) and \(18\%\) in mean average precision on TP2VR-objects and TP2VR-actions respectively. 

%% file: graphics/charts/example.tex
\input{figures/bus_stop_example}

%% file: figures/bus_stop_example.tex
\begin{figure}[H]
\centering
\begingroup
\definecolor{busTeal}{HTML}{2D7D46}
\definecolor{busRed}{HTML}{A13939}
\definecolor{busAmber}{HTML}{96720B}
\definecolor{busBlue}{HTML}{235ABC}
\definecolor{busInk}{HTML}{24363B}
\definecolor{busLine}{HTML}{D3DEDF}
\definecolor{busLight}{HTML}{EFF5F5}
\definecolor{busGreenLight}{HTML}{EFF7F0}
\definecolor{busGreenStrip}{HTML}{D6E8D9}
\definecolor{busYellowLight}{HTML}{FFFAE9}
\definecolor{busYellowStrip}{HTML}{F6E8AE}
\definecolor{busRedLight}{HTML}{FCF0F0}
\definecolor{busRedStrip}{HTML}{EED7D7}
\definecolor{busParamLight}{HTML}{F1EDFA}
\definecolor{busParamLine}{HTML}{C5B7DE}
\input{figures/bus_stop_until/figure_parameters.tex}
\def\busFrameWidth{2.50}
\def\busFrameHeight{2.00}
\newcommand{\BusStopFont}{\rmfamily\fontsize{9}{10.5}\selectfont}
\newcommand{\BusStopNumber}[1]{\pgfmathprintnumber[fixed,fixed zerofill,precision=2]{#1}}
\newcommand{\BusStopDetection}[7]{%
  \draw[#7,line width=0.7pt,rounded corners=1.5pt]
    ({\frameX+\busFrameWidth*#1},{\rowTop-0.56-\busFrameHeight*#2}) rectangle
    ({\frameX+\busFrameWidth*#3},{\rowTop-0.56-\busFrameHeight*#4});
  \pgfmathsetmacro{\labelX}{min(\busFrameWidth*#1,0.88)}%
  \node[anchor=south west,fill=#7,text=white,inner sep=1.5pt,rounded corners=1.5pt,
        font=\BusStopFont] at
    ({\frameX+\labelX},{\rowTop-0.56-\busFrameHeight*#2}) {#5\ #6};
}
\newcommand{\BusStopRow}[7]{%
  \def\rowTop{#1}%
  \ifcase#2\relax\or
    \def\busRowColor{busTeal}\def\busRowFill{busGreenLight}\def\busStripFill{busGreenStrip}%
  \or
    \def\busRowColor{busAmber}\def\busRowFill{busYellowLight}\def\busStripFill{busYellowStrip}%
  \or
    \def\busRowColor{busRed}\def\busRowFill{busRedLight}\def\busStripFill{busRedStrip}%
  \fi
  \ifdim #5pt>\BusStopThreshold pt
    \def\busRawColor{busTeal}\def\busRawVerdict{Match}\def\busRawSign{>}\def\busRawMark{\checkmark}%
  \else
    \def\busRawColor{busRed}\def\busRawVerdict{Non-match}\def\busRawSign{<}\def\busRawMark{\times}%
  \fi
  \ifdim #6pt>\BusStopThreshold pt
    \def\busSmoothColor{busTeal}\def\busSmoothVerdict{Match}\def\busSmoothSign{>}\def\busSmoothMark{\checkmark}%
  \else
    \def\busSmoothColor{busRed}\def\busSmoothVerdict{Non-match}\def\busSmoothSign{<}\def\busSmoothMark{\times}%
  \fi
  \fill[\busRowFill,rounded corners=6pt] (0,{#1-3.20}) rectangle (16.5,{#1+0.17});
  \node[anchor=west,font=\BusStopFont\bfseries,text=\busRowColor]
    at (0.94,#1) {Video #2: #3};
  \node[font=\BusStopFont,text=busInk] at (0.40,{#1-1.05}) {Rank};
  \node[circle,fill=\busRowColor,text=white,minimum size=15pt,
        font=\BusStopFont\bfseries] at (0.40,{#1-1.48}) {#2};
  \filldraw[fill=\busStripFill,draw=\busRowColor!45,line width=0.45pt,rounded corners=4pt]
    (0.76,{#1-2.87}) rectangle (12.23,{#1-0.29});
  \foreach \hole in {0,...,28}{%
    \fill[white,rounded corners=0.8pt]
      ({1.19+\hole*0.38},{#1-0.47}) rectangle +(0.18,0.08);
    \fill[white,rounded corners=0.8pt]
      ({1.19+\hole*0.38},{#1-2.78}) rectangle +(0.18,0.08);
  }
  \node[text=\busRowColor,font=\BusStopFont] at (0.97,{#1-1.58}) {$\cdots$};
  \node[text=\busRowColor,font=\BusStopFont] at (11.98,{#1-1.58}) {$\cdots$};
  \foreach[count=\j from 0] \sample/\imagepath/\state/\boxpath in {#7}{%
    \pgfmathsetmacro{\frameX}{1.19+\j*2.67}%
    \begin{scope}
      \clip[rounded corners=3pt] (\frameX,{#1-0.56})
        rectangle +(\busFrameWidth,-\busFrameHeight);
      \node[anchor=north west] at (\frameX,{#1-0.56})
        {\includegraphics[width=\busFrameWidth cm]{\imagepath}};
    \end{scope}
    \ifnum#2=2\relax\ifnum\j=1\relax
      \draw[busAmber,line width=0.8pt,rounded corners=3pt] (\frameX,{#1-0.56}) rectangle +(\busFrameWidth,-\busFrameHeight);
    \else
      \draw[busLine,line width=0.45pt,rounded corners=3pt] (\frameX,{#1-0.56}) rectangle +(\busFrameWidth,-\busFrameHeight);
    \fi\else
      \draw[busLine,line width=0.45pt,rounded corners=3pt] (\frameX,{#1-0.56}) rectangle +(\busFrameWidth,-\busFrameHeight);
    \fi
    \input{\boxpath}%
  }%
  \draw[busLine] (12.40,{#1-0.19})--(12.40,{#1-2.99});
  \ifnum#2=2\relax\def\busScoreOffset{0.70}\else\def\busScoreOffset{1.30}\fi
  \node[font=\BusStopFont,text=\busSmoothColor]
    at (14.30,{#1-\busScoreOffset}) {$\BusStopNumber{#6}$};
  \node[draw=\busSmoothColor,fill=\busSmoothColor!18,rounded corners=3pt,
        minimum size=0.44cm,line width=0.55pt,font=\BusStopFont,text=\busSmoothColor]
    at (15.45,{#1-\busScoreOffset}) {$\busSmoothMark$};
  \ifnum#2=2\relax
    \node[font=\BusStopFont,text=busInk,align=center]
      at (14.45,{#1-1.35}) {No smoothing ($w_{\mathsf{person}}=0$)};
    \node[font=\BusStopFont,text=\busRawColor]
      at (14.30,{#1-1.84}) {$\BusStopNumber{#5}$};
    \node[draw=\busRawColor,fill=\busRawColor!18,rounded corners=3pt,
          minimum size=0.44cm,line width=0.55pt,font=\BusStopFont,text=\busRawColor]
      at (15.45,{#1-1.84}) {$\busRawMark$};
    \pgfmathsetmacro{\busConstantThreshold}{ln(0.5)}%
    \ifdim #6pt>\busConstantThreshold pt
      \def\busConstantColor{busTeal}\def\busConstantMark{\checkmark}%
    \else
      \def\busConstantColor{busRed}\def\busConstantMark{\times}%
    \fi
    \node[font=\BusStopFont,text=busInk]
      at (14.45,{#1-2.34}) {Constant threshold $\tau=\log 0.5$};
    \node[font=\BusStopFont,text=\busConstantColor]
      at (14.30,{#1-2.78}) {$\BusStopNumber{#6}<\tau$};
    \node[draw=\busConstantColor,fill=\busConstantColor!18,rounded corners=3pt,
          minimum size=0.44cm,line width=0.55pt,font=\BusStopFont,text=\busConstantColor]
      at (15.45,{#1-2.78}) {$\busConstantMark$};
  \fi
}
\resizebox{\linewidth}{!}{%
\begin{tikzpicture}[x=1cm,y=1cm,line cap=round,line join=round,
  every node/.style={inner sep=0pt,font=\BusStopFont}]
\path[use as bounding box] (0,0.05) rectangle (16.5,13.90);
\filldraw[fill=busLight,draw=busLine,line width=0.55pt,rounded corners=6pt]
  (0.05,12.33) rectangle (16.45,13.80);
\node[font=\BusStopFont,text=busInk]
  at (8.25,13.49) {Retrieve videos with a $2$-second scene that satisfies the following temporal property:};
\node[font=\BusStopFont\bfseries,text=busInk]
  at (8.25,13.03) {A person is waiting at a bus stop until a bus appears.};
\node[font=\BusStopFont,text=busInk] at (8.25,12.58)
  {$\varphi=(\mathsf{person}\land\neg\mathsf{bus})\ \land\
   \bigcirc\!\left((\mathsf{person}\land\neg\mathsf{bus})\,\mathcal U\,\mathsf{bus}\right)$};
\node[anchor=north west,draw=busParamLine,fill=busParamLight,rounded corners=6pt,line width=0.45pt,
      inner sep=2pt,text width=16.1cm,font=\BusStopFont,text=busInk,align=left]
  at (0.05,0.73) {\textbf{LogSTOP parameters}\\[1pt]
    Adaptive threshold for $\varphi$ ($2$-second video):
    $\tau=\BusStopNumber{\BusStopThreshold}$\hfill
    Smoothing window radii: $w_{\mathsf{person}}=1$, $w_{\mathsf{bus}}=0$};
\filldraw[fill=busLight,draw=busLine,line width=0.45pt,rounded corners=4pt]
  (0.05,11.66) rectangle (16.45,12.10);
\draw[busLine,line width=0.5pt] (12.40,11.66)--(12.40,12.10);
\node[anchor=west,font=\BusStopFont\bfseries,text=busInk]
  at (0.94,11.88) {Candidate videos};
\node[font=\BusStopFont\bfseries,text=busInk]
  at (14.45,11.88) {Scene LogSTOP};
\begin{scope}[yshift=0.4cm]
\input{figures/bus_stop_until/figure_data.tex}
\end{scope}
\end{tikzpicture}%
}
\caption{LogSTOP supports efficient and interpretable ranked retrieval of videos by scenes relevant to a temporal query\protect\footnotemark. 
A passing car briefly occludes the waiting person in the second scene. Windowed smoothing and the adaptive threshold result in this video matching the query. The scores rank the clean scene above the occluded scene, followed by the non-matching scene in which no bus appears.}
\vspace{-0.1in}
\label{fig:bus_stop_example}
\label{fig:method_overview}
\endgroup
\end{figure}

\footnotetext{Videos are scenes from \url{https://www.pexels.com/video/people-waiting-for-the-bus-5030855/}.}

%% file: figures/bus_stop_until/figure_parameters.tex
\def\BusStopThreshold{-1.946154319487}
\def\BusStopPersonReference{0.50}
\def\BusStopBusReference{0.50}

%% file: figures/bus_stop_until/figure_data.tex
\BusStopRow{10.85}{1}{Both person and bus appear as expected}{6.0--8.0}{-0.5681}{-0.5708}{1/{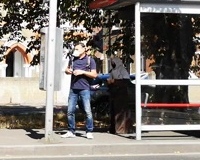}/Visible/{figures/bus_stop_until/figure_assets/clean/frame_06.0s.tex},
2/{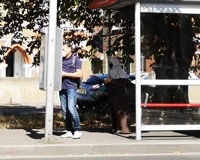}/Visible/{figures/bus_stop_until/figure_assets/clean/frame_06.5s.tex},
3/{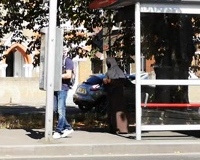}/Visible/{figures/bus_stop_until/figure_assets/clean/frame_07.0s.tex},
4/{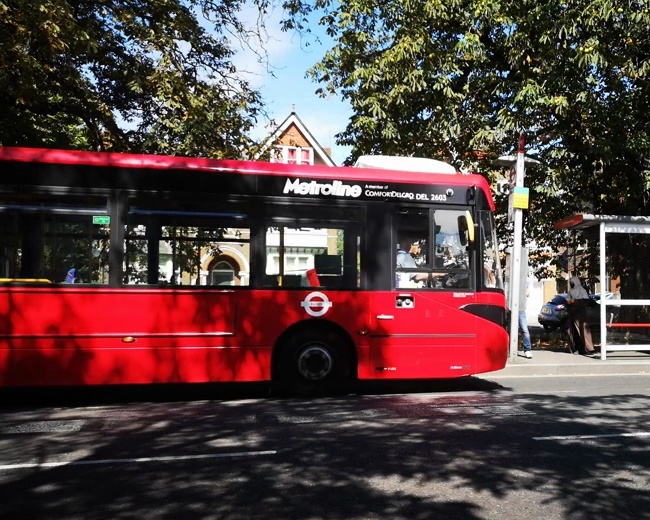}/Bus present/{figures/bus_stop_until/figure_assets/clean/frame_08.0s.tex}}
\BusStopRow{7.25}{2}{Person is briefly occluded}{4.5--6.5}{-2.0900}{-1.8811}{1/{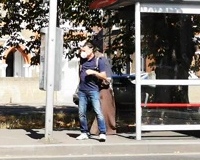}/Visible/{figures/bus_stop_until/figure_assets/occluded/frame_04.5s.tex},
2/{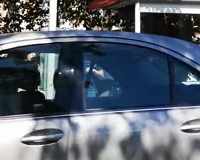}/Occluded/{figures/bus_stop_until/figure_assets/occluded/frame_05.0s.tex},
3/{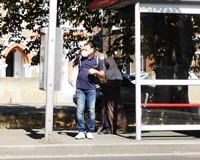}/Visible again/{figures/bus_stop_until/figure_assets/occluded/frame_05.5s.tex},
4/{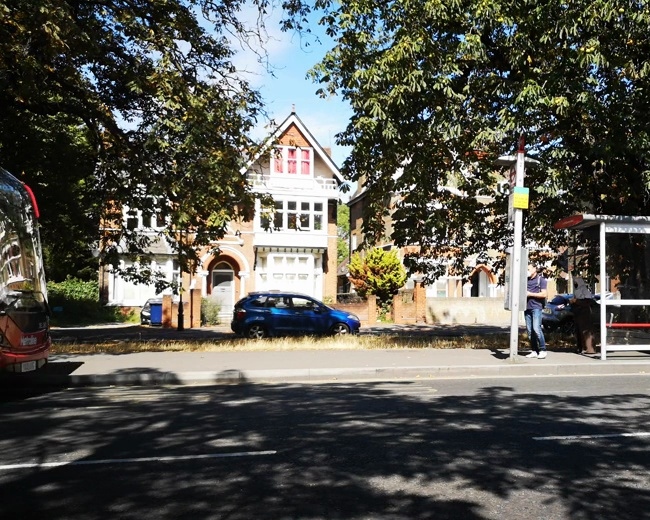}/Bus appears/{figures/bus_stop_until/figure_assets/occluded/frame_06.5s.tex}}
\BusStopRow{3.65}{3}{No bus in the scene}{2.0--4.0}{-2.2881}{-2.2913}{1/{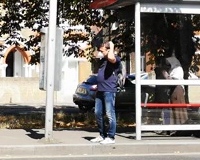}/Visible/{figures/bus_stop_until/figure_assets/no_bus/frame_02.0s.tex},
2/{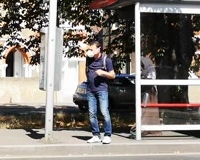}/Visible/{figures/bus_stop_until/figure_assets/no_bus/frame_03.0s.tex},
3/{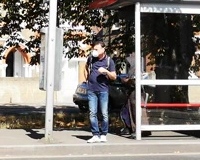}/Visible/{figures/bus_stop_until/figure_assets/no_bus/frame_03.5s.tex},
4/{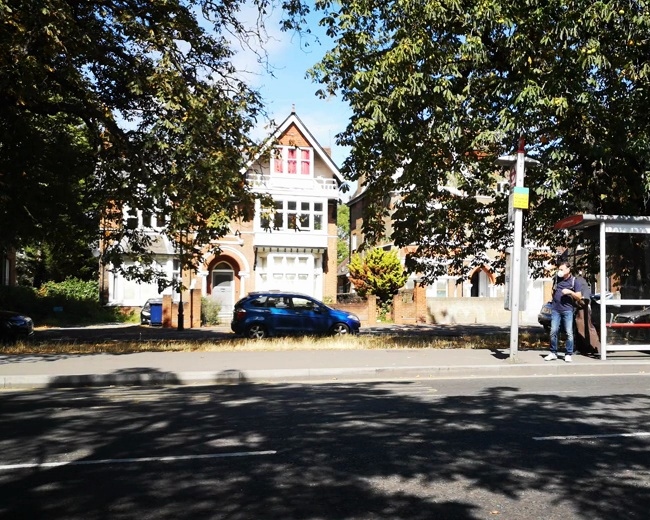}/Bus absent/{figures/bus_stop_until/figure_assets/no_bus/frame_04.0s.tex}}

%% file: src/setup.tex
\section{Representing Temporal Properties over Prediction Sequences}
\label{sec:def}

Let \(X = [x_1, \ldots, x_T]\) 
denote a sequence of data items of length \(T\), where \(x_t\) denotes the data item at timestep \(1 \leq t \leq T\). 
Let \(\mathcal{X}\) denote the set of all such sequences, and \(\mathcal{T}\) denote the set of all timesteps \(\{1, \ldots, \mathbb{N}\}\).
Further, let \(C\) denote a finite set of local properties of interest. 
In general, \(X\) and \(C\) could correspond to sequences of arbitrary modalities and properties respectively,
including but not limited to objects or actions in videos, and speakers or emotions in audio clips.
While these sequences could be over continuous time, we assume that they are discretized into \(T\) timesteps for simplicity.

We assume that there exists a true labeling function \(y : \mathcal{X} \times C \times \mathcal{T} \mapsto \{0, 1\}\) 
such that \(y(X, c, t) = 1\)
if the property \(c \in C\) is expressed by \(x_t\), and
\(y(X, c, t) = 0\)
otherwise. 
For example,
\(X\) could correspond to a video with frames \(x_t\) at timestep \(t\) and \(C\) could be objects of interest such as "car" and "pedestrian". In this case, \(y(X, c, t) = 1\) would indicate that object \(c\) is present in frame \(x_t\).

In this work, we are interested in temporal compositions of these local properties. 
For example, given the true labels for the object "car" in individual frames, 
how can we define the label for a car being present in all frames or alternatively any frame? 
Or, given the true labels for "car" and "pedestrian", 
how can we define the label for a car being present in all frames until a pedestrian is detected? 

We find that Linear Temporal Logic (LTL), 
widely used for formal specification and verification of reactive systems ~\citep{ltl}, 
provides a suitable \textit{specification language} for expressing such temporal properties. 
Formally, a temporal property \(\varphi\) over local properties \(C\) can be expressed as follows:  
\[\varphi \coloneq \top \mid \bot \mid c \mid \neg c \mid \varphi_1 \land \varphi_2 \mid \varphi_1 \lor \varphi_2 \mid \bigcirc \varphi \mid \BoxOp{I} \varphi \mid \varphi_1 \UntilOp{I} \varphi_2 \] 
where, \(c \in C\) is a local property and \(\varphi_1, \varphi_2\) are temporal properties. \(\neg\) is only applied to local properties, while \(\land\) and \(\lor\) are the logical \textit{and} and \textit{or} operators respectively. \(\bigcirc\), \(\BoxOp{I}\) and \(\UntilOp{I}\)  are temporal operators \textit{Next}, \textit{Always}, and \textit{Until} respectively. Other temporal operators such as "Eventually \(\varphi\)" (\(\DiaOp{I} \varphi\)) can then be derived as \(\top \UntilOp{I} \varphi\). 

This language can now be used to represent properties from the previous examples. For instance,  the temporal properties for a car being present in all frames and any frame in a video can be represented as \(\varphi = \BoxOp{I}\, \mathsf{car}\) and \(\varphi = \DiaOp{I}\, \mathsf{car}\) respectively. Furthermore, the property that a car is present in all frames until a pedestrian is present can be represented as \(\varphi = \mathsf{car} \UntilOp{I} \mathsf{pedestrian}\).

The ground truth labeling function \(y(X, c, t)\) for local properties can be lifted to \(y(X, \varphi, t)\), meaning that the sequence \(X\) expresses temporal property \(\varphi\) starting at timestep \(t\), using the standard semantics for LTL over finite sequences~\citep{ltlf} as follows: for \(1 \leq t \leq T\),

\begin{itemize}
    \item \(y(X, \top, t) = 1\) and \(y(X, \bot, t) = 0\)
    \item \(y(X, \neg c, t) = 1\) iff \(y(X, c, t) = 0\)
    \item \(y(X, \varphi_1 \land \varphi_2, t) = 1\) iff \(y(X, \varphi_1, t) = 1\) and \(y(X, \varphi_2, t) = 1\)
    \item \(y(X, \varphi_1 \lor \varphi_2, t) = 1\) iff \(y(X, \varphi_1, t) = 1\) or \(y(X, \varphi_2, t) = 1\)
    \item \(y(X, \bigcirc\varphi, t) = 1\) iff \(t < T\) and \(y(X, \varphi, t+1) = 1\)
    \item \(y(X, \BoxOp{I}\varphi, t) = 1\) iff \(y(X, \varphi, t') = 1\) for all \(t \leq t' \leq T\)
    \item \(y(X, \varphi_1 \UntilOp{I} \varphi_2, t) = 1\) iff there exists a \(t \leq t' \leq T\) such that \(y(X, \varphi_2, t') = 1\) and \(y(X, \varphi_1, t'') = 1\) for all \(t \leq t'' < t'\)
\end{itemize}

Given the true labeling functions for local properties \(y(X, c, \cdot)\), 
this semantics can perfectly determine if a sequence \(X\) expresses a temporal property \(\varphi\) 
(if and only if \(y(X, \varphi, 1) = 1\)). For instance,
a 4-frame video expresses the temporal property \(\varphi = \mathsf{car} \UntilOp{I} \mathsf{pedestrian}\) under this semantics 
with the following true labels:
\[y(X, \mathsf{car}, 1:4) = [1, 1, 1, 0], \qquad y(X, \mathsf{pedestrian}, 1:4) = [0, 0, 0, 1]\]

as \(\mathsf{pedestrian}\) is expressed at \(t = 4\) while \(\mathsf{car}\) is expressed in all frames before that.

In practice, however, we do not have access to these true labeling functions. 
We assume that noisy predictors can be used instead to provide 
\textit{scores} \(\hat{y} : \mathcal{X} \times C \times \mathcal{T} \mapsto [0, 1]\) for the label being 1. 
For instance, an object detection model such as YOLO could assign the following scores to the local properties \(\mathsf{car}\) and \(\mathsf{pedestrian}\)
from the example above:
\[\hat{y}(X, \mathsf{car}, 1:4) = [0.9,0.05,0.95,0.95], \qquad
\hat{y}(X, \mathsf{pedestrian}, 1:4) = [0.1,0.1,0.1,0.95].\]

Semantically, this is a favorable trace for \(\mathsf{car} \UntilOp{} \mathsf{pedestrian}\): pedestrian is strongly supported at \(t=4\) and car is strongly supported at most previous timesteps. However, the local property predictor contains a transient low score for car at \(t=2\), as may occur under occlusion or a missed detection.
Such transient errors, and in turn noisy predictions, are common for real datasets. 

We formally introduce the problem of assigning Scores for TempOral Properties (STOPs) as follows: 
\textit{
Given predictors for local properties \(\hat{y} : \mathcal{X} \times C \times \mathcal{T} \mapsto [0, 1]\) 
and a temporal property \(\varphi\) defined over local properties in \(C\), 
how can a score for \(\varphi\) and sequence \(X\) at time step \(1 \leq t \leq T\), \(\hat{y}(X, \varphi, t)\), be assigned?
}

%% file: src/method.tex
\section{LogSTOP: An Algorithm for Computing STOPs}
\label{sec:method}

\begin{algorithm}[t]
\caption{LogSTOP \(\hat{y}(X, \varphi, t_s, t_e, \{w_c\})\)}\label{alg:cstoptlsmooth}
\begin{algorithmic}[1]
\normalsize
\Require Sequence \(X\), temporal property \(\varphi\), current timestep \(1 \leq t_s \leq T\), end timestep \(t_s \leq t_e \leq T\), local-property-specific smoothing radii \(\{w_c\}_{c\in C}\)
\Ensure Log-space temporal score \(\hat{y}(X, \varphi, t_s, t_e, \{w_c\})\)
\State \textbf{Precompute}
    \(\tilde{y}(X,c,t) \gets
      \mathrm{avg}_{t' \in W_{c,t}} \hat{y}(X,c,t')\)
    \Comment{Windowed average smoothing}
    \label{alg:smoothing_op}
\Statex \hspace{\algorithmicindent}
    for all \(c \in C\), \(t \in [t_s,t_e]\),
\Statex \hspace{\algorithmicindent}
    where \(W_{c,t} =
    [\max\{t-w_c,t_s\},\,\min\{t+w_c,t_e\}]\).
\Function{$\mathrm{Score}$}{$X, \varphi, t, t_e, \{\tilde{y}\}$}
\State \textbf{case} \(t > t_e\): \Return \(-\infty\) \label{alg:termination}
\State \textbf{case} \(\varphi = \top\): \Return \(0\)
\State \textbf{case} \(\varphi = \bot\): \Return \(-\infty\)
\State \textbf{case} \(\varphi = c\): \Return \(\log \tilde{y}(X,c,t)\)
\State \textbf{case} \(\varphi = \neg c\): \Return \(\log (1 - \tilde{y}(X,c,t))\) \label{alg:not_op}
\State \textbf{case} \(\varphi = \varphi_1 \land \varphi_2\): \Return \(\mathrm{Score}(X, \varphi_1, t, t_e, \{\tilde{y}\}) + \mathrm{Score}(X, \varphi_2, t, t_e, \{\tilde{y}\})\) \label{alg:and_op}
\State \textbf{case} \(\varphi = \varphi_1 \lor \varphi_2\):
    \State \quad \(s_1 \gets \mathrm{Score}(X, \varphi_1, t, t_e, \{\tilde{y}\})\); \(s_2 \gets \mathrm{Score}(X, \varphi_2, t, t_e, \{\tilde{y}\})\)
    \State \quad \Return \(\log\!\left(e^{s_1} + e^{s_2} - e^{s_1+s_2}\right)\) \label{alg:or_op}
\State \textbf{case} \(\varphi = \bigcirc\, \varphi'\): \Return \(\mathrm{Score}(X, \varphi', t+1, t_e, \{\tilde{y}\})\) \label{alg:next_op}
\State \textbf{case} \(\varphi = \BoxOp{I}\, \varphi'\): \Return \(\begin{cases}
    \mathrm{Score}(X, \varphi', t, t_e, \{\tilde{y}\}) & \text{if } t = t_e,\\
    \mathrm{Score}(X, \varphi' \land \bigcirc\, \BoxOp{I}\, \varphi', t, t_e, \{\tilde{y}\}) & \text{otherwise}
\end{cases}\) \label{alg:always_op}
\State \textbf{case} \(\varphi = \varphi_1 \UntilOp{I} \varphi_2\): \Return \(\begin{cases}
    \mathrm{Score}(X, \varphi_2, t, t_e, \{\tilde{y}\}) & \text{if } t = t_e,\\
    \mathrm{Score}(X, \varphi_2 \lor (\varphi_1 \land \bigcirc\,(\varphi_1 \UntilOp{I} \varphi_2)), t, t_e, \{\tilde{y}\}) & \text{otherwise}
\end{cases}\) \label{alg:until_op}
\EndFunction
\State \Return \(\mathrm{Score}(X, \varphi, t_s, t_e, \{\tilde{y}\})\)
\end{algorithmic}
\label{alg:LogSTOP_complete}
\end{algorithm}

In this work, we propose a practical solution to the STOP problem designed for the downstream applications of query matching and ranked retrieval.
While we formally define the setups for these tasks later, a key observation is that sequences with stronger evidence for a temporal property should receive higher scores for both tasks. Further, for query matching, a suitable threshold should be available to separate matching and non-matching sequences.

Inspired by quantitative semantics for LTL~\citep{FAINEKOS20094262,donze2010robust,Akazaki2015TimeRI,gmrob}, we propose a scoring function \textit{LogSTOP}, 
that recursively computes a log-space score 
for a sequence \(X[t_s:t_e]\) satisfying temporal property \(\varphi\) 
, given local property scores \(\hat{y}(X,c,t) \in [0,1]\), start and end timesteps \(t_s\) and \(t_e\), and local-property-specific smoothing radii \(\{w_c\}_{c\in C}\) (Algorithm~\ref{alg:LogSTOP_complete}).
LogSTOP provides a solution to the STOP problem with \(t_s = t\) and \(t_e = T\).

\parhead{Temporal smoothing of local property predictions.}
Given a sequence of predictions for each local property \(c \in \varphi\), LogSTOP first employs a windowed-average smoothing step with a radius of size \(w_c\). This step is applied once over the target evaluation interval \([t_s,t_e]\), before the recursive temporal aggregation. When \(w_c=0\), no smoothing is applied and the local probability at the current timestep is used directly. When \(w_c>0\), the post-processed local evidence is
\[
\tilde{y}(X,c,t)
=
\frac{1}{|W_{c,t}|}
\sum_{t' \in W_{c,t}} \hat{y}(X,c,t')
,
\quad
W_{c,t} =
\left[\max\{t-w_c,t_s\},\,\min\{t+w_c,t_e\}\right].
\]

This step targets short-lived local prediction errors, such as transient low scores for present properties or transient high scores for absent ones, which can disproportionately affect temporal properties such as \(\BoxOp{}\, c\), \(\DiaOp{}\, c\), or \(c_1 \UntilOp{} c_2\). 
In the running example, the low car score at \(t=2\) can be viewed as a transient detection error. With \(w_{\mathsf{car}}=w_{\mathsf{pedestrian}}=1\), the smoothed car and pedestrian evidence sequences become \([0.475, 0.633, 0.650, 0.950]\) and \([0.100, 0.100, 0.383, 0.525]\), respectively, increasing the LogSTOP for \(\mathsf{car}\UntilOp{I}\mathsf{pedestrian}\) from \(-1.541\) to \(-1.247\).

\textit{\underline{Choosing a smoothing window}:} 
We propose selecting \(w_c\) on a held-out data split with true labels as follows: given a set of choices \(W\) for each \(w_c\), select the choice that maximizes the ratio of correct to total evidence updates. Concretely, for a given \(w\), a correct update includes a frame \(t\) in sequence \(X\) if smoothing increases the local evidence \(\tilde{y}(X, c, t) > \hat{y}(X, c, t)\) when the true label \(y(X, c, t) = 1\) (or decreases it otherwise). If no \(w \in W\) reports a correct to total ratio over \(0.5\), \(w_c = 0\). This selection strategy ensures that smoothing has a \textit{net positive effect} on the prediction sequences. We provide more details on this selection setup in practice when discussing experimental results in Section~\ref{sec:experiments}.

\parhead{Recursive definitions of Boolean and Temporal operators.}
Algorithm~\ref{alg:LogSTOP_complete} defines LogSTOP recursively over the structure of \(\varphi\), after smoothing of local property evidence. Negation reverses support for a local property, conjunction aggregates support for both sub-properties, and disjunction aggregates support for either sub-property. The Next operator shifts evaluation by one timestep. The Always and Until operators recursively compose evidence across timesteps in log-space. The resulting score \(\hat{y}(X,\varphi,t_s,t_e,\{w_c\})\) is thus obtained by recursively aggregating smoothed local evidence over the syntax of the temporal property. For the running example, LogSTOP decomposes the computation for \(\varphi = \mathsf{car} \UntilOp{I} \mathsf{pedestrian}\) as:
\[
\hat{y}(X,\varphi,1,4,\{w_c\})
=
\hat{y}\big(X,\mathsf{pedestrian} \lor (\mathsf{car} \land \bigcirc(\mathsf{car} \UntilOp{} \mathsf{pedestrian})),1,4,\{w_c\}\big).
\]
The score is computed by evaluating whether \(\mathsf{pedestrian}\) is expressed at \(t=1\), or whether \(\mathsf{car}\) is supported at \(t=1\) and the score for \(\mathsf{car} \UntilOp{} \mathsf{pedestrian}\) is recursively computed from \(t=2\), and so on.

\textit{\underline{Efficient computation via dynamic programming}:}
The recursive definition in Algorithm~\ref{alg:LogSTOP_complete} can be evaluated using dynamic programming over sub-properties and timesteps. The key observation is that the score for any sub-property at timestep \(t\) depends only on a constant number of scores at timestep \(t\) and at the next timestep \(t+1\). Since the smoothing step can be computed in constant time per local property and timestep using prefix sums, and the resulting sub-problems can be cached, the overall time complexity is linear in both the sequence length and the temporal property size. Concretely, given a sequence of length \(T\) and a temporal property \(\varphi\) with size \(|\varphi|\), LogSTOP with local-property-specific smoothing radii \(\{w_c\}_{c\in C}\) can be computed using Algorithm~\ref{alg:LogSTOP_complete} in \(\mathcal{O}(T \cdot |\varphi|)\) time. We present a proof in Appendix~\ref{proof:complexity}.

\parhead{Monotonicity of LogSTOP in post-processed local evidence.}
LogSTOP is monotone in the post-processed local evidence for temporal properties where each local property has a unique polarity. Informally, increasing evidence for a positively occurring local property, or decreasing evidence for a negatively occurring one, cannot decrease the final LogSTOP score. For example, in \(\mathsf{car}\UntilOp{}\mathsf{pedestrian}\), both local properties occur positively; increasing the car evidence at the first timestep in the running example from \(0.9\) to \(0.95\), with all other evidence unchanged, raises the LogSTOP from \(-1.541\) to \(-1.511\). This makes LogSTOP useful for ranked retrieval and threshold-based matching, where scores should respond predictably to improvements in local evidence. We state and prove the formal result in Appendix Proposition~\ref{thm:monotonicity}.

The guarantee applies when each local property has a unique polarity in the formula; mixed-polarity formulas such as \(p\UntilOp{}\neg p\) do not have a natural global monotonicity direction because increasing \(p\) can help one part of the formula while hurting another. The guarantee is also useful only when local evidence is meaningful, i.e., higher local scores correspond to higher empirical likelihood of the local property being present. In our experiments, we assess this using reliability curves and use calibration as a preprocessing step to improve the quality of the local evidence.

\parhead{Probabilistic solutions to the STOP problem.}
Exact probabilistic STOP computation would require modeling the joint distribution of local properties over time and marginalizing over all temporally valid satisfaction events. Prior automaton-based approaches~\cite{yang2023specification,Choi2024TowardsNV} assume independence and construct a sequence model with \(\mathcal{O}(T \cdot 2^{|C|})\) states, which is costly for retrieval over many candidate sequences. LogSTOP is instead a pragmatic scoring surrogate optimized for efficient ranking and matching, rather than a calibrated probabilistic semantics.

%% file: src/applications.tex
\section{Applications of LogSTOP}

\subsection{LogSTOP for Query Matching}
\label{sec:logstop_qm}

We define query matching with temporal properties as the task of predicting whether a temporal property/query is expressed by a sequence. 
LogSTOP matches a sequence \(X\) of length \(T\) with query \(\varphi\) by comparing \(\hat{y}(X, \varphi, 1, T, \cdot)\) with a threshold. Existing works on video query matching~\citep{yang2023specification,Choi2024TowardsNV} use a constant threshold, e.g., \(\tau = \log 0.5\), but such thresholds do not account for query structure and sequence length. We therefore define a length- and query-adaptive threshold with respect to a target local evidence level \(\alpha\). For each local property \(c\), let \(r_c(\alpha)\) denote the local score threshold corresponding to empirical detection probability \(\alpha\), estimated on a held-out split. Let \(X_\alpha\) be a reference sequence of length \(T\), where every local property \(c\) is assigned score \(r_c(\alpha)\) at every timestep. The adaptive matching threshold is:
\[\tau_\alpha(\varphi, T, \{w_c\}) = \hat{y}(X_\alpha, \varphi, 1, T, \{w_c\})\]

A sequence matches when its LogSTOP exceeds this reference score. For perfectly calibrated local property predictions, \(r_c(\alpha)=\alpha\); in practice, we estimate \(r_c(\alpha)\) from reliability curves over a held out dataset. For the running \(\mathsf{car}\UntilOp{}\mathsf{pedestrian}\) example, a held-out split gives \(r_{\mathsf{car}}(0.5)=0.5\) and \(r_{\mathsf{pedestrian}}(0.5)=0.15\), under which this adaptive threshold correctly classifies the smoothed trace as a match, unlike the constant \(\log 0.5\) threshold.
Query matching with this adaptive threshold requires \(\mathcal{O}(T \cdot |\varphi|)\) time, since both the sequence and reference scores are computed with Algorithm~\ref{alg:cstoptlsmooth}.

\subsection{LogSTOP for Ranked Retrieval}

LogSTOP can also be used for ranked retrieval of sequences relevant to temporal properties of interest. Formally, given a database of \(N\) sequences \(\mathcal{D} = \{X_1, \ldots, X_N\}\), a temporal property \(\varphi\), and a range of event lengths \([T_{lo}, T_{hi}]\), the goal is to rank each \(X_i\) according to whether it contains a subsequence \(X_i[t:t']\) of length \((t' - t + 1)\in [T_{lo}, T_{hi}]\) that expresses \(\varphi\). An example query is ``videos with a 10 to 20 second scene where a person is sitting until they stand up.'' Unlike query matching, ranked retrieval emphasizes the relative ordering of sequences, and relevance may depend on only a short subsequence rather than the full sequence.

A sequence is relevant if it contains at least one subsequence that strongly matches the query, so we define its retrieval score as the maximum LogSTOP over all subsequences with lengths in \([T_{lo}, T_{hi}]\), and rank sequences by this score. The detailed algorithm is given in Appendix~\ref{appendix:retrieval_algorithm}.
Given a database \(\mathcal{D}\) of \(N\) sequences with maximum length \(T\), a temporal property \(\varphi\), target subsequence length range \([T_{lo}, T_{hi}]\), retrieval count \(k\), and maximum smoothing radius \(w_{\max}\), this computes the ranked list in \(\mathcal{O}(N \cdot ((T-T_{lo}+1) \cdot T_{hi} \cdot (1+w_{\max}) \cdot |\varphi| + \log k))\) time. Here \(w_{\max}\) is a small hyperparameter, with \(w_{\max} \ll T\) in our experiments. The proof is given in Appendix~\ref{proof:retrieval_complexity}. For the special case where only sequence-level relevance is important (\(T_{lo}=T_{hi}=T\)), this reduces to \(\mathcal{O}(N \cdot (T \cdot (1+w_{\max}) \cdot |\varphi| + \log k))\).

%% file: src/experiments.tex
\section{Empirical Evaluation}
\label{sec:experiments}

\subsection{Controlled benchmarks for temporal query matching and ranked retrieval}

We introduce two controlled benchmarks for query matching and ranked retrieval, with a diverse set of 15 temporal property templates from five categories: simple temporal operators, Boolean over temporal operators, temporal over Boolean operators, temporal over temporal operators, and mixed Boolean-temporal compositions. The templates are instantiated over three existing datasets with frame/segment-level annotations for local properties:
RealTLV~\citep{Choi2024TowardsNV}
for objects in videos (6 classes),
IEMOCAP~\citep{busso2008iemocap}
for emotions in speech (4 classes), and
AVA~\citep{gu2018ava} 
for actions in videos (80 classes). We construct sequence level labels for temporal properties using the standard LTL semantics on ground-truth annotations for local properties. \textit{There are no existing benchmarks that evaluate query matching and ranked retrieval on 
video and speech sequences with the breadth of temporal and local properties considered in this work. }
The full template list and benchmark construction details are provided in Appendix~\ref{appendix:datasets}. We present a brief discussion on benchmark statistics here:

\parhead{The QMTP benchmark.}
The QMTP benchmark evaluates query matching with temporal properties over objects in video and emotions in speech sequences.
\textbf{QMTP-video} consists of \(7468\) samples 
(\(3750\) matching and \(3718\) non-matching) 
with \(10-50\) frames per sample.
\textbf{QMTP-speech} contains \(3300\) samples (balanced), 
including speech sequences with \(5-30\) segments per sample.

\parhead{The TP2VR benchmark.}
The TP2VR benchmark evaluates ranked retrieval of video sequences given temporal property queries over objects and actions.
The \textbf{TP2VR-objects dataset} consists of \(746\) videos with \(39-199\) frames, collected from the RealTLV dataset, and \(42\) queries over objects. Each query corresponds to \(25-50\) frame temporal events and is relevant to no more than \(250\) videos in the dataset.
Similarly, the \textbf{TP2VR-actions dataset} consists of \(952\) videos with \(300\) frames each, collected from 1-min segments of videos in the AVA dataset,
with \(70\) queries over actions.
Each query corresponds to \(10\)-second temporal events and is relevant to no more than \(50\) videos in the dataset.

\subsection{Evaluation protocol}

\parhead{Calibration and hyperparameter selection for LogSTOP.}
For QMTP-video, we use a disjoint held-out split of \(37\) NuScenes videos to select all data-dependent LogSTOP parameters. 
We choose the smoothing windows \(\{w_c\}\) from \(\{1, 2, \ldots, 5\}\) using the protocol described in Section~\ref{sec:method}. For query matching, we set the local evidence level \(\alpha = 0.5\) and compute reference local thresholds \(r_c(\alpha=0.5)\) using the protocol described in Section~\ref{sec:logstop_qm}.
For QMTP-video, we also use this dataset for per-class isotonic calibration of object detectors, to evaluate the effect of calibration on query matching. Details of this calibration setup are provided in Appendix~\ref{appendix:baselines_qm}.
For QMTP-speech and TP2VR-actions, where no comparable held-out split is available, we use a fixed protocol without post-hoc calibration or held-out class-specific tuning: we set \(w_c=1\) for all local properties and, for query matching, use the adaptive threshold with \(r_c(\cdot)=0.5\). Details are provided in Appendix~\ref{appendix:datasets}.

\parhead{Metrics.}
We measure balanced accuracy for query matching. For retrieval, we report Recall at \(r\) (R@\(r\), where \(r\) is the number of relevant results), Precision (P@\(\{1, r\}\)), mean average precision (mAP), and mean rank of the first relevant result (MnR).

\subsection{Results}

\input{graphics/tables/query_matching_main_results}

\parhead{R1. LogSTOP is effective for temporal query matching.}
For query matching, LogSTOP uses calibrated YOLOv8~\citep{yolov8_ultralytics}, OWLv2~\citep{minderer2023scaling}, and Grounding DINO~\citep{liu2024grounding} scores for video, and HuBERT~\citep{superb} scores for speech. We compare against LVLMs (\texttt{Video-LLava-7B}~\citep{videollava}, \texttt{LongVA-7B}~\citep{zhang2024longva}, \texttt{InternVL2-26B}~\citep{chen2024internvl}), LALMs (\texttt{Qwen-Audio-Chat}~\citep{Qwen-Audio}, \texttt{Qwen2-Audio-7B-Instruct}~\citep{Qwen2-Audio}), a supervised DistilBERT trace classifier trained on YOLO detections over \(15\)K NuScenes query-matching samples, and NSVS-TL~\citep{Choi2024TowardsNV}. We also include Spatial Robustness, a quantitative semantics for Signal Temporal Logic (STL)~\citep{FAINEKOS20094262}, with the same models and preprocessing setup as LogSTOP. We discuss the semantics and comparison to LogSTOP in detail in Appendix~\ref{appendix:semantics}. Details of question templates, training setup, PCTL queries, and baselines are in Appendix~\ref{appendix:baselines_qm}.

Table~\ref{tab:results_qm} shows the balanced accuracies of different methods on the QMTP-video and QMTP-speech datasets.
On QMTP-video, LogSTOP with calibrated YOLO reaches 0.80 overall balanced accuracy, outperforming the strongest LVLMs by 17\%; on QMTP-speech, LogSTOP with HuBERT reaches 0.84, outperforming the LALMs by 16--21\%. The gains are largest on categories involving temporal or Boolean composition, suggesting that current LVLMs/LALMs struggle with logically composed queries.
The supervised DistilBERT trace classifier reaches 0.67 overall on QMTP-video, 13\% below LogSTOP with calibrated YOLO, with larger gaps on composition-heavy categories: 12\% on Boolean-over-temporal, 25\% on temporal-over-Boolean, and 14\% on mixed formulas. Among temporal-logic baselines, LogSTOP improves over NSVS-TL by 22\% overall. STL Spatial Robustness is much stronger, reaching up to 0.79 overall, showing that quantitative temporal aggregation with the STOP preprocessing protocol already gives large gains over generic and learned baselines. LogSTOP matches or improves over STL Spatial Robustness across calibrated detectors, with gains of 4\% for Grounding DINO and 3\% for YOLO, giving more consistent performance across local predictors. We also present a comparison with \texttt{InternVL2-26B} in Appendix~\ref{appendix:internvl}.

\input{graphics/tables/retrieval_results}

\parhead{R2. LogSTOP improves top-rank behavior for ranked retrieval.}
For ranked retrieval, LogSTOP uses object detectors (YOLOv8, Grounding DINO, OWLv2) and an action detector (Detectron2~\citep{wu2019detectron2} with SlowR50~\citep{feichtenhofer2019slowfast}). Since no prior method is designed for temporal property to sequence retrieval, we compare against zero-shot text-to-video retrieval baselines \texttt{mPLUG}~\citep{li2022mplug} and \texttt{CaptionSim}, as well as DistilBERT and STL Spatial Robustness; details are in Appendix~\ref{appendix:baselines_vr}.

Table~\ref{tab:retrieval_results}
presents the results for ranked retrieval on the TP2VR benchmark.
The performance of all methods on TP2VR-actions is lower than that on TP2VR-objects due to the significantly higher number of classes (\(80\) actions) and lower number of relevant results (\(21\) relevant videos on average). 
On TP2VR-objects, LogSTOP with OWLv2 achieves the best mAP and R@\(r\), improving over \texttt{CaptionSim} by 28\% and 23\%, retrieving a relevant video at rank 1 for 81\% of queries, and reducing MnR from 3.1 to 1.7.
DistilBERT improves over \texttt{mPLUG} on mAP and R@\(r\), but with the same YOLO detections remains 10\% below LogSTOP in P@1 and retrieves the first relevant result much later on average (MnR 9.1 vs. 4.8). STL Spatial Robustness is again the strongest controlled baseline, confirming that explicit temporal aggregation is useful for retrieval. LogSTOP gives higher top-rank precision across predictors, improving P@1 over STL Spatial Robustness by 19\% with YOLO, 5\% with Grounding DINO, and 12\% with OWLv2, while matching or slightly improving mAP and R@\(r\).

On TP2VR-actions, LogSTOP with SlowR50 improves over the \texttt{CaptionSim} by \(18\%\) in mAP and retrieves the first relevant result earlier, with \(24\%\) higher P@1 and better mean rank; 
LogSTOP retrieves the first relevant result at rank \(8.3\) while  \texttt{CaptionSim} and \texttt{mPLUG} retrieve it at ranks \(>20\). 
With the same SlowR50 scores, LogSTOP matches STL Spatial Robustness on P@1 but reduces MnR from \(11.7\) to \(8.3\), retrieving the first relevant result earlier.
We present examples for retrieval in Appendix~\ref{appendix:examples}, including qualitative comparison to STL Robustness for ranking in Appendix~\ref{appendix:semantics}.

%% file: graphics/tables/query_matching_main_results.tex
\begin{table}[t]
\caption{Query matching balanced accuracy on QMTP-video and QMTP-speech
by temporal property category ($\uparrow$).
The best / second-best distinct scores
are \textbf{bold} / \underline{underlined} per dataset per category. Results with \texttt{InternVL2-26B} are presented in Appendix~\ref{appendix:internvl}.}
\vspace{0.1in}
\label{tab:results_qm}
\centering
\begingroup
\setlength{\tabcolsep}{4pt}
\renewcommand{\arraystretch}{1.08}
\begin{tabular*}{\linewidth}{@{\extracolsep{\fill}}lccccc@{\hspace{12pt}}c@{}}
\toprule
Method / local predictor & Simple
& \shortstack{Bool. over\\ Temp.}
& \shortstack{Temp. over\\ Bool.}
& \shortstack{Temp. over\\ Temp.}
& Mixed & Overall \\
\midrule
\multicolumn{7}{@{}c}{\textbf{QMTP-video}} \\
\addlinespace[2pt]
Video-LLaVA-7B       & 0.50 & 0.50 & 0.50 & 0.50 & 0.50 & 0.50 \\
LongVA-7B           & 0.68 & 0.63 & 0.62 & 0.63 & 0.61 & 0.63 \\
DistilBERT (trained on YOLO)          & 0.76 & 0.66 & 0.55 & 0.77 & 0.67 & 0.67 \\
NSVS-TL             & 0.67 & 0.63 & 0.51 & 0.64 & 0.50 & 0.58 \\
\addlinespace[4pt]
\multicolumn{7}{@{}l}{\textit{Spatial Robustness (calibrated predictors)}} \\
\quad Grounding DINO & 0.73 & 0.67 & 0.73 & 0.73 & 0.72 & 0.72 \\
\quad OWLv2 & \underline{0.82} & 0.74 & \underline{0.78} & \textbf{0.79} & \underline{0.80} & \underline{0.79} \\
\quad YOLO  & 0.79 & 0.73 & 0.77 & \underline{0.78} & 0.78 & 0.77 \\
\addlinespace[4pt]
\multicolumn{7}{@{}l}{\textit{LogSTOP (calibrated predictors)}} \\
\quad Grounding DINO & 0.78 & 0.73 & 0.74 & 0.75 & 0.79 & 0.76 \\
\quad OWLv2 & 0.81 & \underline{0.77} & \underline{0.78} & \underline{0.78} & \textbf{0.81} & \underline{0.79} \\
\quad YOLO  & \textbf{0.83} & \textbf{0.78} & \textbf{0.80} & \textbf{0.79} & \textbf{0.81} & \textbf{0.80} \\
\midrule
\multicolumn{7}{@{}c}{\textbf{QMTP-speech}} \\
\addlinespace[2pt]
Qwen-Audio-Chat          & \underline{0.70} & 0.58 & \underline{0.71} & \underline{0.66} & \underline{0.64} & \underline{0.68} \\
Qwen2-Audio-7B-Instruct  & 0.64 & \underline{0.64} & 0.65 & 0.58 & 0.56 & 0.63 \\
\addlinespace[3pt]
LogSTOP (HuBERT)        & \textbf{0.90} & \textbf{0.77} & \textbf{0.80} & \textbf{0.83} & \textbf{0.79} & \textbf{0.84} \\
\bottomrule
\end{tabular*}
\endgroup
\vspace{-0.1in}
\end{table}

%% file: graphics/tables/retrieval_results.tex
\begin{table}[t]
\caption{Ranked retrieval results on TP2VR. We report Precision@1 (P@1), mean Average Precision (mAP), mean rank of the first relevant result (MnR), and Recall@r (R@\(r\)), where $r$ is the number of relevant videos per query.
The best/second-best scores are
\textbf{bold}/\underline{underlined} per dataset per metric.}
\vspace{0.1in}
\label{tab:retrieval_results}
\centering
\begingroup
\setlength{\tabcolsep}{4pt}
\renewcommand{\arraystretch}{1.08}
\begin{tabular*}{\linewidth}{@{\extracolsep{\fill}}lcccr@{}}
\toprule
Method / local predictor & P@1 $\uparrow$ & mAP $\uparrow$
& R@$r$ $\uparrow$ & \multicolumn{1}{c@{}}{MnR $\downarrow$} \\
\midrule
\multicolumn{5}{@{}c}{\textbf{TP2VR-objects} (42 queries; mean $r = 163$)} \\
\addlinespace[2pt]
mPLUG & 0.10 & 0.34 & 0.35 & 8.1 \\
CaptionSim (LLaVA-NeXT-Video-7B) & 0.50 & 0.36 & 0.35 & 3.1 \\
DistilBERT (trained on YOLO) & 0.26 & 0.41 & 0.43 & 9.1 \\
\addlinespace[4pt]
\multicolumn{5}{@{}l}{\textit{Spatial Robustness}} \\
\quad Grounding DINO & \underline{0.76} & 0.57 & 0.51 & 2.2 \\
\quad OWLv2 & 0.69 & \underline{0.62} & \underline{0.57} & 2.1 \\
\quad YOLO & 0.17 & 0.42 & 0.41 & 4.7 \\
\addlinespace[4pt]
\multicolumn{5}{@{}l}{\textit{LogSTOP}} \\
\quad Grounding DINO & \textbf{0.81} & 0.57 & 0.52 & \underline{2.0} \\
\quad OWLv2 & \textbf{0.81} & \textbf{0.64} & \textbf{0.58} & \textbf{1.7} \\
\quad YOLO & 0.36 & 0.42 & 0.41 & 4.8 \\
\midrule
\multicolumn{5}{@{}c}{\textbf{TP2VR-actions} (70 queries; mean $r = 21$)} \\
\addlinespace[2pt]
mPLUG & 0.07 & 0.06 & 0.05 & 60.7 \\
CaptionSim (LLaVA-NeXT-Video-7B) & \underline{0.23} & 0.09 & 0.11 & 20.4 \\
\addlinespace[3pt]
Spatial Robustness (SlowR50) & \textbf{0.47} & \underline{0.26} & \underline{0.25} & \underline{11.7} \\
LogSTOP (SlowR50) & \textbf{0.47} & \textbf{0.27} & \textbf{0.26} & \textbf{8.3} \\
\bottomrule
\end{tabular*}
\endgroup
\vspace{-0.1in}
\end{table}

%% file: src/discussion.tex
\section{Discussion}
\label{sec:discussion}

\parhead{Component ablations.}
\label{subsec:component_ablations}
To isolate the importance of calibration, windowed smoothing, and adaptive thresholding for query matching, we ran query matching on QMTP-Video and ranked retrieval on TP2VR-Objects with each component independently removed or replaced with alternatives. We present a detailed discussion of this analysis in Appendix~\ref{appendix:discussion} and summarize the key findings here:

\begin{itemize}[leftmargin=*]
    \item The adaptive threshold is the most important component for query matching. Replacing it with the constant \(\ln 0.5\) threshold reduces query matching accuracy by at least \(8.55\%\) across detectors.
    \item Performance gains with calibration are correlated with monotonicity improvements for local property predictors: OWLv2's scores follow empirical accuracy more reliably post calibration (Appendix~\ref{appendix:discussion} Figure~\ref{fig:reliability_diagram_qmtp}) and this results in a \(5.12\%\) higher query matching accuracy.
    \item Most prediction errors are \textit{transient}, i.e., concentrated in short sequences of 1-2 frames across predictors (Appendix~\ref{appendix:discussion} Figure~\ref{fig:smoothing_stats}). While this qualitatively validates the need for windowed smoothing, quantitatively query matching accuracy dips by at most \(0.83\%\) and retrieval Precision@1 by at most \(0.05\) without smoothing.
\end{itemize}

\parhead{Comparison of LogSTOP to other aggregators.}
\label{subsec:score_aggregators}
To assess whether the performance gains stem from using LogSTOP as a scoring function or primarily from access to sequences of local property scores, we compare LogSTOP against other scoring functions. As we discuss in Section~\ref{sec:experiments}, LogSTOP outperforms a trained DistilBERT transformer on query matching by \(13\%\) in accuracy (Table~\ref{tab:results_qm}) and on ranked retrieval by \(10\%\) in Precision@1 (Table~\ref{tab:retrieval_results}). While STL Spatial Robustness, a \(\min-\max\) based quantitative semantics, is competitive with LogSTOP on query matching, LogSTOP reports strictly better Precision@1 for ranked retrieval (up to \(19\%\) higher in Table~\ref{tab:retrieval_results}).

On a stratified sample of 150 queries from QMTP-Video, we also evaluate 2 LLMs as matchers: \texttt{Llama-3.1-8B-Instruct} and \texttt{Qwen3-14B} (in thinking mode). The prompts include the same frame-level YOLOv8 scores used by LogSTOP (post calibration and smoothing) and the class-specific thresholds. 
We find that the two LLMs often engage in long reasoning traces, providing a final matching decision for \(38.7-74\%\) of samples (and only around \(3\%\) for queries with nested temporal and Boolean operators). On the subset where final decisions are produced, LogSTOP outperforms Llama-3.1-8B-Instruct by \(16\%\) and Qwen3-14B by \(1.8\%\), suggesting that stronger reasoning models perform better in cases when they produce valid responses. Overall, the best LLM setting (Llama-3.1-8B-Instruct with Natural Language queries) achieves \(44.0\%\) query matching accuracy, compared to \(78.0\%\) with LogSTOP. We present more details on this experiment in Appendix~\ref{appendix:score_aggregators}.

\parhead{Other analyses.}
In Appendix~\ref{appendix:comp_efficiency}, we present empirical computational efficiency plots of LogSTOP on QMTP-Video that demonstrate that the time taken by LogSTOP scales linearly with sequence length for any given temporal property.
In Appendix~\ref{appendix:llm_translate}, we conduct an experiment on using LLMs, \texttt{Llama-3.1-8B-Instruct} and \texttt{Qwen3-14B} (in thinking mode), to translate Natural Language queries to temporal properties in LTL. We find that LLMs successfully translate \(73.3\%-84\%\) queries, with model-specific prompt designs that improve performance. 

%% file: src/related.tex
\section{Related Work}

\parhead{Neural matching and retrieval.}
General query matching methods use Large Multimodal Models (LMMs) for video~\citep{videollava,xue2024longvila,zhang2024longva} or audio~\citep{Qwen-Audio,Qwen2-Audio}, while text-to-video retrieval methods compute video-text embeddings~\citep{luo2021clip4clip,liu2022ts2,Bain21,li2022mplug,liu-etal-2025-eliot}. Although effective for semantic matching, these models often struggle with complex temporal reasoning in video~\citep{videomme,chen2024rextime,nextqa,qvhighlights,cai2024temporalbench,liu2024tempcompass} and audio~\citep{sakshi2024mmau,ghosh2023compa}. LogSTOP  accounts for the temporal structure of the query, improving performance.

\parhead{Temporal logic methods.}
Prior neuro-symbolic works represent video queries as LTL or PCTL specifications, detect local properties with models such as YOLO~\citep{yolo}, and use probabilistic model checkers such as STORM~\citep{storm}~\citep{yang2023specification,Choi2024TowardsNV} and probabilistic automata~\citep{nesya} for matching; these pipelines do not directly address transient detector errors, rely on fixed thresholds, and scale as \(\mathcal{O}(T \cdot 2^{|C|})\); LogSTOP targets scoring over noisy local prediction sequences efficiently in \(\mathcal{O}(T |\varphi|)\).
Quantitative semantics for MTL/STL are widely used for monitoring, falsification, and control synthesis~\citep{FAINEKOS20094262,donze2010robust,Akazaki2015TimeRI,gmrob}; STL Spatial Robustness aggregates with \(\min\)/\(\max\) operators~\citep{FAINEKOS20094262}, while space-time robustness rewards earlier satisfaction~\citep{donze2010robust}. These semantics were not designed for ranked retrieval over noisy local prediction sequences and we include STL Spatial Robustness as a baseline.

\parhead{Benchmarks.}
Existing video/audio benchmarks evaluate temporal understanding in natural language tasks~\citep{videomme,chen2024rextime,nextqa,qvhighlights,cai2024temporalbench,liu2024tempcompass,sakshi2024mmau,ghosh2023compa}, while retrieval datasets such as ActivityNet Captions~\citep{krishna2017dense} and DiDeMo~\citep{didemo} focus on natural language segment retrieval. QMTP and TP2VR instead evaluate fine-grained temporal properties, with labels derived from frame/segment-level annotations.

%% file: src/conclusion.tex
\section{Conclusion, Limitations and Future Work}

In this work, we study Scores for TempOral Properties (STOPs) from noisy local property scores and propose LogSTOP, an LTL-inspired quantitative semantics for assigning them. We introduce benchmarks for temporal query matching and ranked retrieval over objects/actions in videos and emotions in speech, where LogSTOP with simple neural predictors outperforms LVLMs/LALMs and text-to-video retrieval methods while matching or improving over temporal-logic baselines.

\parhead{Limitations.} LogSTOP relies on competent local property predictors, although this is often easier than obtaining reliable end-to-end neural systems for temporal queries. The full protocol also uses a held-out small but well-balanced split to tune smoothing radii and reference thresholds for matching. Finally, LogSTOP depends on the chosen local properties and associated predictors. For example, a property such as "always 2 cars" can currently be scored only if a detector for "\(\mathsf{2 \,cars}\)" is available; open-vocabulary detectors partly address this, while extending LogSTOP to more expressive logics over simpler local properties is an interesting direction~\cite{strem,huang2023laser}.

%% file: appendix/main.tex
\newpage
\appendix

\input{appendix/examples}

\input{appendix/discussion}

\input{appendix/datasets}

\input{appendix/qm_methods}

\input{appendix/retrieval_methods}

\input{appendix/queries_prompts}

\input{appendix/stl_robustness}

\input{appendix/proofs}

\input{appendix/algorithm_details}

\input{appendix/retrieval_algorithm}

\section{Other Experiment Details}
\label{appendix:experiments}

\noindent{\textbf{Compute Resources.}} 
All experiments were run on a shared cluster with the following GPUs:
eight NVIDIA A100 PCIe (80GB RAM each) and eight NVIDIA RTX A6000 (48GB RAM each).

\input{appendix/llm_ablations_prompts}

%% file: appendix/examples.tex
\section{Qualitative Retrieval Examples}
\label{appendix:examples}

\input{graphics/charts/retrieval_example.tex}

\parhead{Retrieval successes with LogSTOP.} Figure~\ref{fig:bus_car_example} and Figure~\ref{fig:stand_clap_example} present examples of ranking using LogSTOP, Spatial Robustness, and the two neural baselines: \texttt{CaptionSim} and mPLUG. 
In the first example, the video captions used by \texttt{CaptionSim} do not include smaller objects that might be relevant to the query ("car" in this example). In the second example, the two actions are mentioned in the caption but the video is ranked lower than other videos with more mentions of the actions ("stand" and "hand clap" in this case). 
This demonstrates that while caption-based methods outperform joint model embeddings (\texttt{mPLUG}), they rely on semantic similarity between captions and text to determine relevance, which might not be sufficient for effective retrieval with temporal queries.

\parhead{Retrieval failure cases with mixed Boolean-temporal properties over actions.}
We inspected samples from the TP2VR-Actions benchmark with low Precision@1 and found that the failures are predictable interactions between local detector errors (with the SlowR50 action detector) and LTL formula structure. We present examples from two common failure categories here:

\underline{\emph{Eventually and Boolean disjunction are sensitive to sparse false-positive detections:}}
The query \(\BoxOp{}\,\mathsf{smoke} \lor \DiaOp{}\,\mathsf{lift\ person}\) should retrieve a video with a 50-frame subsequence where either smoking happens throughout or a person is lifted in at least one frame. LogSTOP retrieves an irrelevant video at rank 1 because lift person is incorrectly detected with a high score of 0.746 in 4 consecutive frames. Evidence for smoke is low throughout, but the formula can be satisfied through the \(\DiaOp{}\,\mathsf{lift\ person}\) branch, so this sparse false-positive spike gives the irrelevant video a high score.

\underline{\emph{Always and Boolean conjunction are sensitive to repeated false-negative detections:}}
The query \(\BoxOp{}\,(\mathsf{write} \land \mathsf{hold\ object})\) should retrieve a video with a 50-frame subsequence where writing and holding an object happen simultaneously throughout. LogSTOP retrieves the first relevant video only at rank 51 because evidence for write is missing throughout: the highest score for write by the SlowR50 action detector is only 0.097, while evidence for hold object is high throughout. Since the formula requires both actions in every frame, repeated false-negative scores for write produce a low LogSTOP score.

Together, these examples show that LogSTOP’s failures are interpretable: the formula structure determines how local detector errors are amplified. This gives users a clear diagnostic path: improving local action detection, or adding stronger temporal denoising for sustained actions, should directly improve retrieval for these cases.

%% file: graphics/charts/retrieval_example.tex
\begin{figure}[h]
  \centering
  \input{figures/retrieval_selected_scenes/bus_car}
  \caption{An example of ranked retrieval from TP2VR-objects. LogSTOP and Spatial Robustness use Grounding DINO as the local property predictor.}
  \label{fig:bus_car_example}
\end{figure}

\begin{figure}[H]
  \centering
  \input{figures/retrieval_selected_scenes/stand_clap}
  \caption{An example of ranked retrieval from TP2VR-actions. LogSTOP and Spatial Robustness use SlowR50 as the local property predictor.}
  \label{fig:stand_clap_example}
\end{figure}

%% file: figures/retrieval_selected_scenes/bus_car.tex
\begingroup
\input{figures/retrieval_qualitative_style}
\begin{tikzpicture}[x=1cm,y=-1cm,font=\normalfont\normalsize,every node/.style={inner sep=0pt,outer sep=0pt},line cap=round]
\path[use as bounding box] (0,0) rectangle (13.97,9.15);
\filldraw[fill=rqLight,draw=rqLine,line width=.55pt,rounded corners=6pt] (0,.42) rectangle (13.97,1.80);
\node[text=rqInk] at (6.985,.70) {Retrieve videos with a scene of 25--50 frames that satisfies:};
\node[text=rqInk,font=\bfseries,align=center,text width=13.30cm] at (6.985,1.10) {A bus is present throughout, and a car appears at least once.};
\node[text=rqInk] at (6.985,1.54) {$\varphi = \square\,\mathsf{bus}\;\land\;\Diamond\,\mathsf{car}$};
\filldraw[fill=rqLight,draw=rqLine,line width=.55pt,rounded corners=5pt] (0,1.97) rectangle (13.97,3.40);
\node[anchor=west,text=rqInk] at (.15,2.18) {Full video: 40 frames};
\node[anchor=east,text=black!60] at (13.82,2.18) {time $\longrightarrow$};
\fill[black!18] (.10,2.42) rectangle (13.87,3.29);
\node[anchor=north west,opacity=.27] at (0.1700,2.54) {\includegraphics[width=1.1108cm]{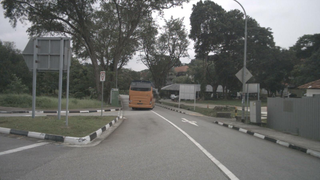}};
\node[anchor=north west,opacity=.27] at (1.3058,2.54) {\includegraphics[width=1.1108cm]{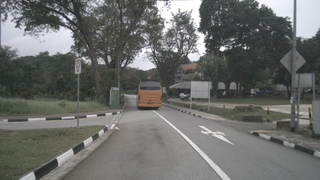}};
\node[anchor=north west,opacity=.27] at (2.4417,2.54) {\includegraphics[width=1.1108cm]{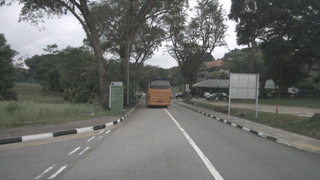}};
\node[anchor=north west,opacity=.27] at (3.5775,2.54) {\includegraphics[width=1.1108cm]{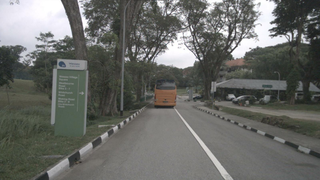}};
\node[anchor=north west,opacity=.27] at (4.7133,2.54) {\includegraphics[width=1.1108cm]{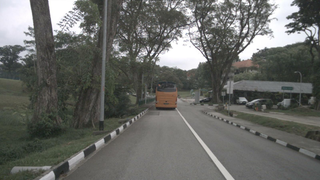}};
\node[anchor=north west,opacity=.27] at (5.8492,2.54) {\includegraphics[width=1.1108cm]{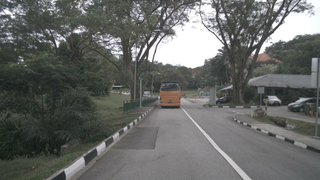}};
\node[anchor=north west,opacity=.27] at (6.9850,2.54) {\includegraphics[width=1.1108cm]{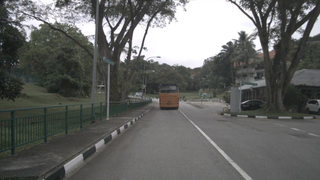}};
\node[anchor=north west,opacity=.27] at (8.1208,2.54) {\includegraphics[width=1.1108cm]{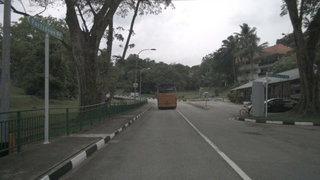}};
\node[anchor=north west,opacity=.27] at (9.2567,2.54) {\includegraphics[width=1.1108cm]{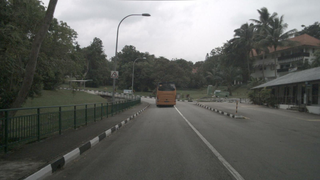}};
\node[anchor=north west,opacity=.27] at (10.3925,2.54) {\includegraphics[width=1.1108cm]{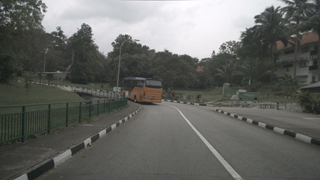}};
\node[anchor=north west,opacity=.27] at (11.5283,2.54) {\includegraphics[width=1.1108cm]{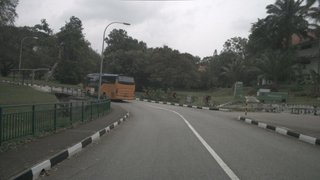}};
\node[anchor=north west,opacity=.27] at (12.6642,2.54) {\includegraphics[width=1.1108cm]{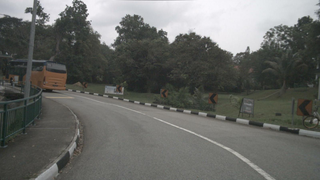}};
\fill[white,rounded corners=.6pt] (0.220,2.460) rectangle +(.16,.04);
\fill[white,rounded corners=.6pt] (0.220,3.210) rectangle +(.16,.04);
\fill[white,rounded corners=.6pt] (0.610,2.460) rectangle +(.16,.04);
\fill[white,rounded corners=.6pt] (0.610,3.210) rectangle +(.16,.04);
\fill[white,rounded corners=.6pt] (1.000,2.460) rectangle +(.16,.04);
\fill[white,rounded corners=.6pt] (1.000,3.210) rectangle +(.16,.04);
\fill[white,rounded corners=.6pt] (1.390,2.460) rectangle +(.16,.04);
\fill[white,rounded corners=.6pt] (1.390,3.210) rectangle +(.16,.04);
\fill[white,rounded corners=.6pt] (1.780,2.460) rectangle +(.16,.04);
\fill[white,rounded corners=.6pt] (1.780,3.210) rectangle +(.16,.04);
\fill[white,rounded corners=.6pt] (2.170,2.460) rectangle +(.16,.04);
\fill[white,rounded corners=.6pt] (2.170,3.210) rectangle +(.16,.04);
\fill[white,rounded corners=.6pt] (2.560,2.460) rectangle +(.16,.04);
\fill[white,rounded corners=.6pt] (2.560,3.210) rectangle +(.16,.04);
\fill[white,rounded corners=.6pt] (2.950,2.460) rectangle +(.16,.04);
\fill[white,rounded corners=.6pt] (2.950,3.210) rectangle +(.16,.04);
\fill[white,rounded corners=.6pt] (3.340,2.460) rectangle +(.16,.04);
\fill[white,rounded corners=.6pt] (3.340,3.210) rectangle +(.16,.04);
\fill[white,rounded corners=.6pt] (3.730,2.460) rectangle +(.16,.04);
\fill[white,rounded corners=.6pt] (3.730,3.210) rectangle +(.16,.04);
\fill[white,rounded corners=.6pt] (4.120,2.460) rectangle +(.16,.04);
\fill[white,rounded corners=.6pt] (4.120,3.210) rectangle +(.16,.04);
\fill[white,rounded corners=.6pt] (4.510,2.460) rectangle +(.16,.04);
\fill[white,rounded corners=.6pt] (4.510,3.210) rectangle +(.16,.04);
\fill[white,rounded corners=.6pt] (4.900,2.460) rectangle +(.16,.04);
\fill[white,rounded corners=.6pt] (4.900,3.210) rectangle +(.16,.04);
\fill[white,rounded corners=.6pt] (5.290,2.460) rectangle +(.16,.04);
\fill[white,rounded corners=.6pt] (5.290,3.210) rectangle +(.16,.04);
\fill[white,rounded corners=.6pt] (5.680,2.460) rectangle +(.16,.04);
\fill[white,rounded corners=.6pt] (5.680,3.210) rectangle +(.16,.04);
\fill[white,rounded corners=.6pt] (6.070,2.460) rectangle +(.16,.04);
\fill[white,rounded corners=.6pt] (6.070,3.210) rectangle +(.16,.04);
\fill[white,rounded corners=.6pt] (6.460,2.460) rectangle +(.16,.04);
\fill[white,rounded corners=.6pt] (6.460,3.210) rectangle +(.16,.04);
\fill[white,rounded corners=.6pt] (6.850,2.460) rectangle +(.16,.04);
\fill[white,rounded corners=.6pt] (6.850,3.210) rectangle +(.16,.04);
\fill[white,rounded corners=.6pt] (7.240,2.460) rectangle +(.16,.04);
\fill[white,rounded corners=.6pt] (7.240,3.210) rectangle +(.16,.04);
\fill[white,rounded corners=.6pt] (7.630,2.460) rectangle +(.16,.04);
\fill[white,rounded corners=.6pt] (7.630,3.210) rectangle +(.16,.04);
\fill[white,rounded corners=.6pt] (8.020,2.460) rectangle +(.16,.04);
\fill[white,rounded corners=.6pt] (8.020,3.210) rectangle +(.16,.04);
\fill[white,rounded corners=.6pt] (8.410,2.460) rectangle +(.16,.04);
\fill[white,rounded corners=.6pt] (8.410,3.210) rectangle +(.16,.04);
\fill[white,rounded corners=.6pt] (8.800,2.460) rectangle +(.16,.04);
\fill[white,rounded corners=.6pt] (8.800,3.210) rectangle +(.16,.04);
\fill[white,rounded corners=.6pt] (9.190,2.460) rectangle +(.16,.04);
\fill[white,rounded corners=.6pt] (9.190,3.210) rectangle +(.16,.04);
\fill[white,rounded corners=.6pt] (9.580,2.460) rectangle +(.16,.04);
\fill[white,rounded corners=.6pt] (9.580,3.210) rectangle +(.16,.04);
\fill[white,rounded corners=.6pt] (9.970,2.460) rectangle +(.16,.04);
\fill[white,rounded corners=.6pt] (9.970,3.210) rectangle +(.16,.04);
\fill[white,rounded corners=.6pt] (10.360,2.460) rectangle +(.16,.04);
\fill[white,rounded corners=.6pt] (10.360,3.210) rectangle +(.16,.04);
\fill[white,rounded corners=.6pt] (10.750,2.460) rectangle +(.16,.04);
\fill[white,rounded corners=.6pt] (10.750,3.210) rectangle +(.16,.04);
\fill[white,rounded corners=.6pt] (11.140,2.460) rectangle +(.16,.04);
\fill[white,rounded corners=.6pt] (11.140,3.210) rectangle +(.16,.04);
\fill[white,rounded corners=.6pt] (11.530,2.460) rectangle +(.16,.04);
\fill[white,rounded corners=.6pt] (11.530,3.210) rectangle +(.16,.04);
\fill[white,rounded corners=.6pt] (11.920,2.460) rectangle +(.16,.04);
\fill[white,rounded corners=.6pt] (11.920,3.210) rectangle +(.16,.04);
\fill[white,rounded corners=.6pt] (12.310,2.460) rectangle +(.16,.04);
\fill[white,rounded corners=.6pt] (12.310,3.210) rectangle +(.16,.04);
\fill[white,rounded corners=.6pt] (12.700,2.460) rectangle +(.16,.04);
\fill[white,rounded corners=.6pt] (12.700,3.210) rectangle +(.16,.04);
\fill[white,rounded corners=.6pt] (13.090,2.460) rectangle +(.16,.04);
\fill[white,rounded corners=.6pt] (13.090,3.210) rectangle +(.16,.04);
\fill[white,rounded corners=.6pt] (13.480,2.460) rectangle +(.16,.04);
\fill[white,rounded corners=.6pt] (13.480,3.210) rectangle +(.16,.04);
\fill[rqGreenStrip] (4.2590,2.42) -- (12.7777,2.42) -- (12.7777,3.29) -- (13.97,3.89) -- (13.97,8.19) -- (0,8.19) -- (0,3.89) -- (4.2590,3.29) -- cycle;
\begin{scope}\clip (4.2590,2.42) rectangle (12.7777,3.29);
\node[anchor=north west] at (0.1700,2.54) {\includegraphics[width=1.1108cm]{figures/retrieval_selected_scenes/overviews/bus_car/frame_002.png}};
\node[anchor=north west] at (1.3058,2.54) {\includegraphics[width=1.1108cm]{figures/retrieval_selected_scenes/overviews/bus_car/frame_004.png}};
\node[anchor=north west] at (2.4417,2.54) {\includegraphics[width=1.1108cm]{figures/retrieval_selected_scenes/overviews/bus_car/frame_008.png}};
\node[anchor=north west] at (3.5775,2.54) {\includegraphics[width=1.1108cm]{figures/retrieval_selected_scenes/overviews/bus_car/frame_012.png}};
\node[anchor=north west] at (4.7133,2.54) {\includegraphics[width=1.1108cm]{figures/retrieval_selected_scenes/overviews/bus_car/frame_014.png}};
\node[anchor=north west] at (5.8492,2.54) {\includegraphics[width=1.1108cm]{figures/retrieval_selected_scenes/overviews/bus_car/frame_018.png}};
\node[anchor=north west] at (6.9850,2.54) {\includegraphics[width=1.1108cm]{figures/retrieval_selected_scenes/overviews/bus_car/frame_022.png}};
\node[anchor=north west] at (8.1208,2.54) {\includegraphics[width=1.1108cm]{figures/retrieval_selected_scenes/overviews/bus_car/frame_024.png}};
\node[anchor=north west] at (9.2567,2.54) {\includegraphics[width=1.1108cm]{figures/retrieval_selected_scenes/overviews/bus_car/frame_028.png}};
\node[anchor=north west] at (10.3925,2.54) {\includegraphics[width=1.1108cm]{figures/retrieval_selected_scenes/overviews/bus_car/frame_032.png}};
\node[anchor=north west] at (11.5283,2.54) {\includegraphics[width=1.1108cm]{figures/retrieval_selected_scenes/overviews/bus_car/frame_034.png}};
\node[anchor=north west] at (12.6642,2.54) {\includegraphics[width=1.1108cm]{figures/retrieval_selected_scenes/overviews/bus_car/frame_038.png}};
\fill[white,rounded corners=.6pt] (0.220,2.460) rectangle +(.16,.04);
\fill[white,rounded corners=.6pt] (0.220,3.210) rectangle +(.16,.04);
\fill[white,rounded corners=.6pt] (0.610,2.460) rectangle +(.16,.04);
\fill[white,rounded corners=.6pt] (0.610,3.210) rectangle +(.16,.04);
\fill[white,rounded corners=.6pt] (1.000,2.460) rectangle +(.16,.04);
\fill[white,rounded corners=.6pt] (1.000,3.210) rectangle +(.16,.04);
\fill[white,rounded corners=.6pt] (1.390,2.460) rectangle +(.16,.04);
\fill[white,rounded corners=.6pt] (1.390,3.210) rectangle +(.16,.04);
\fill[white,rounded corners=.6pt] (1.780,2.460) rectangle +(.16,.04);
\fill[white,rounded corners=.6pt] (1.780,3.210) rectangle +(.16,.04);
\fill[white,rounded corners=.6pt] (2.170,2.460) rectangle +(.16,.04);
\fill[white,rounded corners=.6pt] (2.170,3.210) rectangle +(.16,.04);
\fill[white,rounded corners=.6pt] (2.560,2.460) rectangle +(.16,.04);
\fill[white,rounded corners=.6pt] (2.560,3.210) rectangle +(.16,.04);
\fill[white,rounded corners=.6pt] (2.950,2.460) rectangle +(.16,.04);
\fill[white,rounded corners=.6pt] (2.950,3.210) rectangle +(.16,.04);
\fill[white,rounded corners=.6pt] (3.340,2.460) rectangle +(.16,.04);
\fill[white,rounded corners=.6pt] (3.340,3.210) rectangle +(.16,.04);
\fill[white,rounded corners=.6pt] (3.730,2.460) rectangle +(.16,.04);
\fill[white,rounded corners=.6pt] (3.730,3.210) rectangle +(.16,.04);
\fill[white,rounded corners=.6pt] (4.120,2.460) rectangle +(.16,.04);
\fill[white,rounded corners=.6pt] (4.120,3.210) rectangle +(.16,.04);
\fill[white,rounded corners=.6pt] (4.510,2.460) rectangle +(.16,.04);
\fill[white,rounded corners=.6pt] (4.510,3.210) rectangle +(.16,.04);
\fill[white,rounded corners=.6pt] (4.900,2.460) rectangle +(.16,.04);
\fill[white,rounded corners=.6pt] (4.900,3.210) rectangle +(.16,.04);
\fill[white,rounded corners=.6pt] (5.290,2.460) rectangle +(.16,.04);
\fill[white,rounded corners=.6pt] (5.290,3.210) rectangle +(.16,.04);
\fill[white,rounded corners=.6pt] (5.680,2.460) rectangle +(.16,.04);
\fill[white,rounded corners=.6pt] (5.680,3.210) rectangle +(.16,.04);
\fill[white,rounded corners=.6pt] (6.070,2.460) rectangle +(.16,.04);
\fill[white,rounded corners=.6pt] (6.070,3.210) rectangle +(.16,.04);
\fill[white,rounded corners=.6pt] (6.460,2.460) rectangle +(.16,.04);
\fill[white,rounded corners=.6pt] (6.460,3.210) rectangle +(.16,.04);
\fill[white,rounded corners=.6pt] (6.850,2.460) rectangle +(.16,.04);
\fill[white,rounded corners=.6pt] (6.850,3.210) rectangle +(.16,.04);
\fill[white,rounded corners=.6pt] (7.240,2.460) rectangle +(.16,.04);
\fill[white,rounded corners=.6pt] (7.240,3.210) rectangle +(.16,.04);
\fill[white,rounded corners=.6pt] (7.630,2.460) rectangle +(.16,.04);
\fill[white,rounded corners=.6pt] (7.630,3.210) rectangle +(.16,.04);
\fill[white,rounded corners=.6pt] (8.020,2.460) rectangle +(.16,.04);
\fill[white,rounded corners=.6pt] (8.020,3.210) rectangle +(.16,.04);
\fill[white,rounded corners=.6pt] (8.410,2.460) rectangle +(.16,.04);
\fill[white,rounded corners=.6pt] (8.410,3.210) rectangle +(.16,.04);
\fill[white,rounded corners=.6pt] (8.800,2.460) rectangle +(.16,.04);
\fill[white,rounded corners=.6pt] (8.800,3.210) rectangle +(.16,.04);
\fill[white,rounded corners=.6pt] (9.190,2.460) rectangle +(.16,.04);
\fill[white,rounded corners=.6pt] (9.190,3.210) rectangle +(.16,.04);
\fill[white,rounded corners=.6pt] (9.580,2.460) rectangle +(.16,.04);
\fill[white,rounded corners=.6pt] (9.580,3.210) rectangle +(.16,.04);
\fill[white,rounded corners=.6pt] (9.970,2.460) rectangle +(.16,.04);
\fill[white,rounded corners=.6pt] (9.970,3.210) rectangle +(.16,.04);
\fill[white,rounded corners=.6pt] (10.360,2.460) rectangle +(.16,.04);
\fill[white,rounded corners=.6pt] (10.360,3.210) rectangle +(.16,.04);
\fill[white,rounded corners=.6pt] (10.750,2.460) rectangle +(.16,.04);
\fill[white,rounded corners=.6pt] (10.750,3.210) rectangle +(.16,.04);
\fill[white,rounded corners=.6pt] (11.140,2.460) rectangle +(.16,.04);
\fill[white,rounded corners=.6pt] (11.140,3.210) rectangle +(.16,.04);
\fill[white,rounded corners=.6pt] (11.530,2.460) rectangle +(.16,.04);
\fill[white,rounded corners=.6pt] (11.530,3.210) rectangle +(.16,.04);
\fill[white,rounded corners=.6pt] (11.920,2.460) rectangle +(.16,.04);
\fill[white,rounded corners=.6pt] (11.920,3.210) rectangle +(.16,.04);
\fill[white,rounded corners=.6pt] (12.310,2.460) rectangle +(.16,.04);
\fill[white,rounded corners=.6pt] (12.310,3.210) rectangle +(.16,.04);
\fill[white,rounded corners=.6pt] (12.700,2.460) rectangle +(.16,.04);
\fill[white,rounded corners=.6pt] (12.700,3.210) rectangle +(.16,.04);
\fill[white,rounded corners=.6pt] (13.090,2.460) rectangle +(.16,.04);
\fill[white,rounded corners=.6pt] (13.090,3.210) rectangle +(.16,.04);
\fill[white,rounded corners=.6pt] (13.480,2.460) rectangle +(.16,.04);
\fill[white,rounded corners=.6pt] (13.480,3.210) rectangle +(.16,.04);
\end{scope}
\draw[rqTeal!65,line width=.7pt,line join=round] (4.2590,2.42) -- (12.7777,2.42) -- (12.7777,3.29) -- (13.97,3.89) -- (13.97,8.19) -- (0,8.19) -- (0,3.89) -- (4.2590,3.29) -- cycle;
\node[text=rqTeal,font=\bfseries] at (6.985,4.12) {Selected scene: video frames 13--37};
\fill[white,rounded corners=.7pt] (0.220,4.300) rectangle +(.16,.05);
\fill[white,rounded corners=.7pt] (0.220,7.780) rectangle +(.16,.05);
\fill[white,rounded corners=.7pt] (0.610,4.300) rectangle +(.16,.05);
\fill[white,rounded corners=.7pt] (0.610,7.780) rectangle +(.16,.05);
\fill[white,rounded corners=.7pt] (1.000,4.300) rectangle +(.16,.05);
\fill[white,rounded corners=.7pt] (1.000,7.780) rectangle +(.16,.05);
\fill[white,rounded corners=.7pt] (1.390,4.300) rectangle +(.16,.05);
\fill[white,rounded corners=.7pt] (1.390,7.780) rectangle +(.16,.05);
\fill[white,rounded corners=.7pt] (1.780,4.300) rectangle +(.16,.05);
\fill[white,rounded corners=.7pt] (1.780,7.780) rectangle +(.16,.05);
\fill[white,rounded corners=.7pt] (2.170,4.300) rectangle +(.16,.05);
\fill[white,rounded corners=.7pt] (2.170,7.780) rectangle +(.16,.05);
\fill[white,rounded corners=.7pt] (2.560,4.300) rectangle +(.16,.05);
\fill[white,rounded corners=.7pt] (2.560,7.780) rectangle +(.16,.05);
\fill[white,rounded corners=.7pt] (2.950,4.300) rectangle +(.16,.05);
\fill[white,rounded corners=.7pt] (2.950,7.780) rectangle +(.16,.05);
\fill[white,rounded corners=.7pt] (3.340,4.300) rectangle +(.16,.05);
\fill[white,rounded corners=.7pt] (3.340,7.780) rectangle +(.16,.05);
\fill[white,rounded corners=.7pt] (3.730,4.300) rectangle +(.16,.05);
\fill[white,rounded corners=.7pt] (3.730,7.780) rectangle +(.16,.05);
\fill[white,rounded corners=.7pt] (4.120,4.300) rectangle +(.16,.05);
\fill[white,rounded corners=.7pt] (4.120,7.780) rectangle +(.16,.05);
\fill[white,rounded corners=.7pt] (4.510,4.300) rectangle +(.16,.05);
\fill[white,rounded corners=.7pt] (4.510,7.780) rectangle +(.16,.05);
\fill[white,rounded corners=.7pt] (4.900,4.300) rectangle +(.16,.05);
\fill[white,rounded corners=.7pt] (4.900,7.780) rectangle +(.16,.05);
\fill[white,rounded corners=.7pt] (5.290,4.300) rectangle +(.16,.05);
\fill[white,rounded corners=.7pt] (5.290,7.780) rectangle +(.16,.05);
\fill[white,rounded corners=.7pt] (5.680,4.300) rectangle +(.16,.05);
\fill[white,rounded corners=.7pt] (5.680,7.780) rectangle +(.16,.05);
\fill[white,rounded corners=.7pt] (6.070,4.300) rectangle +(.16,.05);
\fill[white,rounded corners=.7pt] (6.070,7.780) rectangle +(.16,.05);
\fill[white,rounded corners=.7pt] (6.460,4.300) rectangle +(.16,.05);
\fill[white,rounded corners=.7pt] (6.460,7.780) rectangle +(.16,.05);
\fill[white,rounded corners=.7pt] (6.850,4.300) rectangle +(.16,.05);
\fill[white,rounded corners=.7pt] (6.850,7.780) rectangle +(.16,.05);
\fill[white,rounded corners=.7pt] (7.240,4.300) rectangle +(.16,.05);
\fill[white,rounded corners=.7pt] (7.240,7.780) rectangle +(.16,.05);
\fill[white,rounded corners=.7pt] (7.630,4.300) rectangle +(.16,.05);
\fill[white,rounded corners=.7pt] (7.630,7.780) rectangle +(.16,.05);
\fill[white,rounded corners=.7pt] (8.020,4.300) rectangle +(.16,.05);
\fill[white,rounded corners=.7pt] (8.020,7.780) rectangle +(.16,.05);
\fill[white,rounded corners=.7pt] (8.410,4.300) rectangle +(.16,.05);
\fill[white,rounded corners=.7pt] (8.410,7.780) rectangle +(.16,.05);
\fill[white,rounded corners=.7pt] (8.800,4.300) rectangle +(.16,.05);
\fill[white,rounded corners=.7pt] (8.800,7.780) rectangle +(.16,.05);
\fill[white,rounded corners=.7pt] (9.190,4.300) rectangle +(.16,.05);
\fill[white,rounded corners=.7pt] (9.190,7.780) rectangle +(.16,.05);
\fill[white,rounded corners=.7pt] (9.580,4.300) rectangle +(.16,.05);
\fill[white,rounded corners=.7pt] (9.580,7.780) rectangle +(.16,.05);
\fill[white,rounded corners=.7pt] (9.970,4.300) rectangle +(.16,.05);
\fill[white,rounded corners=.7pt] (9.970,7.780) rectangle +(.16,.05);
\fill[white,rounded corners=.7pt] (10.360,4.300) rectangle +(.16,.05);
\fill[white,rounded corners=.7pt] (10.360,7.780) rectangle +(.16,.05);
\fill[white,rounded corners=.7pt] (10.750,4.300) rectangle +(.16,.05);
\fill[white,rounded corners=.7pt] (10.750,7.780) rectangle +(.16,.05);
\fill[white,rounded corners=.7pt] (11.140,4.300) rectangle +(.16,.05);
\fill[white,rounded corners=.7pt] (11.140,7.780) rectangle +(.16,.05);
\fill[white,rounded corners=.7pt] (11.530,4.300) rectangle +(.16,.05);
\fill[white,rounded corners=.7pt] (11.530,7.780) rectangle +(.16,.05);
\fill[white,rounded corners=.7pt] (11.920,4.300) rectangle +(.16,.05);
\fill[white,rounded corners=.7pt] (11.920,7.780) rectangle +(.16,.05);
\fill[white,rounded corners=.7pt] (12.310,4.300) rectangle +(.16,.05);
\fill[white,rounded corners=.7pt] (12.310,7.780) rectangle +(.16,.05);
\fill[white,rounded corners=.7pt] (12.700,4.300) rectangle +(.16,.05);
\fill[white,rounded corners=.7pt] (12.700,7.780) rectangle +(.16,.05);
\fill[white,rounded corners=.7pt] (13.090,4.300) rectangle +(.16,.05);
\fill[white,rounded corners=.7pt] (13.090,7.780) rectangle +(.16,.05);
\fill[white,rounded corners=.7pt] (13.480,4.300) rectangle +(.16,.05);
\fill[white,rounded corners=.7pt] (13.480,7.780) rectangle +(.16,.05);
\node[text=rqTeal!65] at (4.650,6.06) {$\cdots$};
\node[text=rqTeal!65] at (9.320,6.06) {$\cdots$};
\node[anchor=north west] at (0.130,4.420) {\includegraphics[width=4.37cm]{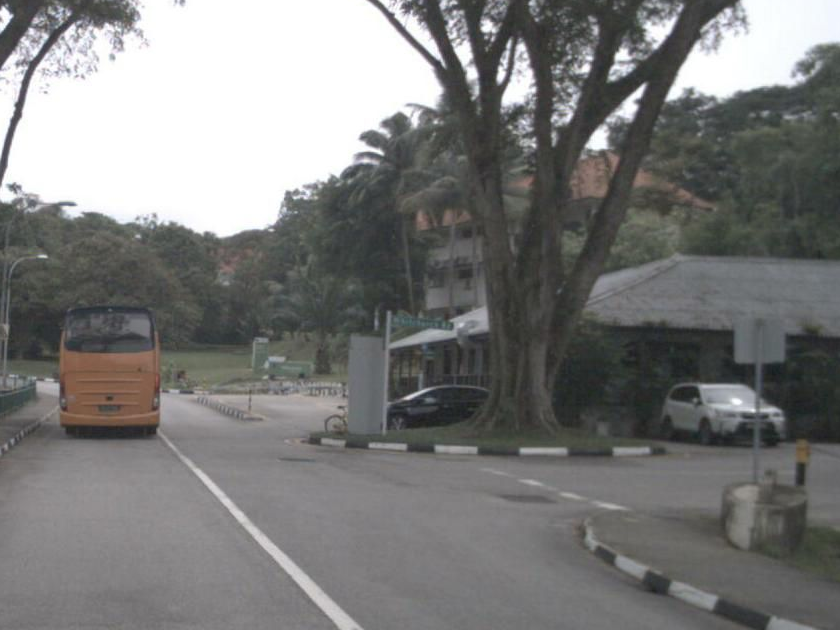}};
\node[text=rqInk] at (2.315,8.00) {Frame 21};
\draw[rqTeal,line width=.75pt,rounded corners=1pt] (0.4305,6.0029) rectangle (0.9716,6.6934);
\node[anchor=north west,fill=rqTeal,text=white,inner sep=1.3pt,rounded corners=1pt,font=\normalfont\fontsize{9}{10}\selectfont] at (0.4305,5.5629) {bus 0.81};
\draw[rqBlue,line width=.75pt,rounded corners=1pt] (3.5558,6.4051) rectangle (4.2242,6.7396);
\node[anchor=north west,fill=rqBlue,text=white,inner sep=1.3pt,rounded corners=1pt,font=\normalfont\fontsize{9}{10}\selectfont] at (3.2550,5.9651) {car 0.54};
\node[anchor=north west] at (4.800,4.420) {\includegraphics[width=4.37cm]{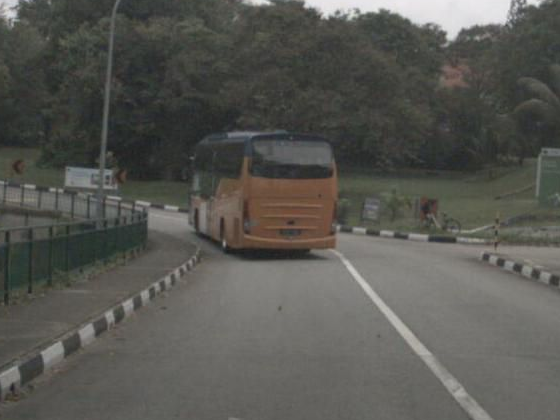}};
\node[text=rqInk] at (6.985,8.00) {Frame 31};
\draw[rqTeal,line width=.75pt,rounded corners=1pt] (6.2642,5.4307) rectangle (7.4492,6.4111);
\node[anchor=north west,fill=rqTeal,text=white,inner sep=1.3pt,rounded corners=1pt,font=\normalfont\fontsize{9}{10}\selectfont] at (6.2642,4.9907) {bus 0.83};
\node[anchor=north west] at (9.470,4.420) {\includegraphics[width=4.37cm]{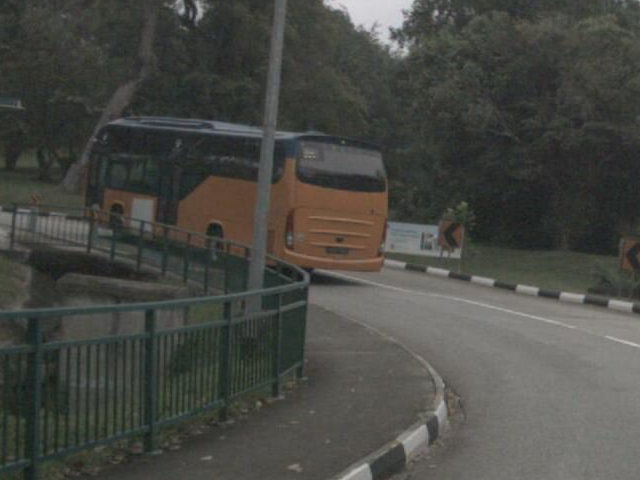}};
\node[text=rqInk] at (11.655,8.00) {Frame 37};
\draw[rqTeal,line width=.75pt,rounded corners=1pt] (10.0391,5.2006) rectangle (12.1344,6.3227);
\node[anchor=north west,fill=rqTeal,text=white,inner sep=1.3pt,rounded corners=1pt,font=\normalfont\fontsize{9}{10}\selectfont] at (10.0391,4.7606) {bus 0.81};
\draw[black!20,line width=.4pt] (0,8.35) -- (13.97,8.35);
\node[anchor=west] at (0,8.61) {Video rank $\downarrow$};
\node[anchor=west,text=black!65] at (0,9.01) {of 746};
\fill[rqTeal!8] (2.57,8.41) rectangle (4.62,9.13);
\node[font=\bfseries,text=rqTeal] at (3.600,8.61) {LogSTOP};
\node[font=\bfseries,text=rqTeal] at (3.600,9.01) {2};
\node[] at (6.680,8.61) {Spatial robustness};
\node[] at (6.680,9.01) {2};
\node[] at (10.000,8.61) {CaptionSim};
\node[] at (10.000,9.01) {150};
\node[] at (12.830,8.61) {mPLUG};
\node[] at (12.830,9.01) {349};
\end{tikzpicture}
\endgroup

%% file: figures/retrieval_selected_scenes/stand_clap.tex
\begingroup
\input{figures/retrieval_qualitative_style}
\begin{tikzpicture}[x=1cm,y=-1cm,font=\normalfont\normalsize,every node/.style={inner sep=0pt,outer sep=0pt},line cap=round]
\path[use as bounding box] (0,0) rectangle (13.97,9.15);
\filldraw[fill=rqLight,draw=rqLine,line width=.55pt,rounded corners=6pt] (0,.42) rectangle (13.97,1.80);
\node[text=rqInk] at (6.985,.70) {Retrieve videos with a 50-frame scene where:};
\node[text=rqInk,font=\bfseries,align=center,text width=13.30cm] at (6.985,1.10) {Standing holds until hand clapping starts and continues.};
\node[text=rqInk] at (6.985,1.54) {$\varphi = \mathsf{stand}\;\mathcal{U}\;(\square\,\mathsf{hand\ clap})$};
\filldraw[fill=rqLight,draw=rqLine,line width=.55pt,rounded corners=5pt] (0,1.97) rectangle (13.97,3.40);
\node[anchor=west,text=rqInk] at (.15,2.18) {Full video: 300 frames};
\node[anchor=east,text=black!60] at (13.82,2.18) {time $\longrightarrow$};
\fill[black!18] (.10,2.42) rectangle (13.87,3.29);
\node[anchor=north west,opacity=.27] at (0.1700,2.54) {\includegraphics[width=1.1108cm]{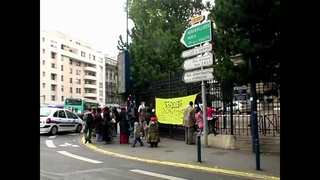}};
\node[anchor=north west,opacity=.27] at (1.3058,2.54) {\includegraphics[width=1.1108cm]{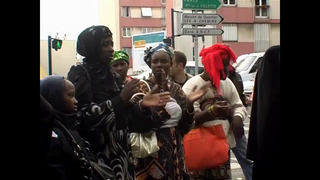}};
\node[anchor=north west,opacity=.27] at (2.4417,2.54) {\includegraphics[width=1.1108cm]{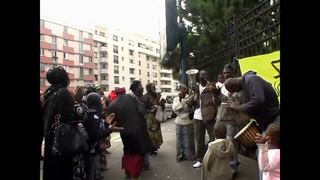}};
\node[anchor=north west,opacity=.27] at (3.5775,2.54) {\includegraphics[width=1.1108cm]{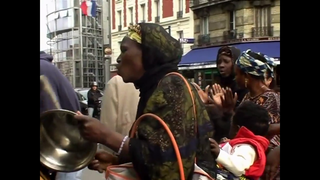}};
\node[anchor=north west,opacity=.27] at (4.7133,2.54) {\includegraphics[width=1.1108cm]{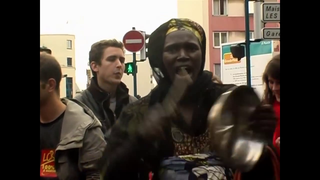}};
\node[anchor=north west,opacity=.27] at (5.8492,2.54) {\includegraphics[width=1.1108cm]{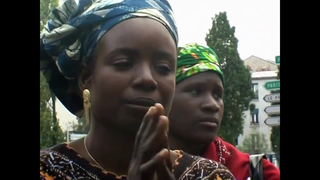}};
\node[anchor=north west,opacity=.27] at (6.9850,2.54) {\includegraphics[width=1.1108cm]{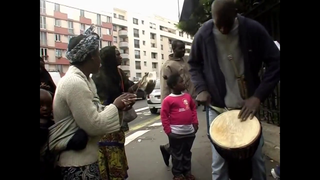}};
\node[anchor=north west,opacity=.27] at (8.1208,2.54) {\includegraphics[width=1.1108cm]{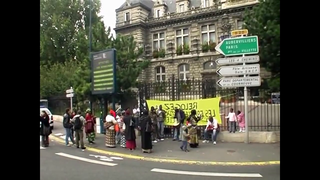}};
\node[anchor=north west,opacity=.27] at (9.2567,2.54) {\includegraphics[width=1.1108cm]{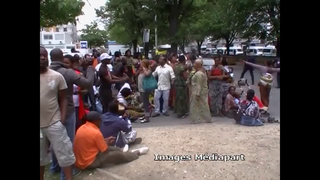}};
\node[anchor=north west,opacity=.27] at (10.3925,2.54) {\includegraphics[width=1.1108cm]{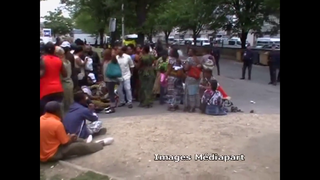}};
\node[anchor=north west,opacity=.27] at (11.5283,2.54) {\includegraphics[width=1.1108cm]{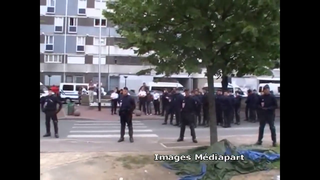}};
\node[anchor=north west,opacity=.27] at (12.6642,2.54) {\includegraphics[width=1.1108cm]{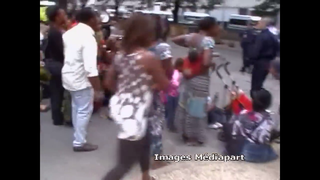}};
\fill[white,rounded corners=.6pt] (0.220,2.460) rectangle +(.16,.04);
\fill[white,rounded corners=.6pt] (0.220,3.210) rectangle +(.16,.04);
\fill[white,rounded corners=.6pt] (0.610,2.460) rectangle +(.16,.04);
\fill[white,rounded corners=.6pt] (0.610,3.210) rectangle +(.16,.04);
\fill[white,rounded corners=.6pt] (1.000,2.460) rectangle +(.16,.04);
\fill[white,rounded corners=.6pt] (1.000,3.210) rectangle +(.16,.04);
\fill[white,rounded corners=.6pt] (1.390,2.460) rectangle +(.16,.04);
\fill[white,rounded corners=.6pt] (1.390,3.210) rectangle +(.16,.04);
\fill[white,rounded corners=.6pt] (1.780,2.460) rectangle +(.16,.04);
\fill[white,rounded corners=.6pt] (1.780,3.210) rectangle +(.16,.04);
\fill[white,rounded corners=.6pt] (2.170,2.460) rectangle +(.16,.04);
\fill[white,rounded corners=.6pt] (2.170,3.210) rectangle +(.16,.04);
\fill[white,rounded corners=.6pt] (2.560,2.460) rectangle +(.16,.04);
\fill[white,rounded corners=.6pt] (2.560,3.210) rectangle +(.16,.04);
\fill[white,rounded corners=.6pt] (2.950,2.460) rectangle +(.16,.04);
\fill[white,rounded corners=.6pt] (2.950,3.210) rectangle +(.16,.04);
\fill[white,rounded corners=.6pt] (3.340,2.460) rectangle +(.16,.04);
\fill[white,rounded corners=.6pt] (3.340,3.210) rectangle +(.16,.04);
\fill[white,rounded corners=.6pt] (3.730,2.460) rectangle +(.16,.04);
\fill[white,rounded corners=.6pt] (3.730,3.210) rectangle +(.16,.04);
\fill[white,rounded corners=.6pt] (4.120,2.460) rectangle +(.16,.04);
\fill[white,rounded corners=.6pt] (4.120,3.210) rectangle +(.16,.04);
\fill[white,rounded corners=.6pt] (4.510,2.460) rectangle +(.16,.04);
\fill[white,rounded corners=.6pt] (4.510,3.210) rectangle +(.16,.04);
\fill[white,rounded corners=.6pt] (4.900,2.460) rectangle +(.16,.04);
\fill[white,rounded corners=.6pt] (4.900,3.210) rectangle +(.16,.04);
\fill[white,rounded corners=.6pt] (5.290,2.460) rectangle +(.16,.04);
\fill[white,rounded corners=.6pt] (5.290,3.210) rectangle +(.16,.04);
\fill[white,rounded corners=.6pt] (5.680,2.460) rectangle +(.16,.04);
\fill[white,rounded corners=.6pt] (5.680,3.210) rectangle +(.16,.04);
\fill[white,rounded corners=.6pt] (6.070,2.460) rectangle +(.16,.04);
\fill[white,rounded corners=.6pt] (6.070,3.210) rectangle +(.16,.04);
\fill[white,rounded corners=.6pt] (6.460,2.460) rectangle +(.16,.04);
\fill[white,rounded corners=.6pt] (6.460,3.210) rectangle +(.16,.04);
\fill[white,rounded corners=.6pt] (6.850,2.460) rectangle +(.16,.04);
\fill[white,rounded corners=.6pt] (6.850,3.210) rectangle +(.16,.04);
\fill[white,rounded corners=.6pt] (7.240,2.460) rectangle +(.16,.04);
\fill[white,rounded corners=.6pt] (7.240,3.210) rectangle +(.16,.04);
\fill[white,rounded corners=.6pt] (7.630,2.460) rectangle +(.16,.04);
\fill[white,rounded corners=.6pt] (7.630,3.210) rectangle +(.16,.04);
\fill[white,rounded corners=.6pt] (8.020,2.460) rectangle +(.16,.04);
\fill[white,rounded corners=.6pt] (8.020,3.210) rectangle +(.16,.04);
\fill[white,rounded corners=.6pt] (8.410,2.460) rectangle +(.16,.04);
\fill[white,rounded corners=.6pt] (8.410,3.210) rectangle +(.16,.04);
\fill[white,rounded corners=.6pt] (8.800,2.460) rectangle +(.16,.04);
\fill[white,rounded corners=.6pt] (8.800,3.210) rectangle +(.16,.04);
\fill[white,rounded corners=.6pt] (9.190,2.460) rectangle +(.16,.04);
\fill[white,rounded corners=.6pt] (9.190,3.210) rectangle +(.16,.04);
\fill[white,rounded corners=.6pt] (9.580,2.460) rectangle +(.16,.04);
\fill[white,rounded corners=.6pt] (9.580,3.210) rectangle +(.16,.04);
\fill[white,rounded corners=.6pt] (9.970,2.460) rectangle +(.16,.04);
\fill[white,rounded corners=.6pt] (9.970,3.210) rectangle +(.16,.04);
\fill[white,rounded corners=.6pt] (10.360,2.460) rectangle +(.16,.04);
\fill[white,rounded corners=.6pt] (10.360,3.210) rectangle +(.16,.04);
\fill[white,rounded corners=.6pt] (10.750,2.460) rectangle +(.16,.04);
\fill[white,rounded corners=.6pt] (10.750,3.210) rectangle +(.16,.04);
\fill[white,rounded corners=.6pt] (11.140,2.460) rectangle +(.16,.04);
\fill[white,rounded corners=.6pt] (11.140,3.210) rectangle +(.16,.04);
\fill[white,rounded corners=.6pt] (11.530,2.460) rectangle +(.16,.04);
\fill[white,rounded corners=.6pt] (11.530,3.210) rectangle +(.16,.04);
\fill[white,rounded corners=.6pt] (11.920,2.460) rectangle +(.16,.04);
\fill[white,rounded corners=.6pt] (11.920,3.210) rectangle +(.16,.04);
\fill[white,rounded corners=.6pt] (12.310,2.460) rectangle +(.16,.04);
\fill[white,rounded corners=.6pt] (12.310,3.210) rectangle +(.16,.04);
\fill[white,rounded corners=.6pt] (12.700,2.460) rectangle +(.16,.04);
\fill[white,rounded corners=.6pt] (12.700,3.210) rectangle +(.16,.04);
\fill[white,rounded corners=.6pt] (13.090,2.460) rectangle +(.16,.04);
\fill[white,rounded corners=.6pt] (13.090,3.210) rectangle +(.16,.04);
\fill[white,rounded corners=.6pt] (13.480,2.460) rectangle +(.16,.04);
\fill[white,rounded corners=.6pt] (13.480,3.210) rectangle +(.16,.04);
\fill[rqGreenStrip] (1.0787,2.42) -- (3.3503,2.42) -- (3.3503,3.29) -- (13.97,3.89) -- (13.97,8.19) -- (0,8.19) -- (0,3.89) -- (1.0787,3.29) -- cycle;
\begin{scope}\clip (1.0787,2.42) rectangle (3.3503,3.29);
\node[anchor=north west] at (0.1700,2.54) {\includegraphics[width=1.1108cm]{figures/retrieval_selected_scenes/overviews/street_actions/frame_012.png}};
\node[anchor=north west] at (1.3058,2.54) {\includegraphics[width=1.1108cm]{figures/retrieval_selected_scenes/overviews/street_actions/frame_037.png}};
\node[anchor=north west] at (2.4417,2.54) {\includegraphics[width=1.1108cm]{figures/retrieval_selected_scenes/overviews/street_actions/frame_062.png}};
\node[anchor=north west] at (3.5775,2.54) {\includegraphics[width=1.1108cm]{figures/retrieval_selected_scenes/overviews/street_actions/frame_087.png}};
\node[anchor=north west] at (4.7133,2.54) {\includegraphics[width=1.1108cm]{figures/retrieval_selected_scenes/overviews/street_actions/frame_112.png}};
\node[anchor=north west] at (5.8492,2.54) {\includegraphics[width=1.1108cm]{figures/retrieval_selected_scenes/overviews/street_actions/frame_137.png}};
\node[anchor=north west] at (6.9850,2.54) {\includegraphics[width=1.1108cm]{figures/retrieval_selected_scenes/overviews/street_actions/frame_162.png}};
\node[anchor=north west] at (8.1208,2.54) {\includegraphics[width=1.1108cm]{figures/retrieval_selected_scenes/overviews/street_actions/frame_187.png}};
\node[anchor=north west] at (9.2567,2.54) {\includegraphics[width=1.1108cm]{figures/retrieval_selected_scenes/overviews/street_actions/frame_212.png}};
\node[anchor=north west] at (10.3925,2.54) {\includegraphics[width=1.1108cm]{figures/retrieval_selected_scenes/overviews/street_actions/frame_237.png}};
\node[anchor=north west] at (11.5283,2.54) {\includegraphics[width=1.1108cm]{figures/retrieval_selected_scenes/overviews/street_actions/frame_262.png}};
\node[anchor=north west] at (12.6642,2.54) {\includegraphics[width=1.1108cm]{figures/retrieval_selected_scenes/overviews/street_actions/frame_287.png}};
\fill[white,rounded corners=.6pt] (0.220,2.460) rectangle +(.16,.04);
\fill[white,rounded corners=.6pt] (0.220,3.210) rectangle +(.16,.04);
\fill[white,rounded corners=.6pt] (0.610,2.460) rectangle +(.16,.04);
\fill[white,rounded corners=.6pt] (0.610,3.210) rectangle +(.16,.04);
\fill[white,rounded corners=.6pt] (1.000,2.460) rectangle +(.16,.04);
\fill[white,rounded corners=.6pt] (1.000,3.210) rectangle +(.16,.04);
\fill[white,rounded corners=.6pt] (1.390,2.460) rectangle +(.16,.04);
\fill[white,rounded corners=.6pt] (1.390,3.210) rectangle +(.16,.04);
\fill[white,rounded corners=.6pt] (1.780,2.460) rectangle +(.16,.04);
\fill[white,rounded corners=.6pt] (1.780,3.210) rectangle +(.16,.04);
\fill[white,rounded corners=.6pt] (2.170,2.460) rectangle +(.16,.04);
\fill[white,rounded corners=.6pt] (2.170,3.210) rectangle +(.16,.04);
\fill[white,rounded corners=.6pt] (2.560,2.460) rectangle +(.16,.04);
\fill[white,rounded corners=.6pt] (2.560,3.210) rectangle +(.16,.04);
\fill[white,rounded corners=.6pt] (2.950,2.460) rectangle +(.16,.04);
\fill[white,rounded corners=.6pt] (2.950,3.210) rectangle +(.16,.04);
\fill[white,rounded corners=.6pt] (3.340,2.460) rectangle +(.16,.04);
\fill[white,rounded corners=.6pt] (3.340,3.210) rectangle +(.16,.04);
\fill[white,rounded corners=.6pt] (3.730,2.460) rectangle +(.16,.04);
\fill[white,rounded corners=.6pt] (3.730,3.210) rectangle +(.16,.04);
\fill[white,rounded corners=.6pt] (4.120,2.460) rectangle +(.16,.04);
\fill[white,rounded corners=.6pt] (4.120,3.210) rectangle +(.16,.04);
\fill[white,rounded corners=.6pt] (4.510,2.460) rectangle +(.16,.04);
\fill[white,rounded corners=.6pt] (4.510,3.210) rectangle +(.16,.04);
\fill[white,rounded corners=.6pt] (4.900,2.460) rectangle +(.16,.04);
\fill[white,rounded corners=.6pt] (4.900,3.210) rectangle +(.16,.04);
\fill[white,rounded corners=.6pt] (5.290,2.460) rectangle +(.16,.04);
\fill[white,rounded corners=.6pt] (5.290,3.210) rectangle +(.16,.04);
\fill[white,rounded corners=.6pt] (5.680,2.460) rectangle +(.16,.04);
\fill[white,rounded corners=.6pt] (5.680,3.210) rectangle +(.16,.04);
\fill[white,rounded corners=.6pt] (6.070,2.460) rectangle +(.16,.04);
\fill[white,rounded corners=.6pt] (6.070,3.210) rectangle +(.16,.04);
\fill[white,rounded corners=.6pt] (6.460,2.460) rectangle +(.16,.04);
\fill[white,rounded corners=.6pt] (6.460,3.210) rectangle +(.16,.04);
\fill[white,rounded corners=.6pt] (6.850,2.460) rectangle +(.16,.04);
\fill[white,rounded corners=.6pt] (6.850,3.210) rectangle +(.16,.04);
\fill[white,rounded corners=.6pt] (7.240,2.460) rectangle +(.16,.04);
\fill[white,rounded corners=.6pt] (7.240,3.210) rectangle +(.16,.04);
\fill[white,rounded corners=.6pt] (7.630,2.460) rectangle +(.16,.04);
\fill[white,rounded corners=.6pt] (7.630,3.210) rectangle +(.16,.04);
\fill[white,rounded corners=.6pt] (8.020,2.460) rectangle +(.16,.04);
\fill[white,rounded corners=.6pt] (8.020,3.210) rectangle +(.16,.04);
\fill[white,rounded corners=.6pt] (8.410,2.460) rectangle +(.16,.04);
\fill[white,rounded corners=.6pt] (8.410,3.210) rectangle +(.16,.04);
\fill[white,rounded corners=.6pt] (8.800,2.460) rectangle +(.16,.04);
\fill[white,rounded corners=.6pt] (8.800,3.210) rectangle +(.16,.04);
\fill[white,rounded corners=.6pt] (9.190,2.460) rectangle +(.16,.04);
\fill[white,rounded corners=.6pt] (9.190,3.210) rectangle +(.16,.04);
\fill[white,rounded corners=.6pt] (9.580,2.460) rectangle +(.16,.04);
\fill[white,rounded corners=.6pt] (9.580,3.210) rectangle +(.16,.04);
\fill[white,rounded corners=.6pt] (9.970,2.460) rectangle +(.16,.04);
\fill[white,rounded corners=.6pt] (9.970,3.210) rectangle +(.16,.04);
\fill[white,rounded corners=.6pt] (10.360,2.460) rectangle +(.16,.04);
\fill[white,rounded corners=.6pt] (10.360,3.210) rectangle +(.16,.04);
\fill[white,rounded corners=.6pt] (10.750,2.460) rectangle +(.16,.04);
\fill[white,rounded corners=.6pt] (10.750,3.210) rectangle +(.16,.04);
\fill[white,rounded corners=.6pt] (11.140,2.460) rectangle +(.16,.04);
\fill[white,rounded corners=.6pt] (11.140,3.210) rectangle +(.16,.04);
\fill[white,rounded corners=.6pt] (11.530,2.460) rectangle +(.16,.04);
\fill[white,rounded corners=.6pt] (11.530,3.210) rectangle +(.16,.04);
\fill[white,rounded corners=.6pt] (11.920,2.460) rectangle +(.16,.04);
\fill[white,rounded corners=.6pt] (11.920,3.210) rectangle +(.16,.04);
\fill[white,rounded corners=.6pt] (12.310,2.460) rectangle +(.16,.04);
\fill[white,rounded corners=.6pt] (12.310,3.210) rectangle +(.16,.04);
\fill[white,rounded corners=.6pt] (12.700,2.460) rectangle +(.16,.04);
\fill[white,rounded corners=.6pt] (12.700,3.210) rectangle +(.16,.04);
\fill[white,rounded corners=.6pt] (13.090,2.460) rectangle +(.16,.04);
\fill[white,rounded corners=.6pt] (13.090,3.210) rectangle +(.16,.04);
\fill[white,rounded corners=.6pt] (13.480,2.460) rectangle +(.16,.04);
\fill[white,rounded corners=.6pt] (13.480,3.210) rectangle +(.16,.04);
\end{scope}
\draw[rqTeal!65,line width=.7pt,line join=round] (1.0787,2.42) -- (3.3503,2.42) -- (3.3503,3.29) -- (13.97,3.89) -- (13.97,8.19) -- (0,8.19) -- (0,3.89) -- (1.0787,3.29) -- cycle;
\node[text=rqTeal,font=\bfseries] at (6.985,4.12) {Selected scene: video frames 21--70};
\fill[white,rounded corners=.7pt] (0.220,4.300) rectangle +(.16,.05);
\fill[white,rounded corners=.7pt] (0.220,7.780) rectangle +(.16,.05);
\fill[white,rounded corners=.7pt] (0.610,4.300) rectangle +(.16,.05);
\fill[white,rounded corners=.7pt] (0.610,7.780) rectangle +(.16,.05);
\fill[white,rounded corners=.7pt] (1.000,4.300) rectangle +(.16,.05);
\fill[white,rounded corners=.7pt] (1.000,7.780) rectangle +(.16,.05);
\fill[white,rounded corners=.7pt] (1.390,4.300) rectangle +(.16,.05);
\fill[white,rounded corners=.7pt] (1.390,7.780) rectangle +(.16,.05);
\fill[white,rounded corners=.7pt] (1.780,4.300) rectangle +(.16,.05);
\fill[white,rounded corners=.7pt] (1.780,7.780) rectangle +(.16,.05);
\fill[white,rounded corners=.7pt] (2.170,4.300) rectangle +(.16,.05);
\fill[white,rounded corners=.7pt] (2.170,7.780) rectangle +(.16,.05);
\fill[white,rounded corners=.7pt] (2.560,4.300) rectangle +(.16,.05);
\fill[white,rounded corners=.7pt] (2.560,7.780) rectangle +(.16,.05);
\fill[white,rounded corners=.7pt] (2.950,4.300) rectangle +(.16,.05);
\fill[white,rounded corners=.7pt] (2.950,7.780) rectangle +(.16,.05);
\fill[white,rounded corners=.7pt] (3.340,4.300) rectangle +(.16,.05);
\fill[white,rounded corners=.7pt] (3.340,7.780) rectangle +(.16,.05);
\fill[white,rounded corners=.7pt] (3.730,4.300) rectangle +(.16,.05);
\fill[white,rounded corners=.7pt] (3.730,7.780) rectangle +(.16,.05);
\fill[white,rounded corners=.7pt] (4.120,4.300) rectangle +(.16,.05);
\fill[white,rounded corners=.7pt] (4.120,7.780) rectangle +(.16,.05);
\fill[white,rounded corners=.7pt] (4.510,4.300) rectangle +(.16,.05);
\fill[white,rounded corners=.7pt] (4.510,7.780) rectangle +(.16,.05);
\fill[white,rounded corners=.7pt] (4.900,4.300) rectangle +(.16,.05);
\fill[white,rounded corners=.7pt] (4.900,7.780) rectangle +(.16,.05);
\fill[white,rounded corners=.7pt] (5.290,4.300) rectangle +(.16,.05);
\fill[white,rounded corners=.7pt] (5.290,7.780) rectangle +(.16,.05);
\fill[white,rounded corners=.7pt] (5.680,4.300) rectangle +(.16,.05);
\fill[white,rounded corners=.7pt] (5.680,7.780) rectangle +(.16,.05);
\fill[white,rounded corners=.7pt] (6.070,4.300) rectangle +(.16,.05);
\fill[white,rounded corners=.7pt] (6.070,7.780) rectangle +(.16,.05);
\fill[white,rounded corners=.7pt] (6.460,4.300) rectangle +(.16,.05);
\fill[white,rounded corners=.7pt] (6.460,7.780) rectangle +(.16,.05);
\fill[white,rounded corners=.7pt] (6.850,4.300) rectangle +(.16,.05);
\fill[white,rounded corners=.7pt] (6.850,7.780) rectangle +(.16,.05);
\fill[white,rounded corners=.7pt] (7.240,4.300) rectangle +(.16,.05);
\fill[white,rounded corners=.7pt] (7.240,7.780) rectangle +(.16,.05);
\fill[white,rounded corners=.7pt] (7.630,4.300) rectangle +(.16,.05);
\fill[white,rounded corners=.7pt] (7.630,7.780) rectangle +(.16,.05);
\fill[white,rounded corners=.7pt] (8.020,4.300) rectangle +(.16,.05);
\fill[white,rounded corners=.7pt] (8.020,7.780) rectangle +(.16,.05);
\fill[white,rounded corners=.7pt] (8.410,4.300) rectangle +(.16,.05);
\fill[white,rounded corners=.7pt] (8.410,7.780) rectangle +(.16,.05);
\fill[white,rounded corners=.7pt] (8.800,4.300) rectangle +(.16,.05);
\fill[white,rounded corners=.7pt] (8.800,7.780) rectangle +(.16,.05);
\fill[white,rounded corners=.7pt] (9.190,4.300) rectangle +(.16,.05);
\fill[white,rounded corners=.7pt] (9.190,7.780) rectangle +(.16,.05);
\fill[white,rounded corners=.7pt] (9.580,4.300) rectangle +(.16,.05);
\fill[white,rounded corners=.7pt] (9.580,7.780) rectangle +(.16,.05);
\fill[white,rounded corners=.7pt] (9.970,4.300) rectangle +(.16,.05);
\fill[white,rounded corners=.7pt] (9.970,7.780) rectangle +(.16,.05);
\fill[white,rounded corners=.7pt] (10.360,4.300) rectangle +(.16,.05);
\fill[white,rounded corners=.7pt] (10.360,7.780) rectangle +(.16,.05);
\fill[white,rounded corners=.7pt] (10.750,4.300) rectangle +(.16,.05);
\fill[white,rounded corners=.7pt] (10.750,7.780) rectangle +(.16,.05);
\fill[white,rounded corners=.7pt] (11.140,4.300) rectangle +(.16,.05);
\fill[white,rounded corners=.7pt] (11.140,7.780) rectangle +(.16,.05);
\fill[white,rounded corners=.7pt] (11.530,4.300) rectangle +(.16,.05);
\fill[white,rounded corners=.7pt] (11.530,7.780) rectangle +(.16,.05);
\fill[white,rounded corners=.7pt] (11.920,4.300) rectangle +(.16,.05);
\fill[white,rounded corners=.7pt] (11.920,7.780) rectangle +(.16,.05);
\fill[white,rounded corners=.7pt] (12.310,4.300) rectangle +(.16,.05);
\fill[white,rounded corners=.7pt] (12.310,7.780) rectangle +(.16,.05);
\fill[white,rounded corners=.7pt] (12.700,4.300) rectangle +(.16,.05);
\fill[white,rounded corners=.7pt] (12.700,7.780) rectangle +(.16,.05);
\fill[white,rounded corners=.7pt] (13.090,4.300) rectangle +(.16,.05);
\fill[white,rounded corners=.7pt] (13.090,7.780) rectangle +(.16,.05);
\fill[white,rounded corners=.7pt] (13.480,4.300) rectangle +(.16,.05);
\fill[white,rounded corners=.7pt] (13.480,7.780) rectangle +(.16,.05);
\node[text=rqTeal!65] at (4.650,6.06) {$\cdots$};
\node[text=rqTeal!65] at (9.320,6.06) {$\cdots$};
\node[anchor=north west] at (0.130,4.420) {\includegraphics[width=4.37cm]{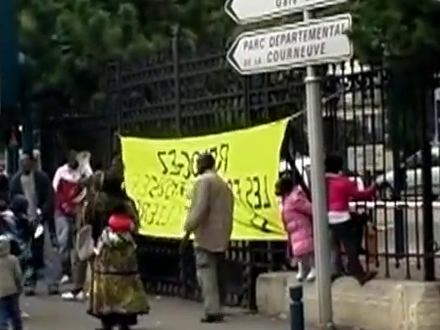}};
\node[text=rqInk] at (2.315,8.00) {Frame 21};
\draw[rqTeal,line width=.75pt,rounded corners=1pt] (1.9352,5.9298) rectangle (2.4732,7.6251);
\node[anchor=north west,fill=rqTeal,text=white,inner sep=1.3pt,rounded corners=1pt,font=\normalfont\fontsize{9}{10}\selectfont] at (1.9352,5.4898) {stand 0.99};
\draw[rqBlue,line width=.75pt,rounded corners=1pt] (0.9908,6.4649) rectangle (1.6075,7.6725);
\draw[rqBlue,line width=.45pt] (1.6575,6.4649) -- (0.9908,6.4649);
\node[anchor=north west,fill=rqBlue,text=white,inner sep=1.3pt,rounded corners=1pt,font=\normalfont\fontsize{9}{10}\selectfont] at (1.6575,6.4649) {hand clap 0.05};
\node[anchor=north west] at (4.800,4.420) {\includegraphics[width=4.37cm]{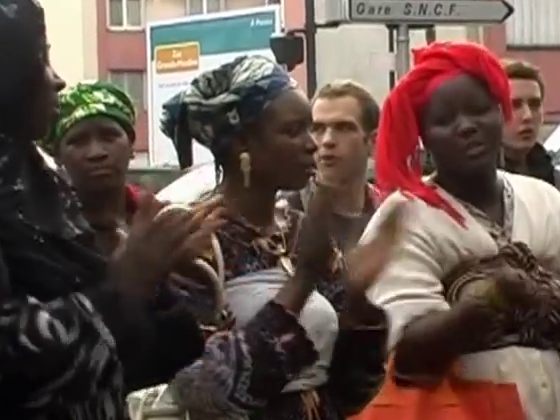}};
\node[text=rqInk] at (6.985,8.00) {Frame 26};
\draw[rqBlue,line width=.75pt,rounded corners=1pt] (5.9620,4.8497) rectangle (7.2693,6.5106);
\node[anchor=north west,fill=rqBlue,text=white,inner sep=1.3pt,rounded corners=1pt,font=\normalfont\fontsize{9}{10}\selectfont] at (5.9620,4.4450) {hand clap 1.00};
\draw[rqTeal,line width=.75pt,rounded corners=1pt] (5.1174,5.0009) rectangle (6.5368,7.4178);
\draw[rqTeal,line width=.45pt] (5.1174,5.7179) -- (5.1174,5.0009);
\node[anchor=north west,fill=rqTeal,text=white,inner sep=1.3pt,rounded corners=1pt,font=\normalfont\fontsize{9}{10}\selectfont] at (4.8300,5.7179) {stand 0.92};
\node[anchor=north west] at (9.470,4.420) {\includegraphics[width=4.37cm]{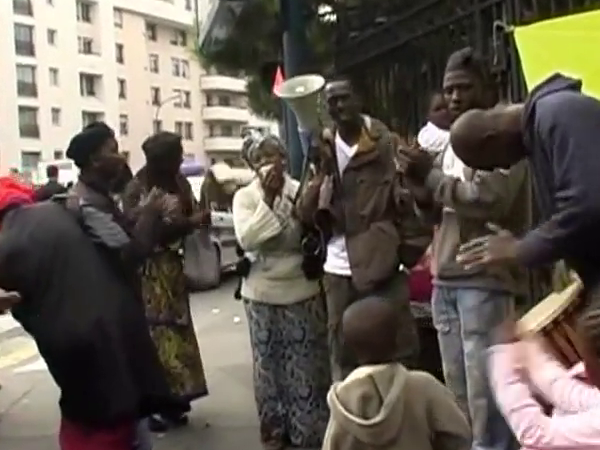}};
\node[text=rqInk] at (11.655,8.00) {Frame 66};
\draw[rqTeal,line width=.75pt,rounded corners=1pt] (11.1533,5.2836) rectangle (11.8536,7.6975);
\node[anchor=north west,fill=rqTeal,text=white,inner sep=1.3pt,rounded corners=1pt,font=\normalfont\fontsize{9}{10}\selectfont] at (11.1533,4.4450) {stand 1.00};
\node[anchor=north west,fill=rqBlue,text=white,inner sep=1.3pt,rounded corners=1pt,font=\normalfont\fontsize{9}{10}\selectfont] at (11.1533,4.8450) {hand clap 0.69};
\draw[black!20,line width=.4pt] (0,8.35) -- (13.97,8.35);
\node[anchor=west] at (0,8.61) {Video rank $\downarrow$};
\node[anchor=west,text=black!65] at (0,9.01) {of 952};
\fill[rqTeal!8] (2.57,8.41) rectangle (4.62,9.13);
\node[font=\bfseries,text=rqTeal] at (3.600,8.61) {LogSTOP};
\node[font=\bfseries,text=rqTeal] at (3.600,9.01) {2};
\node[] at (6.680,8.61) {Spatial robustness};
\node[] at (6.680,9.01) {3};
\node[] at (10.000,8.61) {CaptionSim};
\node[] at (10.000,9.01) {20};
\node[] at (12.830,8.61) {mPLUG};
\node[] at (12.830,9.01) {80};
\end{tikzpicture}
\endgroup

%% file: appendix/discussion.tex
\section{Additional discussion}

\subsection{Component ablations}
\label{appendix:discussion}

To isolate the importance of calibration, windowed smoothing, adaptive thresholding for matching, we ran query matching on QMTP-Video and ranked retrieval on TP2VR-Objects by ablating these components independently. The results of these component isolation ablations are presented in Table~\ref{tab:qm_ablations_full} for query matching and Table~\ref{tab:rr_ablations_full} for ranked retrieval.

\input{graphics/tables/ablations_tables}
\input{graphics/charts/calibration_plots}

\parhead{Calibration improves monotonicity of local property predictors.}
Figure~\ref{fig:reliability_diagram_qmtp} presents the reliability diagrams for YOLOv8x, Grounding DINO, and OWLv2 on the QMTP-video calibration split, before and after isotonic calibration. Before calibration, YOLOv8x already exhibits relatively monotonic reliability behavior, whereas Grounding DINO and especially OWLv2 show larger deviations from monotonicity. After calibration, all three predictors either improve or retain this monotonic reliability ordering. These trends are consistent with the role of calibration in making local predictor scores better aligned with empirical correctness.

Table~\ref{tab:qm_ablations_full} shows how these changes affect downstream query matching, with various choices of thresholds. The accuracy gains from calibration are aligned with the relative improvements in monotonicity: OWLv2 improves by \(5.12\%\) in query matching accuracy, Grounding DINO by \(1.62\%\), and YOLOv8x remains competitive. This pattern is consistent with the monotonicity properties of LogSTOP: when local scores are better ordered with respect to empirical correctness, the resulting sequence-level scores provide a more reliable basis for matching decisions.

\parhead{Most prediction errors are transient but effect of windowed smoothing is minimal.}
In Figure~\ref{fig:smoothing_stats}, we present the error run-length statistics for GroundingDINO, OWLv2, and YOLOv8x, on the calibration split of QMTP-video. The plots report how often these predictors incorrectly score contiguous sequences of various lengths. We find that most prediction errors are \textit{transient}, i.e., concentrated in short sequences of 1-2 frames across predictors, validating the need for windowed smoothing qualitatively. 
Quantitatively, query matching accuracy dips by at most \(0.83\%\) (Table~\ref{tab:qm_ablations_full}) and ranked retrieval mAP by at most \(0.05\) when no windowed smoothing is performed (Table~\ref{tab:rr_ablations_full}).

\input{graphics/charts/smoothing_stats}

\parhead{Adaptive threshold is the most useful component for query matching.}
The adaptive threshold is the largest contributor to query matching performance. When we replace it with a constant \(\ln 0.5\) threshold, the query matching accuracy drops by \(11.18\%\) for GroundingDINO, \(9.64\%\) for OWLv2, and \(8.55\%\) for YOLOv8 (Table~\ref{tab:qm_ablations_full}).

\subsection{Comparison of LogSTOP to other symbolic and neural aggregators}
\label{appendix:score_aggregators}

To isolate the usefulness of LogSTOP as a scoring function, we compare it against (a) a symbolic scoring function, the \(\min-\max\) based STL Spatial Robustness, (b) a learned transformer-based scorer, and (c) LLMs prompted to output a matching decision given the same information as LogSTOP.

\parhead{LogSTOP vs. STL Spatial Robustness.} Table~\ref{tab:results_qm} shows that LogSTOP improves query matching accuracy modestly by up to \(3.9\%\) (with calibrated GroundingDINO). Importantly, Table~\ref{tab:retrieval_results} shows that LogSTOP reports strictly better Precision@1 for ranked retrieval, up to \(19\%\) higher than STL Spatial Robustness with YOLO. This indicates that LogSTOP retrieves a relevant video at the first rank more often than STL robustness, making it a more useful scoring function for ranked retrieval. We present qualitative examples comparing LogSTOP and STL Robustness for ranking in Appendix~\ref{appendix:semantics}.

\parhead{LogSTOP vs. a learned transformer-based scorer.} We also train
a DistilBERT model to score temporal properties from the same calibrated and smoothed YOLOv8x detection traces. As shown in Table~\ref{tab:results_qm} and Table~\ref{tab:retrieval_results}, LogSTOP outperforms this neural scorer by \(13\%\) in query matching accuracy and by \(10\%\) in retrieval Precision@1, indicating that the gains are not merely due to access to detector outputs. The training details for this DistilBERT model are presented in Appendix~\ref{appendix:baselines_qm}.

\parhead{LogSTOP vs. LLMs for query matching.}
On a stratified sample of 150 queries from QMTP-Video, we also evaluate 2 LLMs as matchers: Llama-3.1-8B-Instruct and Qwen3-14B (in thinking mode). The prompts include the same frame-level YOLOv8x detector scores used by LogSTOP (post calibration and smoothing). The LLMs are also provided with the class-specific thresholds that we use for computing the adaptive threshold for matching with LogSTOP. We tested two ways of presenting the temporal queries: in Natural Language (those provided to our LVLM baselines, listed in Appendix~\ref{appendix:queries}) and as LTL formulae (with a syntax and semantics guide).
We present the prompts to the LLMs in Appendix~\ref{appendix:llm_ablation_prompts}.

\input{graphics/charts/llm_query_matching}

The results are presented in Table~\ref{tab:llms_qm}. Here, accuracy is computed over all 150 examples, counting invalid LLM outputs as incorrect. Valid output rate is the fraction of examples for which the model returned a parseable Yes/No matching decision. Accuracy on valid subset evaluates the LLM only on those valid outputs, and the final column reports LogSTOP accuracy on the same valid subset for a matched comparison.

Overall, even with the same detector information, text-only LLMs do not reliably perform temporal aggregation. 
Our key findings are as follows:

\underline{\textit{LLMs often fail to produce a usable matching decision.}} Output-valid rates are low, especially for Qwen3-14B: \(38.7\%\) with NL queries and \(30.7\%\) with LTL formulae given a 1024-token budget. For Qwen with NL queries, valid outputs also decrease sharply as formula complexity increases: from \(66.7\%\) for simple temporal operators to only \(3.3\%\) for formulas combining Boolean and nested temporal structure. This suggests that as temporal composition becomes more nested, the model is more likely to get stuck in reasoning and fail to emit a final decision. LogSTOP does not have this failure mode because it deterministically returns a score for every query.  

\underline{\textit{Even on valid LLM outputs, LogSTOP remains stronger.}} On the valid subset, LogSTOP outperforms the LLM in every setting: \(75.7\%\) vs. \(59.5\%\) for Llama and \(82.8\%\) vs. \(81.0\%\) for Qwen on NL queries. Qwen performs better than Llama on valid outputs, suggesting stronger reasoning models perform well in cases when they produce a valid response.  

\underline{\textit{LLMs perform better with NL queries than with LTL formulae.}} Overall accuracy drops from \(44.0\%\) to \(23.3\%\) for Llama and from \(31.3\%\) to \(22.7\%\) for Qwen when replacing NL queries with LTL formulae. Thus, directly providing the formal query does not solve the aggregation problem; models still struggle to reason reliably over frame-level score traces and temporal semantics.  

\subsection{Larger VLMs for query matching}
\label{appendix:internvl}

We also evaluate \texttt{InternVL2-26B}~\citep{chen2024internvl} on QMTP-Video. This comparison is restricted to videos with up to \(30\) frames due to the $8K$ context window size supported with 5 A100s.

Table~\ref{tab:qm_internvl} presents results on this 4500 sample subset. \texttt{InternVL2-26B} achieves
\(63.09\%\) balanced accuracy, compared with \(62.38\%\) for
\texttt{LongVA-7B}, despite being a significantly larger model.
Furthermore, LogSTOP with calibrated YOLOv8x achieves \(77.76\%\), exceeding
\texttt{InternVL2-26B} by \(14.67\%\).

\input{graphics/tables/query_matching_internvl_subset}

\subsection{Empirical efficiency of LogSTOP}
\label{appendix:comp_efficiency}
\parhead{LogSTOP computation scales linearly with sequence length.}
Figure~\ref{fig:efficiency_plots} shows that the time required to compute LogSTOP scales approximately linearly with sequence length for a fixed temporal property, consistent with the complexity analysis in Appendix~\ref{proof:complexity}.

\input{graphics/charts/efficiency_plots}

\subsection{LLMs for translating Natural Language queries to LTL properties}
\label{appendix:llm_translate}

We evaluated two open-source LLMs, Llama-3.1-8B-Instruct and Qwen3-14B, on translating natural language temporal queries into LTL formulae in a restricted QMTP-Video setting. The NL queries were sampled from the same query distribution as our LVLM baselines, with templates listed in Appendix~\ref{appendix:queries}. The prompt specified the allowed objects, LTL operators, finite-trace semantics, and rules for assigning the arguments of nested operators. We present the prompts to the LLMs in Appendix~\ref{appendix:llm_ablation_prompts}.  Our choice of models and prompts lets us test the effect of model family, the role of reasoning (with Qwen’s thinking mode), and choice of few-shot examples. We evaluated 75 QMTP-Video queries, with 5 queries for each of the 15 temporal templates, and the results are presented in Table~\ref{tab:llm_parsing}.

\input{graphics/charts/llm_parsing}

Overall, NL-to-LTL parsing works reasonably well in this restricted setting, with exact-match accuracy between \(73.3\%\) and \(84.0\%\). 

\underline{\textit{Smaller models such as Llama-3.1-8B-Instruct benefit from more few-shot examples}} included in the prompt: exact match accuracy improves from \(73.3\%\) with \(3\) examples (with nested formulae) to \(84.0\%\) when the prompt includes examples covering all \(15\) templates. Errors with this model primarily involve temporal scope: swapped until roles, e.g.,
\((\mathsf{car} \UntilOp{} \neg\mathsf{bicycle})
\longrightarrow
((\neg\mathsf{bicycle}) \UntilOp{} (\BoxOp{}\,\mathsf{car}))\),
and dropped or inserted temporal operators, e.g.,
\(((\neg\mathsf{car}) \UntilOp{} (\DiaOp{}\,\mathsf{bicycle}))
\longrightarrow
((\neg\mathsf{car}) \UntilOp{} \mathsf{bicycle})\).

\textit{Reasoning models such as Qwen3-14B report high performance with few reasoning-driven examples} (\(82.7\%\)) but could engage in long reasoning chains without outputting the final LTL formula (lower valid output rates of \(73-82\%\)). Interestingly, Qwen3-14B mentions the correct gold formula in the response in \(86.7\%\) of the cases but continues to reason in \(4\%\) of those. 

This experiment suggests that LLMs can be used to parse LTL formulae from NL queries in restricted settings, but robust use requires both good examples and strict output control. Future work can focus on correctly translating queries with nested composition and extracting temporal properties from longer, more general NL queries.

%% file: graphics/tables/ablations_tables.tex
\definecolor{qmpositive}{RGB}{0,110,60}
\definecolor{qmnegative}{RGB}{180,35,35}
\definecolor{qmneutral}{RGB}{110,110,110}

\newcommand{\qmdelta}[1]{%
  \ifdim #1pt>0pt
    \textcolor{qmpositive}{$(+#1\%)$}%
  \else\ifdim #1pt<0pt
    \textcolor{qmnegative}{$(#1\%)$}%
  \else
    \textcolor{qmneutral}{$(#1\%)$}%
  \fi\fi
}

\begin{table}[t]
  \centering
  \caption{Query matching ablations on QMTP-video.
  Parentheses show percentage-point changes relative to
  Default, colored green for positive, red for negative,
  and gray for zero. Best accuracy for each detector is in bold.}
  \label{tab:qm_ablations}

  \setlength{\tabcolsep}{3pt}
  \renewcommand{\arraystretch}{1.08}

  \begin{tabular*}{\linewidth}{@{\extracolsep{\fill}}lcclcr@{}}
    \toprule
    Setting & Calib. & Smooth. & Aggregation & Threshold
    & Accuracy (\%) $\uparrow$ \\
    \midrule

    \multicolumn{6}{@{}l}{\textbf{YOLOv8}} \\
    \addlinespace[2pt]
    LogSTOP (default)  & Yes & Yes & LogSTOP            & Adaptive    & 80.26 \qmdelta{+0.00}  \\
    Replace aggregation & Yes & Yes & Spatial Robustness & Adaptive    & 77.02 \qmdelta{-3.24} \\
    No smoothing        & Yes & No  & LogSTOP            & Adaptive    & 80.04 \qmdelta{-0.22} \\
    No calibration      & No  & Yes & LogSTOP            & Adaptive    & \textbf{80.43} \qmdelta{0.17} \\
    Fixed threshold     & Yes & Yes & LogSTOP            & $\ln(0.5)$ & 71.71 \qmdelta{-8.55} \\
    \midrule

    \multicolumn{6}{@{}l}{\textbf{OWLv2}} \\
    \addlinespace[2pt]
    LogSTOP (default)  & Yes & Yes & LogSTOP            & Adaptive    & \textbf{79.05} \qmdelta{+0.00} \\
    Replace aggregation & Yes & Yes & Spatial Robustness & Adaptive    & 78.92 \qmdelta{-0.13} \\
    No smoothing        & Yes & No  & LogSTOP            & Adaptive    & 78.22 \qmdelta{-0.83} \\
    No calibration      & No  & Yes & LogSTOP            & Adaptive    & 73.93 \qmdelta{-5.12} \\
    Fixed threshold     & Yes & Yes & LogSTOP            & $\ln(0.5)$ & 69.41 \qmdelta{-9.64} \\
    \midrule

    \multicolumn{6}{@{}l}{\textbf{Grounding DINO}} \\
    \addlinespace[2pt]
    LogSTOP (default)  & Yes & Yes & LogSTOP            & Adaptive    & \textbf{75.81} \qmdelta{+0.00}   \\
    Replace aggregation & Yes & Yes & Spatial Robustness & Adaptive    & 71.90 \qmdelta{-3.91}  \\
    No smoothing        & Yes & No  & LogSTOP            & Adaptive    & 75.20 \qmdelta{-0.61}  \\
    No calibration      & No  & Yes & LogSTOP            & Adaptive    & 74.19 \qmdelta{-1.62}  \\
    Fixed threshold     & Yes & Yes & LogSTOP            & $\ln(0.5)$ & 64.63 \qmdelta{-11.18} \\
    
    \bottomrule
  \end{tabular*}
\label{tab:qm_ablations_full}
\end{table}


\definecolor{retgood}{RGB}{0,110,60}
\definecolor{retbad}{RGB}{180,35,35}
\definecolor{retzero}{RGB}{110,110,110}

\newcommand{\retdelta}[2][1]{%
  \begingroup
  \ifdim #2pt=0pt
    \color{retzero}%
  \else\ifdim\dimexpr #2pt*#1\relax>0pt
    \color{retgood}%
  \else
    \color{retbad}%
  \fi\fi
  \ensuremath{(\ifdim #2pt>0pt+\fi#2)}%
  \endgroup
}

\begin{table}[t]
  \centering
  \caption{Retrieval ablations on the same 42-query cohort.
  LogSTOP (default) uses uncalibrated predictions, selected smoothing,
  and LogSTOP aggregation. Each ablation changes one component.
  Without smoothing uses radius zero; replacement aggregation uses
  Spatial Robustness. Parentheses show absolute changes from the default.
  Metrics and deltas are rounded independently to two decimal places.
  Green/red indicate improvement/degradation; gray indicates changes
  that round to zero. Bold marks the best values before rounding.}
  \label{tab:retrieval_ablations}
  \label{tab:rr_ablations_full}
  \vspace{6pt}

  \setlength{\tabcolsep}{3pt}
  \renewcommand{\arraystretch}{1.08}

  \begin{tabular*}{\linewidth}{@{\extracolsep{\fill}}lrrrr@{}}
    \toprule
    Setting & P@1 $\uparrow$ & mAP $\uparrow$
    & R@$r$ $\uparrow$ & MnR $\downarrow$ \\
    \midrule

    \multicolumn{5}{@{}l}{\textbf{YOLOv8}} \\
    \addlinespace[2pt]
    LogSTOP (default)
      & \textbf{0.36} \retdelta{+0.00}
      & 0.42 \retdelta{+0.00}
      & \textbf{0.41} \retdelta{+0.00}
      & 4.81 \retdelta[-1]{+0.00} \\
    With calibration
      & 0.21 \retdelta{-0.14}
      & 0.41 \retdelta{-0.01}
      & 0.40 \retdelta{+0.00}
      & 5.52 \retdelta[-1]{0.71} \\
    Without smoothing
      & \textbf{0.36} \retdelta{+0.00}
      & 0.42 \retdelta{-0.01}
      & 0.41 \retdelta{+0.00}
      & 5.31 \retdelta[-1]{0.50} \\
    Replace aggregation
      & 0.17 \retdelta{-0.19}
      & \textbf{0.42} \retdelta{+0.00}
      & 0.41 \retdelta{+0.00}
      & \textbf{4.69} \retdelta[-1]{-0.12} \\
    \midrule

    \multicolumn{5}{@{}l}{\textbf{OWLv2}} \\
    \addlinespace[2pt]
    LogSTOP (default)
      & 0.81 \retdelta{+0.00}
      & \textbf{0.64} \retdelta{+0.00}
      & \textbf{0.58} \retdelta{+0.00}
      & \textbf{1.71} \retdelta[-1]{+0.00} \\
    With calibration
      & \textbf{0.86} \retdelta{0.05}
      & 0.63 \retdelta{-0.01}
      & 0.58 \retdelta{+0.00}
      & 2.67 \retdelta[-1]{0.95} \\
    Without smoothing
      & 0.76 \retdelta{-0.05}
      & 0.61 \retdelta{-0.03}
      & 0.54 \retdelta{-0.04}
      & 1.76 \retdelta[-1]{0.05} \\
    Replace aggregation
      & 0.69 \retdelta{-0.12}
      & 0.62 \retdelta{-0.02}
      & 0.57 \retdelta{-0.02}
      & 2.12 \retdelta[-1]{0.40} \\
    \midrule

    \multicolumn{5}{@{}l}{\textbf{Grounding DINO}} \\
    \addlinespace[2pt]
    LogSTOP (default)
      & \textbf{0.81} \retdelta{+0.00}
      & \textbf{0.57} \retdelta{+0.00}
      & \textbf{0.52} \retdelta{+0.00}
      & \textbf{1.98} \retdelta[-1]{+0.00} \\
    With calibration
      & 0.76 \retdelta{-0.05}
      & 0.53 \retdelta{-0.04}
      & 0.49 \retdelta{-0.03}
      & 3.12 \retdelta[-1]{1.14} \\
    Without smoothing
      & 0.76 \retdelta{-0.05}
      & 0.52 \retdelta{-0.05}
      & 0.45 \retdelta{-0.07}
      & 3.79 \retdelta[-1]{1.81} \\
    Replace aggregation
      & 0.76 \retdelta{-0.05}
      & 0.57 \retdelta{+0.00}
      & 0.51 \retdelta{-0.01}
      & 2.17 \retdelta[-1]{0.19} \\
    \bottomrule
  \end{tabular*}
\end{table}

%% file: graphics/charts/calibration_plots.tex
\begin{figure}[H]
\centering
\includegraphics[width=0.95\linewidth]{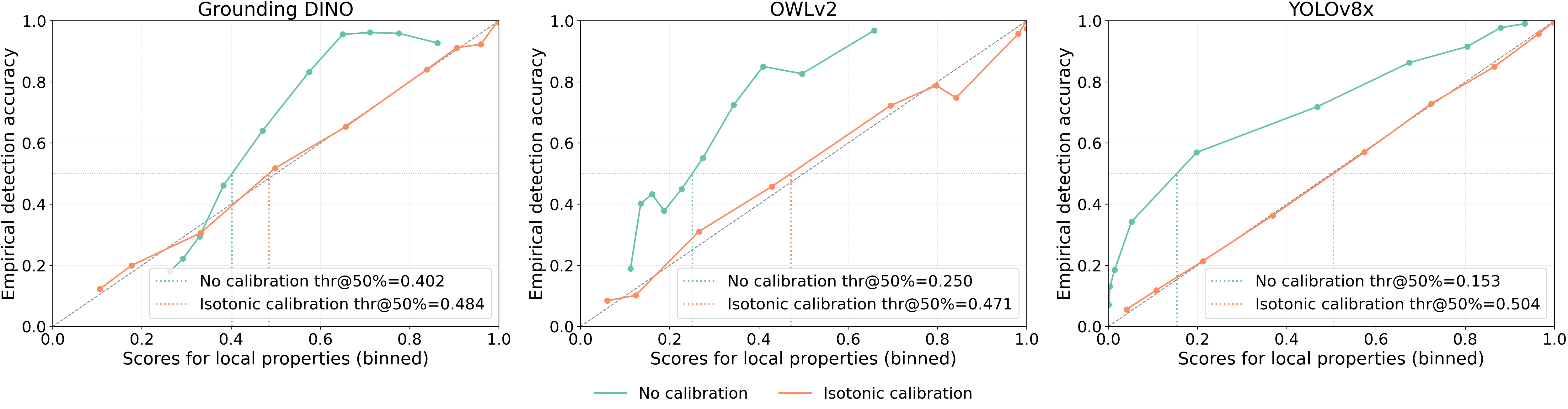}
\caption{
Reliability plots for the three predictors on the calibration split of QMTP-video.
}
\label{fig:reliability_diagram_qmtp}
\end{figure}

%% file: graphics/charts/smoothing_stats.tex
\begin{figure}[h]
\centering
\includegraphics[width=0.95\linewidth]{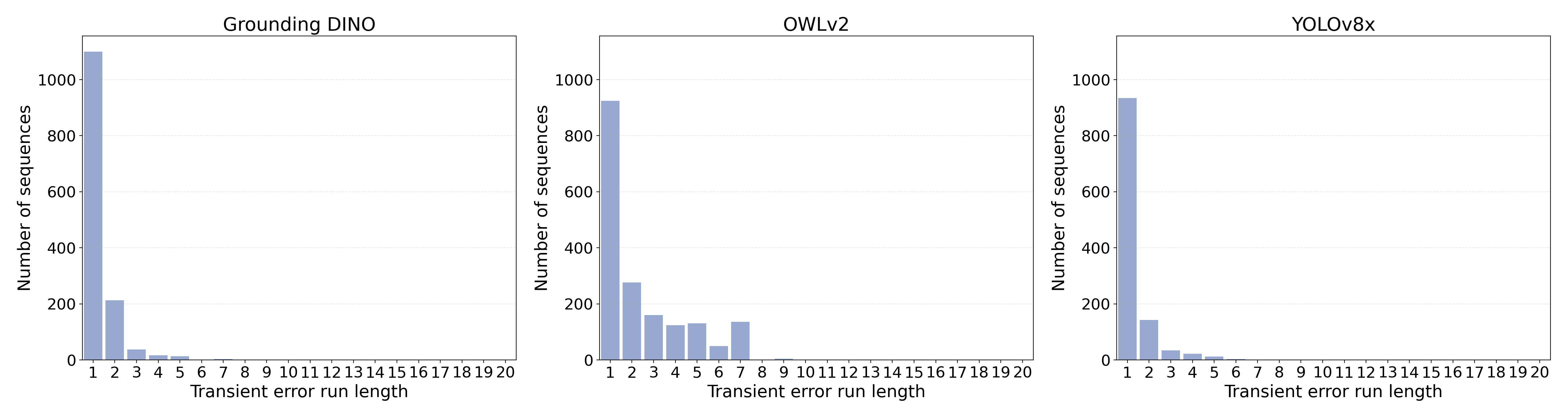}
\caption{
Length of transient error runs for the three predictors.
}
\label{fig:smoothing_stats}
\end{figure}

%% file: graphics/charts/llm_query_matching.tex
\begin{table}[t]
  \centering
  \caption{Query matching performance and output validity.
  The last two columns compare each method and LogSTOP on the same
  subset of examples with valid outputs from that method.
  All numerical values are percentages.}
  \label{tab:llm_query_matching}
  \vspace{6pt}

  \setlength{\tabcolsep}{3pt}
  \renewcommand{\arraystretch}{1.08}

  \begin{tabular*}{\linewidth}{@{\extracolsep{\fill}}llrrrr@{}}
    \toprule
    & & & &
    \multicolumn{2}{c}{\shortstack{Accuracy on\\valid output subset (\%)}} \\
    \cmidrule(lr){5-6}
    Method / model
    & \shortstack{Query\\type}
    & \shortstack{Accuracy\\(\%)}
    & \shortstack{Valid outputs\\(\%)}
    & Method
    & LogSTOP \\
    \midrule

    Llama-3.1-8B-Instruct
      & NL query    & 44.0 & 74.0 & 59.5 & 75.7 \\
      & LTL formula & 23.3 & 33.3 & 70.0 & 82.0 \\
    \addlinespace[3pt]

    Qwen3-14B
      & NL query    & 31.3 & 38.7 & 81.0 & 82.8 \\
      & LTL formula & 22.7 & 30.7 & 73.9 & 84.8 \\
    \midrule

    LogSTOP
      & LTL formula & \textbf{78.0} & \textbf{100.0} & 78.0 & 78.0 \\
    \bottomrule
  \end{tabular*}
\label{tab:llms_qm}
\end{table}

%% file: graphics/tables/query_matching_internvl_subset.tex
\begin{table}[h]
\caption{Query matching balanced accuracy by temporal property category ($\uparrow$). QMTP-video is restricted to the common 4,500 examples with 10, 20, or 30 frames (2,250 matching and 2,250 non-matching). The best / second-best distinct scores at the displayed precision are \textbf{bold} / \underline{underlined} within each dataset and column.}
\centering
\begingroup
\setlength{\tabcolsep}{4pt}
\renewcommand{\arraystretch}{1.08}
\begin{tabular*}{\linewidth}{@{\extracolsep{\fill}}lccccc@{\hspace{12pt}}c@{}}
\toprule
Method / local predictor & Simple & \shortstack{Bool. over\\ Temp.} & \shortstack{Temp. over\\ Bool.} & \shortstack{Temp. over\\ Temp.} & Mixed & Overall \\
\midrule
\addlinespace[2pt]
Video-LLaVA-7B & 0.50 & 0.50 & 0.50 & 0.50 & 0.50 & 0.50 \\
LongVA-7B & 0.69 & 0.63 & 0.61 & 0.60 & 0.60 & 0.62 \\
InternVL2-26B & 0.69 & 0.59 & 0.63 & 0.63 & 0.60 & 0.63 \\
DistilBERT (trained on YOLO) & 0.76 & 0.66 & 0.57 & \textbf{0.76} & 0.72 & 0.69 \\
NSVS-TL & 0.65 & 0.62 & 0.51 & 0.61 & 0.50 & 0.57 \\
\addlinespace[4pt]
\multicolumn{7}{@{}l}{\textit{Spatial Robustness (calibrated predictors)}} \\
\quad Grounding DINO & 0.74 & 0.68 & 0.70 & 0.70 & 0.73 & 0.71 \\
\quad OWLv2 & \underline{0.79} & 0.71 & 0.74 & \underline{0.75} & \underline{0.77} & 0.75 \\
\quad YOLO & 0.78 & 0.71 & 0.74 & \underline{0.75} & \underline{0.77} & 0.75 \\
\addlinespace[4pt]
\multicolumn{7}{@{}l}{\textit{LogSTOP (calibrated predictors)}} \\
\quad Grounding DINO & 0.77 & 0.72 & 0.71 & 0.71 & 0.75 & 0.73 \\
\quad OWLv2 & \underline{0.79} & \underline{0.75} & \underline{0.75} & 0.74 & \textbf{0.78} & \underline{0.76} \\
\quad YOLO & \textbf{0.81} & \textbf{0.76} & \textbf{0.78} & \textbf{0.76} & \textbf{0.78} & \textbf{0.78} \\
\bottomrule
\end{tabular*}
\endgroup
\label{tab:qm_internvl}
\end{table}

%% file: graphics/charts/efficiency_plots.tex
\begin{figure}[h]

    \begin{subfigure}{0.32\textwidth} 
        \centering 
        \includegraphics[width=\linewidth]{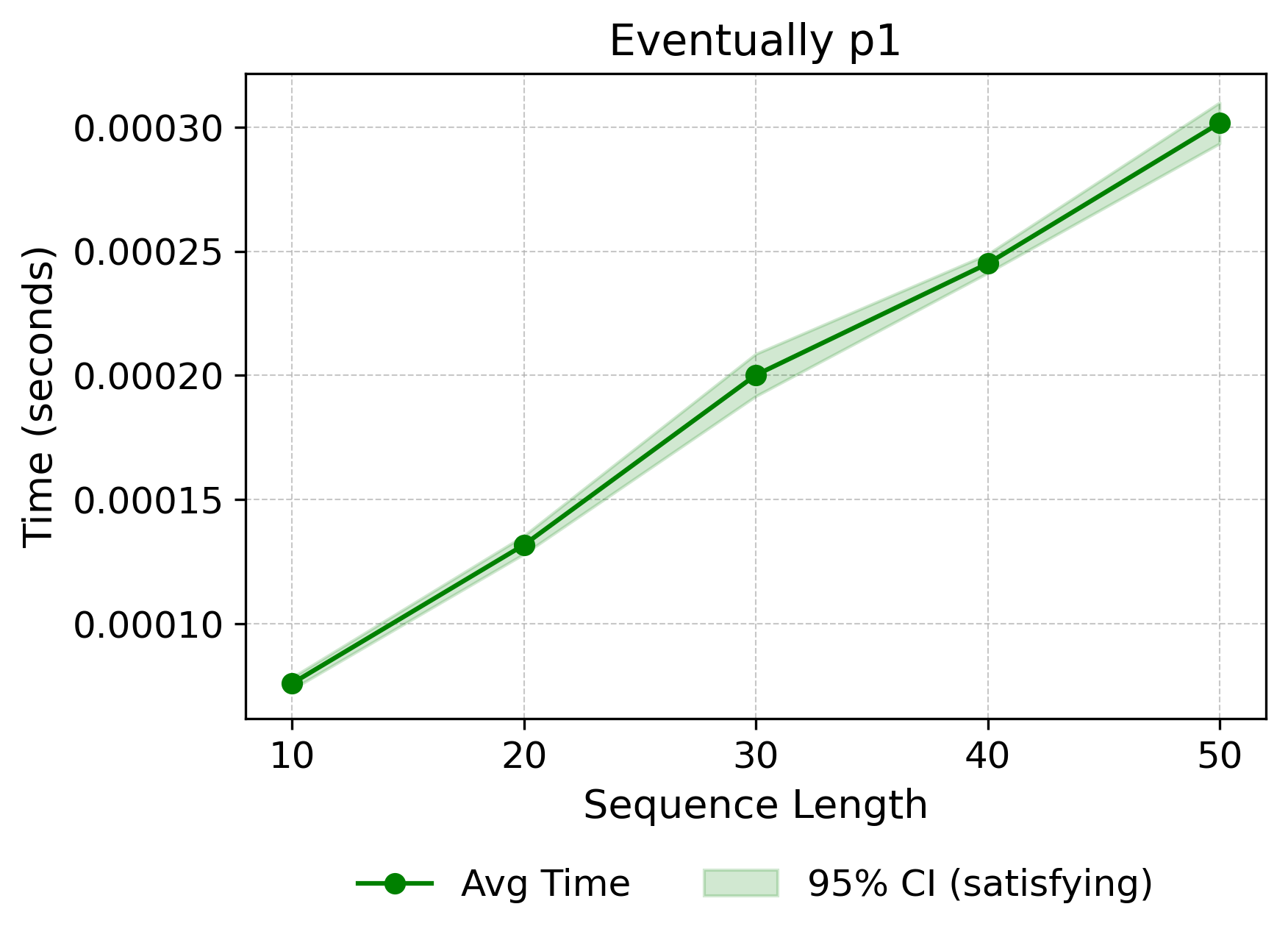} %
    \end{subfigure}
    \hfill 
    \begin{subfigure}{0.32\textwidth} 
        \centering 
        \includegraphics[width=\linewidth]{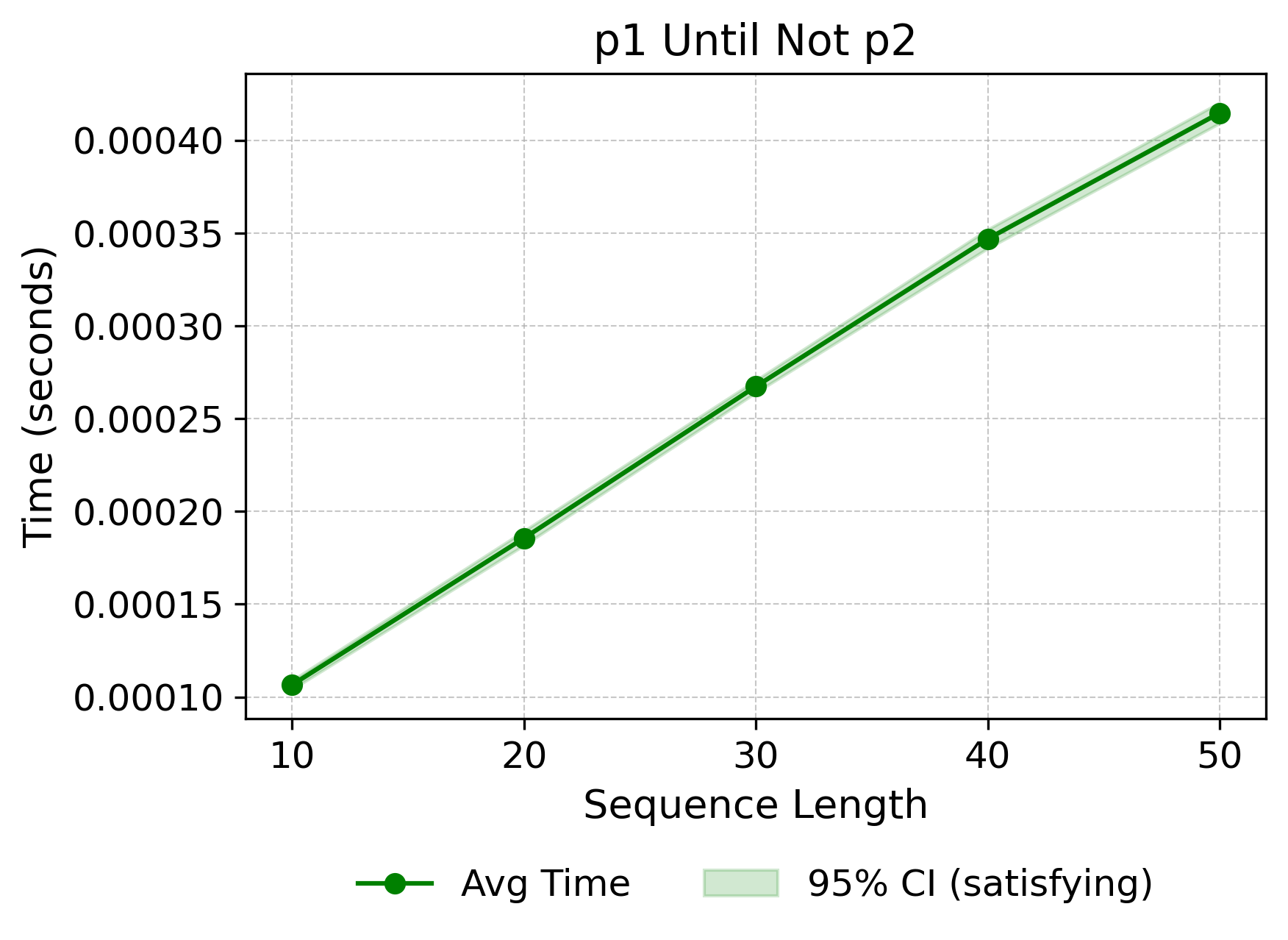} %
    \end{subfigure}
    \hfill 
    \begin{subfigure}{0.32\textwidth} 
        \centering 
        \includegraphics[width=\linewidth]{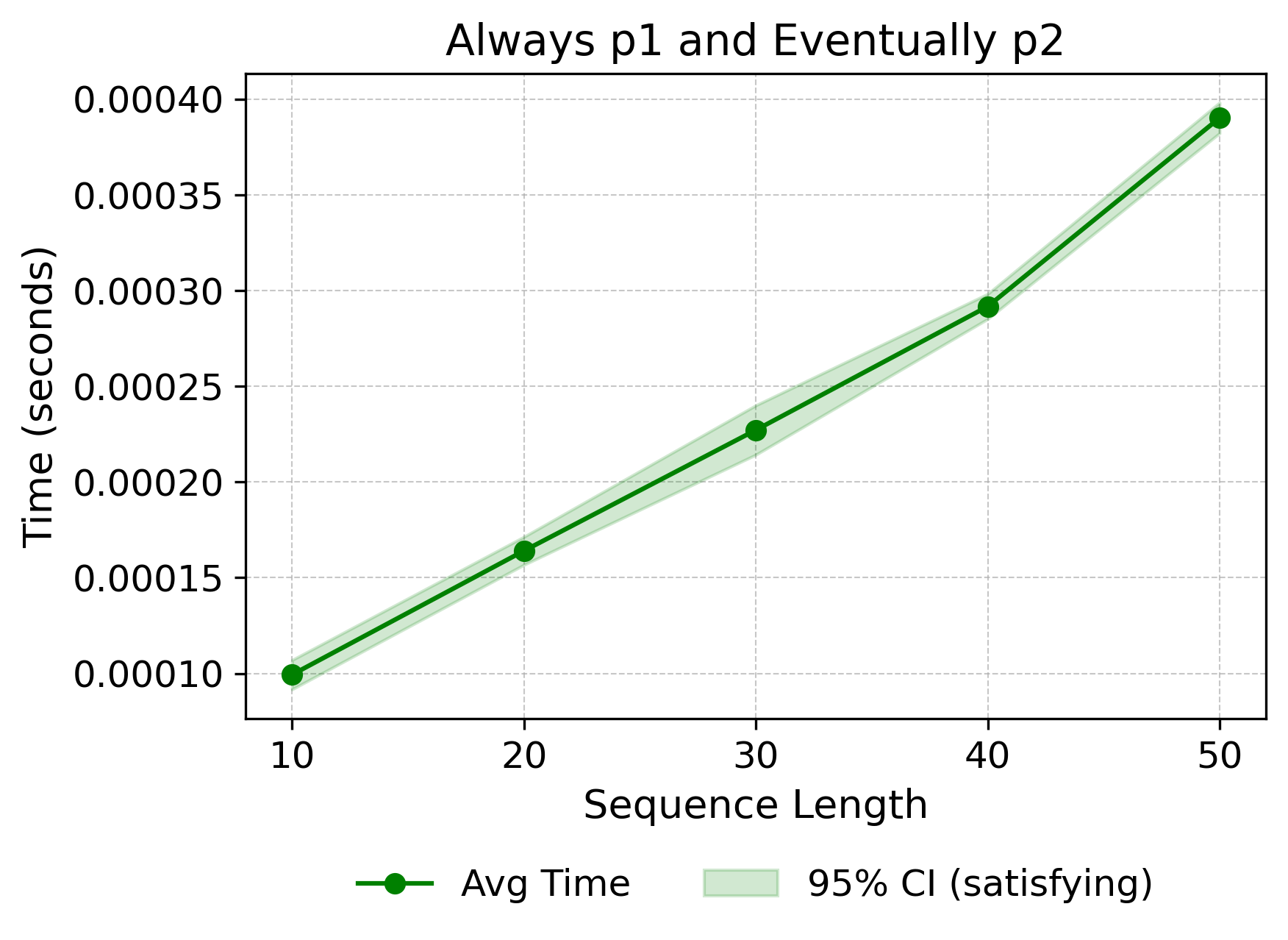} %
    \end{subfigure}
    \caption{Time taken for query matching with LogSTOP scales linearly with time for temporal properties of varying lengths. The results are using YOLOv8x on the QMTP-Video benchmark. Note that this does not include the time taken by the object detector.}
    \label{fig:efficiency_plots}
\end{figure}

%% file: graphics/charts/llm_parsing.tex
\begin{table}[t]
  \centering
  \caption{Few-shot formula generation results.
  The final column reports the percentage of responses mentioning
  the gold formula. All performance values are percentages.}
  \label{tab:few_shot_formula_generation}
  \vspace{6pt}

  \setlength{\tabcolsep}{3pt}
  \renewcommand{\arraystretch}{1.08}

  \begin{tabular*}{\linewidth}{@{\extracolsep{\fill}}llcrrr@{}}
    \toprule
    Model
    & \shortstack{Few-shot\\examples}
    & Thinking
    & \shortstack{Exact\\match (\%)}
    & \shortstack{Valid\\outputs (\%)}
    & \shortstack{Gold formula\\in response\\(\%)} \\
    \midrule

    Llama-3.1-8B-Instruct
      & 15 (all templates)
      & No & 84.0 & 100.0 & 84.0 \\
      & 3 (selected queries)
      & No & 73.3 & 97.3 & 73.3 \\
    \midrule

    Qwen3-14B
      & 3 (selected queries)
      & No & 76.0 & 100.0 & 76.0 \\
      & 3 (selected queries)
      & Yes & 73.3 & 73.3 & 82.7 \\
      & \shortstack[l]{3 (selected queries,\\with reasoning)}
      & Yes & 82.7 & 82.7 & 86.7 \\
    \bottomrule
  \end{tabular*}
\label{tab:llm_parsing}
\end{table}

%% file: appendix/datasets.tex
\section{More Details on Datasets}
\label{appendix:datasets}

In Section~\ref{sec:experiments}, we briefly discuss the QMTP and TP2VR benchmarks for evaluation.
For
constructing these benchmarks
, we use 
three existing datasets with frame/segment-level annotations for local properties:
The RealTLV dataset~\cite{Choi2024TowardsNV}
consists of videos from NuScenes~\cite{nuscenes} and Waymo~\cite{waymo} driving datasets with frame-level annotations for 6 object classes. 
(for example, "car", "truck", etc). 
The IEMOCAP dataset~\cite{busso2008iemocap} provides speech segments from conversations 
between two speakers 
and each segment is labeled with one of the 4 major emotions expressed by the speaker. 
(for example, "happy", "sad", etc.). 
The AVA dataset~\cite{gu2018ava} consists of frame-level action annotations for 80 actions in 15-min clips from YouTube. We only consider the validation subset of this dataset and sample 5 frames per second.
These datasets can be used for evaluating
temporal properties over objects in videos 
("car until pedestrian", for example)
, emotions in speech 
("always happy")
, and actions in videos 
("a person sits until they stand up") 
respectively.

\noindent{\textbf{The QMTP dataset.}}
For any temporal property template (for example, "p1 Until p2") and samples from these datasets, we identify matching and non-matching sequences of desired length as follows: for every sample, we first identify candidates for local properties in the template (p1, p2, etc.) as the set of all ground-truth objects / emotions in the sequence. We then use the standard LTL semantics over the frame/segment-level ground-truth labels to collect matching and non-matching subsequences of the desired length. This creates a TP-query matching dataset for an arbitrary set of temporal properties as long as these properties are sufficiently expressed by sequences from the underlying dataset. Moreover, this pipeline is agnostic to the choice of the dataset since it only requires sequences of ground-truth labels for local properties. We use this pipeline to create  
the \textbf{QMTP-video dataset} with \(7468\) samples (\(3750\) matching and \(3718\) non-matching) with video sequences of lengths \(\{10, 20, 30, 40, 50\}\). For each target length, this dataset contains approximately \(100\) samples corresponding to each of the \(15\) property templates. Similarly, we create the \textbf{QMTP-speech dataset} with \(3300\) samples, including speech sequences of lengths in ranges \(\{5-10, 10-20, 20-30\}\). The QMTP-speech dataset only contains samples from \(11/15\) property templates. This is because there are no sequences matching \(4\) properties "Always p1 and Eventually p2", "Always (p1 and p2)", "(p1 and p2) Until p3", and "(p1 and p2) Until Eventually p3". 

\noindent{\textbf{The TP2VR dataset.}}
We restrict queries to a maximum of 5 per temporal property template. For TP2VR objects, we aim to find \(25-50\) frame-long subsequences satisfying a temporal property; we only include a temporal property if less than \(250\) videos are retrieved using the standard LTL semantics with the ground truth labels. 
With all possible combinations of objects, this gives us a total of \(42\) queries, with an average of \(163\) videos relevant to a query.
Similarly, for TP2VR-actions, we aim to find 10-second long subsequences (\(50\) frames) satisfying a temporal property; we only include a temporal property if less than \(50\) videos are retrieved using the standard LTL semantics with the ground truth labels.  With all possible combinations of actions, this gives us a total of \(70\) queries, with an average of \(21\) videos relevant to a query.

We present a list of all 15 temporal query templates in Appendix~\ref{appendix:queries}.

%% file: appendix/qm_methods.tex
\section{More details on Query Matching Methods}
\label{appendix:baselines_qm}

\noindent{\textbf{LogSTOP with simple trained neural predictors:}} We use YOLOv8~\cite{yolov8_ultralytics}, Grounding DINO~\cite{liu2024grounding}, and OWLv2~\cite{minderer2023scaling} to get confidence scores for object detection in video frames, and HuBERT~\cite{superb} for speech emotion recognition in audio segments. These scores are used in their \([0,1]\) range; LogSTOP applies the \(\log\) operation internally when local evidence enters the temporal aggregation.
A video matches query \(\varphi\) if LogSTOP \(\hat{y}(X, \varphi, 1, T, \{w_c\}) > \tau_\alpha(\varphi, T, \{w_c\})\) (and vice versa for non-matching examples). The estimates are evaluated against ground-truth labels \(y(X, \varphi, 1)\). For QMTP-video, we use the held-out NuScenes split to calibrate detector scores, choose local-property-specific smoothing radii \(\{w_c\}\) by the correct-direction movement protocol, and estimate the reference thresholds used by the adaptive threshold. For QMTP-speech, where no comparable held-out split is available, we set \(w_c=1\) for all local properties and use the adaptive threshold with the shared reference threshold \(r_c(0.5)=0.5\).

\noindent{\textbf{Calibration details.}} 
For QMTP-video, we collect 37 videos from NuScenes~\cite{nuscenes} which are disjoint from the QMTP-video benchmark. We then aggregate frame-level label-score pairs for all local properties across all videos. For each local property \(c\), we collect the label-score pairs for all frames and all videos. Let \(p(X,c,t) \in [0,1]\) denote the raw predictor confidence. We fit an isotonic regressor \(f_c\) (monotone, range-clipped) on \((p(X,c,t), y(X,c,t))\), obtain calibrated probabilities \(p_{\mathrm{cal}}(X,c,t)=f_c(p(X,c,t))\), and then use \(\hat{y}_{\mathrm{cal}}(X,c,t)=p_{\mathrm{cal}}(X,c,t)\) as the local evidence supplied to LogSTOP. 
  We evaluate calibration using reliability diagrams and ECE
  with binned confidence statistics (10 bins using quantile binning), comparing empirical accuracy in each bin against mean predicted confidence (Figure~\ref{fig:reliability_diagram_qmtp}).
  We also derive operating thresholds per local property \(r_c(0.5)\) as the confidence where the reliability curve crosses
  \(50\%\) empirical accuracy, computed per class and overall (with per-class fallback to
  overall if no crossing exists).

\noindent{\textbf{DistilBERT trace classifier.}}
Our transformer baseline is a binary temporal-matching classifier built on DistilBERT (\texttt{distilbert-base-uncased}) with a 2-way classification head. Each training instance is
  text-serialized from detector outputs and the target query using XML-style fields: local-property names are paired with their score traces \((s_{i,1},\ldots,s_{i,T})\), followed by the temporal property \(\phi\).
where \(p_i\) are local properties and \(\phi\) is the temporal formula. The dataset was constructed using videos from NuScenes and has 18,356 query matching examples (14,958/1,526/1,872 train/val/test; 8,395 positive, 9,961 negative), spanning the 15 formula templates we use in our experiments. We train for 3 epochs (batch size 8, learning rate \(2\times 10^{-5}\), weight decay 0.01, seed 13), selecting the best checkpoint by
  validation F1. Final performance is 0.880 val accuracy /
  0.859 val F1 / 0.949 val AUROC and 0.858 test accuracy / 0.854 test F1 / 0.919 test
  AUROC.

\noindent{\textbf{STL Spatial Robustness.}}
We also include STL Spatial Robustness as a baseline with the same components used by LogSTOP (calibration, smoothing, threshold selection) but only differing in the aggregation scheme. This standard quantitative STL semantics aggregates with \(\min\)/\(\max\) operators; we provide details and examples comparing it with LogSTOP in Section~\ref{appendix:semantics}. 

\noindent{\textbf{Large Vision Language Models (LVLMs)}.} We evaluate three popular LVLMs on query matching for videos: \texttt{Video-LLava-7B}~\cite{videollava}, \texttt{LongVA-7B}~\cite{zhang2024longva}, and \texttt{InternVL2-26B}. 
For the "always car" example, we provide the models with the video sequence and a text prompt "Is a car detected in all frames of this video?". The response is considered correct if the model responds with "Yes" or "No" for matching and non-matching samples respectively. \texttt{Video-LLava-7B} supports a context window of \(4096\) tokens while \texttt{LongVA-7B} can handle up to 2000 frames. We set the maximum tokens to generate to \(60\) and \(1024\) respectively and use a temperature of 0.1 and standard values for the other parameters. We present a list of all natural language query templates in Appendix~\ref{appendix:queries}.

\noindent{\textbf{Large Audio Language Models (LALMs)}.} Similarly, we evaluate two popular LALMs on query matching for speech: \texttt{Qwen-Audio-Chat}~\cite{Qwen-Audio} and \texttt{Qwen2-Audio-7B-Instruct} ~\cite{Qwen2-Audio}. 
For the "eventually happy" example, we provide the models with the audio sequence and a text prompt "Does the speaker sound happy at some time in this audio clip?". We set the sampling rate to \(16000\) and generate a maximum of \(256\) new tokens, with standard values for other parameters.
We present a list of all natural language query templates in Appendix~\ref{appendix:queries}.

\noindent{\textbf{NSVS-TL~\cite{Choi2024TowardsNV}.}} Proposed for event detection in videos, NSVS-TL~\cite{Choi2024TowardsNV} uses the PCTL-based model checker STORM~\cite{storm} to identify video frame subsequences where a certain event is detected. NSVS-TL reports state-of-the-art performance on detecting temporal events in videos, surpassing large language models such as GPT-4. 
For our task, we specify the target query in PCTL ("always car" is \(P > 0.5 [G \) "\({car}\)" \(]\)) and the response is considered correct if NSVS-TL returns / does not return the entire video sequence as output for a matching / non-matching query respectively. We present a list of all PCTL query templates in Appendix~\ref{appendix:queries}.

We do not evaluate the method from ~\cite{yang2023specification} since the implementation is not publicly available and LTL model checking with STORM is not well-documented.

%% file: appendix/retrieval_methods.tex
\section{More details on the Ranked Retrieval methods}
\label{appendix:baselines_vr}

\noindent{\textbf{LogSTOP.}} 
We use LogSTOP with YOLOv8~\cite{yolov8_ultralytics}, Grounding DINO~\cite{liu2024grounding}, and OWLv2~\cite{minderer2023scaling} for TP2VR-objects and with SlowR50~\cite{feichtenhofer2019slowfast} for TP2VR-actions respectively. 
We repurpose the script from \texttt{tutorials/video\_detection\_example} at \texttt{https://github.com/facebookresearch/pytorchvideo/}
to run SlowR50 on videos from the TP2VR-actions dataset.
For TP2VR-objects, we use the same RealTLV held-out split to choose local-property-specific smoothing radii. For TP2VR-actions, where no comparable held-out split is available, we set \(w_c=1\) for all local properties and do not perform post-hoc calibration or held-out class-specific smoothing selection.

\noindent{\textbf{DistilBERT trace classifier.}}
We use the same DistilBERT trace classifier described in Appendix~\ref{appendix:baselines_qm}. For retrieval, each candidate video-query pair is serialized in the same detector-trace format and scored by the classifier head; videos are then ranked by the predicted match score.

\noindent{\textbf{STL Spatial Robustness.}}
We use the same STL Spatial Robustness setup described above for query matching, including the same smoothed local evidence and class-specific predicate thresholds. For retrieval, videos are ranked directly by their STL Spatial Robustness scores.

\noindent{\textbf{mPLUG.}} 
We use the implementation from \texttt{https://github.com/alibaba/AliceMind}. We repurpose the \texttt{mPLUG/retrieval\_vid\_mplug.py} script to run \texttt{mPLUG\_{large}\_v2} on videos and queries from the TP2VR datasets. We present a list of all natural language query templates in Appendix~\ref{appendix:queries}.

\noindent{\textbf{CaptionSim.}} 
Inspired by ELIOT~\cite{liu-etal-2025-eliot}, we also include embedding similarity between video captions and text queries as a baseline.
We refer to this as \texttt{CaptionSim} in the discussion, and use 
\texttt{LLaVA-NeXT-Video-7B}~\cite{zhang2024llavanextvideo} for generating video captions and \texttt{SentenceBERT/all-MiniLM-L6-v2}~\cite{reimers-2020-multilingual-sentence-bert} for embedding captions and queries. 
Due to the limited context window of \texttt{LLaVA-NeXT-Video-7B}, we divide videos in sections of 50 frames and generate captions for each before concatenating them together. We use the following prompt to generate the captions for the first 50 frames: "Describe this video in detail, listing objects in each frame. Keep the descriptions concise." for TP2VR-objects.
For any next sections, we use the prompt "Continue describing the video, listing objects in each frame. You are now at frame {i}, you have already described the previous {i} frames."
For TP2VR-actions, we use the prompt "Describe this video in detail, listing actions and objects in each frame. Keep the descriptions concise." We set the \texttt{max\_new\_tokens} to 1024. We present a list of all natural language query templates in Appendix~\ref{appendix:queries}.

We use \texttt{CaptionSim} because the implementation of ELIOT is not publicly available. Our local implementation of ELIOT did not report good results (the video captions generated by ~\cite{tewel2022zero} did not include mentions of objects or actions).

%% file: appendix/queries_prompts.tex
\section{Queries and Prompts}
\label{appendix:queries}

We choose \(15\) temporal property templates for the experiments in Section~\ref{sec:experiments}.

The LogSTOP queries for these templates are as follows:

\begin{enumerate}
    \item Eventually p1: \(\DiaOp{I}\, p1\)
    \item Always p1: \(\BoxOp{I}\,p1\)
    \item p1 Until p2: \(p1 \UntilOp{I} p2\)
    \item Always p1 and Eventually p2: \(\BoxOp{I}\,p1 \land \DiaOp{I}\,p2\)
    \item Always p1 or Eventually p2: \(\BoxOp{I}\,p1 \lor \DiaOp{I}\,p2\)
    \item (Not p1) Until p2: \(\neg p1 \UntilOp{I} p2\)
    \item p1 Until (Not p2): \(p1 \UntilOp{I} \neg p2\)
    \item Always (p1 and p2): \(\BoxOp{I} (p1 \land p2)\)
    \item (p1 and p2) Until p3: \((p1 \land p2) \UntilOp{I} p3\)
    \item p1 Until Always p2: \(p1 \UntilOp{I} \BoxOp{I}\, p2\)
    \item Eventually Always p1: \(\DiaOp{I}\,\BoxOp{I}\,p1\)
    \item Always Eventually p1: \(\BoxOp{I}\,\DiaOp{I}\,p1\)
    \item (Not p1) Until Eventually p2: \(\neg p1 \UntilOp{I} \DiaOp{I}\, p2\)
    \item (Not p1) Until Always p2: \(\neg p1 \UntilOp{I} \BoxOp{I}\, p2\)
    \item (p1 and p2) Until Eventually p3: \((p1 \land p2) \UntilOp{I} \DiaOp{I}\, p3\)
\end{enumerate}

NSVS-TL~\cite{Choi2024TowardsNV} uses the model checker STORM~\cite{storm} to verify if a given sequence satisfies a temporal property, where the temporal properties are represented in Probabilistic Computation Tree Logic (PCTL). In PCTL, the \(F\), \(G\) and \(U\) operators represent the \textit{Eventually}, \textit{Always} and \textit{Until} operators respectively. The \(\sim\), \(\&\) and \(\mid\) operators represent the Boolean \textit{negation}, \textit{and}, and \textit{or} operators respectively. The operator \(P\) is used to indicate the ranges of probability of a given property being satisfied: for example, \(P > 0.5 [F\, \varphi]\) translates to "the probability of \(\varphi\) eventually being satisfied is more than \(0.5\)".

The PCTL queries for the 15 temporal property templates are as follows:

\begin{enumerate}
    \item Eventually p1: \(P > 0.5\, [F \) "\({p1}\)" \(]\)
    \item Always p1: \(P > 0.5\, [G \) "\({p1}\)" \(]\)
    \item p1 Until p2: \(P > 0.5\, [ \) "\({p1}\)" \(U\) "\({p2}\)" \(]\)
    \item Always p1 and Eventually p2: \(P > 0.5\, [\,G \) "\({p1}\)" \(\&\, F\,\) "\({p2}\)" \(]\)
    \item Always p1 or Eventually p2: \(P > 0.5\, [\,G \) "\({p1}\)" \(\mid\, F\,\) "\({p2}\)" \(]\)
    \item (Not p1) Until p2:  \(P > 0.5\, [ \sim \) "\({p1}\)" \(U\) "\({p2}\)" \(]\)
    \item p1 Until (Not p2):  \(P > 0.5\, [\) "\({p1}\)" \(U \sim\) "\({p2}\)" \(]\)
    \item Always (p1 and p2): \(P > 0.5\, [\,G \) "\({p1}\)" \(\&\, G\,\) "\({p2}\)" \(]\)
    \item (p1 and p2) Until p3: \(P > 0.5\, [ \) "\({p1}\)" \(\&\) "\({p2}\)" \(U\) "\({p3}\)" \(]\)
    \item p1 Until Always p2:  \(P > 0.5\, [ \) "\({p1}\)" \(U\,G\,\) "\({p2}\)" \(]\)
    \item Eventually Always p1: \(P > 0.5\, [F\,G\, \) "\({p1}\)" \(]\)
    \item Always Eventually p1: \(P > 0.5\, [G\,F\, \) "\({p1}\)" \(]\)
    \item (Not p1) Until Eventually p2: \(P > 0.5\, [ \sim \) "\({p1}\)" \(U\,F\,\) "\({p2}\)" \(]\)
    \item (Not p1) Until Always p2: \(P > 0.5\, [ \sim \) "\({p1}\)" \(U\,G\,\) "\({p2}\)" \(]\)
    \item (p1 and p2) Until Eventually p3: \(P > 0.5\, [ \) "\({p1}\)" \(\&\) "\({p2}\)" \(U\,F\,\) "\({p3}\)" \(]\)
\end{enumerate}

The prompts for LVLMs for query matching on QMTP-video are as follows:

\begin{enumerate}
    \item Eventually p1: "Is a \({p1}\) present in any frame of this video?"
    \item Always p1: "Is a \({p1}\) present in all frames of this video?"
    \item p1 Until p2: "Is a \({p2}\) present in any frame of this video and \({p1}\) present in all previous frames?"
    \item Always p1 and Eventually p2: "Is a \({p1}\) present in all frames of this video and is a \({p2}\) present in any frame of this video?"
    \item Always p1 or Eventually p2: "Is a \({p1}\) present in all frames of this video or is a \({p2}\) present in any frame of this video?"
    \item (Not p1) Until p2: "Is a \({p2}\) present in any frame of this video and \({p1}\) absent in all previous frames?"
    \item p1 Until (Not p2): "Is a \({p2}\) absent in any frame of this video and \({p1}\) present in all previous frames?"
    \item Always (p1 and p2): "Are both \({p1}\) and \({p2}\) present in all frames of this video?"
    \item (p1 and p2) Until p3: "Is a \({p3}\) present in any frame of this video and both \({p1}\) and \({p2}\) present in all previous frames?"
    \item p1 Until Always p2: "Starting at some frame in this video, is a \({p2}\) present in all subsequent frames and \({p1}\) present in all previous frames?"
    \item Eventually Always p1: "Starting at some frame in this video, is a \({p1}\) present in all subsequent frames?"
    \item Always Eventually p1: "Starting at any frame in this video, is a \({p1}\) present in some subsequent frame?"
    \item (Not p1) Until Eventually p2: "Starting at some frame in this video, is a \({p2}\) present in some subsequent frame and \({p1}\) absent in all previous frames?"
    \item (Not p1) Until Always p2: "Starting at some frame in this video, is a \({p2}\) present in all subsequent frames and \({p1}\) absent in all previous frames?"
    \item (p1 and p2) Until Eventually p3: "Starting at some frame in this video, is a \({p3}\) present in some subsequent frame and both \({p1}\) and \({p2}\) present in all previous frames?"
\end{enumerate}

The prompts for LALMs for query matching on QMTP-speech are as follows:

\begin{enumerate}
    \item Eventually p1: "Does the speaker's emotion sound \({p1}\) at any time?"
    \item Always p1: "Does the speaker's emotion sound \({p1}\) at all times?"
    \item p1 Until p2: "Does the speaker's emotion sound \({p2}\) at any time and \({p1}\) at all times until then?"
    \item Always p1 and Eventually p2: "Does the speaker's emotion sound \({p1}\) at all times and \({p2}\) at any time?"
    \item Always p1 or Eventually p2: "Does the speaker's emotion sound \({p1}\) at all times or \({p2}\) at any time?"
    \item (Not p1) Until p2: "Does the speaker's emotion sound \({p2}\) at any time and not \({p1}\) at all times until then?"
    \item p1 Until (Not p2): "Does the speaker's emotion sound not \({p2}\) at any time and \({p1}\) at all times until then?"
    \item Always (p1 and p2): "Does the speaker's emotion sound both \({p1}\) and \({p2}\) at all times?"
    \item (p1 and p2) Until p3: "Does the speaker's emotion sound \({p3}\) at any time and both \({p1}\) and \({p2}\) at all times until then?"
    \item p1 Until Always p2: "Starting at some time in this audio clip, does the speaker's emotion sound \({p2}\) at all subsequent times and \({p1}\) at all previous times?"
    \item Eventually Always p1: "Starting at some time in this audio clip, does the speaker's emotion sound \({p1}\) at all subsequent times?"
    \item Always Eventually p1: "Starting at any time in this audio clip, does the speaker's emotion sound \({p1}\) at some subsequent time?"
    \item (Not p1) Until Eventually p2: "Starting at some time in this audio clip, does the speaker's emotion sound \({p2}\) at some subsequent time and not \({p1}\) at all previous times?"
    \item (Not p1) Until Always p2: "Starting at some time in this audio clip, does the speaker's emotion sound \({p2}\) at all subsequent times and not \({p1}\) at all previous times?"
    \item (p1 and p2) Until Eventually p3: "Starting at some time in this audio clip, does the speaker's emotion sound \({p3}\) at some subsequent time and both \({p1}\) and \({p2}\) at all previous times?"
\end{enumerate}

The queries for \texttt{mPLUG} and \texttt{CaptionSim} for retrieval on TP2VR-objects are as follows (\(t_{lo} = 25\) and \(t_{hi} = 50\) here):

\begin{enumerate}
    \item Eventually p1: "A sequence of \(t_{lo}\) to \(t_{hi}\) frames where a \({p1}\) appears at some point."
    \item Always p1: "A sequence of \(t_{lo}\) to \(t_{hi}\) frames where a \({p1}\) is always present."
    \item p1 Until p2: "A sequence of \(t_{lo}\) to \(t_{hi}\) frames where a \({p2}\) is present at some point and a \({p1}\) is present in all frames before that."
    \item Always p1 and Eventually p2: "A sequence of \(t_{lo}\) to \(t_{hi}\) frames where a \({p1}\) is always present and a \({p2}\) appears at some point."
    \item Always p1 or Eventually p2: "A sequence of \(t_{lo}\) to \(t_{hi}\) frames where either a \({p1}\) is always present or a \({p2}\) appears at some point."
    \item (Not p1) Until p2: "A sequence of \(t_{lo}\) to \(t_{hi}\) frames where a \({p2}\) is present at some point and a \({p1}\) is absent in all frames before that."
    \item p1 Until (Not p2): "A sequence of \(t_{lo}\) to \(t_{hi}\) frames where a \({p2}\) is absent at some point and a \({p1}\) is present in all frames before that."
    \item Always (p1 and p2): "A sequence of \(t_{lo}\) to \(t_{hi}\) frames where a \({p1}\) and a \({p2}\) are always present."
    \item (p1 and p2) Until p3: "A sequence of \(t_{lo}\) to \(t_{hi}\) frames where a \({p3}\) is present at some point and both a \({p1}\) and a \({p2}\) are present in all frames before that."
    \item p1 Until Always p2: "A sequence of \(t_{lo}\) to \(t_{hi}\) frames where starting at some point, a \({p2}\) is always present and a \({p1}\) is present in all frames before that."
    \item Eventually Always p1: "A sequence of \(t_{lo}\) to \(t_{hi}\) frames where starting at some point, a \({p1}\) starts being always present."
    \item Always Eventually p1: "A sequence of \(t_{lo}\) to \(t_{hi}\) frames where starting at any point, a \({p1}\) appears in some frames."
    \item (Not p1) Until Eventually p2: "A sequence of \(t_{lo}\) to \(t_{hi}\) frames where starting at some point, a \({p2}\) appears in some frames and a \({p1}\) is absent in all frames before that."
    \item (Not p1) Until Always p2: "A sequence of \(t_{lo}\) to \(t_{hi}\) frames where starting at some point, a \({p2}\) is always present and a \({p1}\) is absent in all frames before that."
    \item (p1 and p2) Until Eventually p3: "A sequence of \(t_{lo}\) to \(t_{hi}\) frames where starting at some point, a \({p3}\) appears at some point and both a \({p1}\) and a \({p2}\) are present in all frames before that."
\end{enumerate}

The queries for \texttt{mPLUG} and \texttt{CaptionSim} for retrieval on TP2VR-actions are as follows (note that \(t_{lo} = t_{hi} = 50\) here):

\begin{enumerate}
    \item Eventually p1: "A sequence of \(t_{lo}\) frames where the action '\({p1}\)' happens at some point."
    \item Always p1: "A sequence of \(t_{lo}\) frames where the action '\({p1}\)' is always happening."
    \item p1 Until p2: "A sequence of \(t_{lo}\) frames where the action '\({p2}\)' happens at some point and the action '\({p1}\)' is happening in all frames before that."
    \item Always p1 and Eventually p2: "A sequence of \(t_{lo}\) frames where the action '\({p1}\)' is always happening and the action '\({p2}\)' happens at some point."
    \item Always p1 or Eventually p2: "A sequence of \(t_{lo}\) frames where either the action '\({p1}\)' is always happening or the action '\({p2}\)' happens at some point."
    \item (Not p1) Until p2: "A sequence of \(t_{lo}\) frames where the action '\({p2}\)' happens at some point and the action '\({p1}\)' is not happening in all frames before that."
    \item p1 Until (Not p2): "A sequence of \(t_{lo}\) frames where the action '\({p2}\)' is not happening at some point and the action '\({p1}\)' is happening in all frames before that."
    \item Always (p1 and p2): "A sequence of \(t_{lo}\) frames where the actions '\({p1}\)' and '\({p2}\)' are always happening."
    \item (p1 and p2) Until p3: "A sequence of \(t_{lo}\) frames where the action '\({p3}\)' happens at some point and both the actions '\({p1}\)' and '\({p2}\)' are happening in all frames before that."
    \item p1 Until Always p2: "A sequence of \(t_{lo}\) frames where starting at some point, the action '\({p2}\)' is always happening and the action '\({p1}\)' is happening in all frames before that."
    \item Eventually Always p1: "A sequence of \(t_{lo}\) frames where starting at some point, the action '\({p1}\)' is always happening."
    \item Always Eventually p1: "A sequence of \(t_{lo}\) frames where starting at any point, the action '\({p1}\)' happens in some frames."
    \item (Not p1) Until Eventually p2: "A sequence of \(t_{lo}\) frames where starting at some point, the action '\({p2}\)' happens in some frames and the action '\({p1}\)' is not happening in all frames before that."
    \item (Not p1) Until Always p2: "A sequence of \(t_{lo}\) frames where starting at some point, the action '\({p2}\)' is always happening and the action '\({p1}\)' is not happening in all frames before that."
    \item (p1 and p2) Until Eventually p3: "A sequence of \(t_{lo}\) frames where starting at some point, the action '\({p3}\)' happens at some point and both the actions '\({p1}\)' and '\({p2}\)' are happening in all frames before that."
\end{enumerate}

%% file: appendix/stl_robustness.tex
\section{More Details on Quantitative Semantics for Temporal Logic}
\label{appendix:semantics}

We now present more details on STL Spatial Robustness, the standard quantitative semantics for Signal Temporal Logic (STL) discussed in Section~\ref{sec:method}. A formula in Signal Temporal Logic can be written as:

\[\varphi := \top \mid \mu \mid \neg \varphi \mid \varphi_1 \land \varphi_2 \mid \varphi_1 \UntilOpSTL{I} \varphi_2\]

where \(\mu = f(s(t)) \geq 0\) is a Lipschitz continuous function over the signal \(s\) and \(I = [t_1, t_2]\) is a time interval, \(t_2 \geq t_1 \geq 0\). The operators \textit{Eventually} and \textit{Always} can be defined as follows:

\[\DiaOpSTL{I} \varphi := \top \UntilOpSTL{I} \varphi\]
\[\BoxOpSTL{I} \varphi := \neg \DiaOpSTL{I} \neg\varphi\]

In the context of the STOP problem, \(s(t) = \hat{y}(X, \cdot, t)\) and \(f_c(s(t)) = \hat{y}(X, c, t) - \tau\). The query "car until pedestrian" can be written as \(\mu_{car}(s(t)) \UntilOpSTL{I} \mu_{pedestrian}(s(t))\) where \(I = [0, T-1]\) and \(\mu_{car}(s(t)) = \hat{y}(X, {car}, t) - \tau \geq 0\) (\(\mu_{pedestrian}\) is defined similarly).

The STL quantitative semantics, also called \textit{robustness} \(\rho\)~\cite{FAINEKOS20094262}, is defined as follows to indicate how much a signal satisfies or violates the formula:

\[\rho(\top, s, t) := \rho_\top\]
\[\rho(\mu, s, t) := f(s(t))\]
\[\rho(\neg \varphi, s, t) := -\rho(\varphi, s, t)\]
\[\rho(\varphi_1 \land \varphi_2, s, t) := \min(\rho(\varphi_1, s, t), \rho(\varphi_2, s, t))\]
\[\rho(\varphi_1 \lor \varphi_2, s, t) := \max(\rho(\varphi_1, s, t), \rho(\varphi_2, s, t))\]
\[\rho(\varphi_1 \UntilOpSTL{I} \varphi_2, s, t) := \sup_{t' \in\, t+I}(\min\{\rho(\varphi_2, s, t'), \inf_{t'' \in\, [t, t')}\rho(\varphi_1, s, t'')\})\]
\[\rho(\BoxOpSTL{I}\,\varphi, s, t) := \inf_{t' \in\,[t+I]} \rho(\varphi, s, t')\]
\[\rho(\DiaOpSTL{I}\,\varphi, s, t) := \sup_{t' \in\,[t+I]} \rho(\varphi, s, t')\]

where, \(\rho_\top\) is the maximum robustness, i.e., \(\rho_\top = b -\tau\) for the STOP problem where \(b = \max(\hat{y}(\cdot, \cdot, \cdot))\).

We argue that LogSTOP offers advantages over such semantics in the context of the STOP problem.
This is primarily because the traditional robustness measure is defined using \(\max\) and \(\min\) functions over temporal and logical formulae. 
The measure, hence, only reflects the most violating or most satisfying timestep in the sequence. For example, consider assigning confidence scores to the property "Always car" in two different scenarios:
\[\hat{y}_1(X, {car}, t\in[1,3]) = [0.9, 0.9, 0.1]\]
\[\hat{y}_2(X, {car}, t\in[1,3]) = [0.1, 0.1, 0.1]\]
Ideally, the confidence score for "Always car" should follow the order: 
\(\hat{y}_1(X, \BoxOp{I}{car}, 1, 3) > \hat{y}_2(X, \BoxOp{I}{car}, 1, 3)\). 
STL Spatial Robustness, however, would assign the same robustness to both sequences for any \(\tau > 0.1\) since the most violating local evidence is \(0.1\) in either case. This makes the robustness measure unsuitable for downstream applications that require such ordering: for example, ranking / search. 

\noindent{\textbf{An example with Boolean operators. }} For example, consider assigning confidence scores to the property "car and pedestrian" at \(t=1\) in two different scenarios:
\[\hat{y}_1(X, {car}, t=1) = 0.9, \hat{y}_1(X, {pedestrian}, t=1) = 0.6\]
\[\hat{y}_2(X, {car}, t=1) = 0.6, \hat{y}_2(X, {pedestrian}, t=1) = 0.6\]
Ideally, the confidence scores for "car and pedestrian" should follow the order: 
\(\hat{y}_1(X,{car} \land {pedestrian}, t=1,\cdot) > \hat{y}_2(X,{car} \land {pedestrian}, t=1, \cdot)\) since \(\hat{y}_1(X,{car}, t=1) > \hat{y}_2(X,{car}, t=1)\). The robustness for both the cases is the same, i.e., \(0.6 - 0.5\) for \(\tau=0.5\), because of the \(\min\) semantics for the Boolean \textit{and} operator. The LogSTOPs for the two cases are \(\log(0.9) + \log(0.6)\) and  \(\log(0.6) + \log(0.6)\) respectively, which reflect the expected order.

\noindent{\textbf{An example with the Until operator. }} Consider assigning confidence scores to the property "car Until pedestrian" in two different scenarios:
\[\hat{y}_1(X,{car}, t\in[1,3]) = [0.6, 0.6, 0.6]\]
\[\hat{y}_1(X,{pedestrian}, t\in[1,3]) = [0.4, 0.4, 0.9]\]
and,
\[\hat{y}_2(X,{car}, t\in[1,3]) = [0.6, 0.6, 0.6]\]
\[\hat{y}_2(X,{pedestrian}, t\in[1,3]) = [0.4, 0.4, 0.6]\]

Note that the only difference between the two scenarios is the score for "pedestrian" at \(t = 3\) (a high score of \(0.9\) for the first scenario and a lower score of \(0.6\) for the second scenario) . The robustness for the two cases is the same because of the \(\min\) semantics within the \textit{Until} operator. LogSTOP assigns a higher score for the first scenario because of the difference at \(t=3\).

%% file: appendix/proofs.tex
\section{Proofs}

\subsection{Proof for Proposition~\ref{thm:monotonicity} (Monotonicity of LogSTOP in post-processed local evidence)}
\label{proof:monotonicity}

\begin{proposition}[Monotonicity of LogSTOP in post-processed local evidence]
\label{thm:monotonicity}
Fix a temporal property \(\varphi\), an evaluation interval \([t_s,t_e]\), and post-processed local evidence values \(\tilde{y}(X,c,t)\). Suppose that every post-processed local evidence variable \(\tilde{y}(X,c,t)\), with \(t \in [t_s,t_e]\), that influences the evaluated score \(\hat{y}(X,\varphi,t_s,t_e)\) has a unique effective polarity, either positive or negative. Let \(X\) and \(X'\) be two sequences whose post-processed local evidence values satisfy
\[
\tilde{y}(X',c,t)\ge \tilde{y}(X,c,t)
\quad\text{for every }(c,t)\text{ with positive effective polarity,}
\]
and
\[
\tilde{y}(X',c,t)\le \tilde{y}(X,c,t)
\quad\text{for every }(c,t)\text{ with negative effective polarity.}
\]
Then
\[
\hat{y}(X',\varphi,t_s,t_e)\ge \hat{y}(X,\varphi,t_s,t_e).
\]
\end{proposition}

\begin{proof}
    We first record the monotonicity properties of the primitive operations used by LogSTOP.

    \noindent\textbf{Primitive monotonicity facts.}

    Throughout this proof, let \(\tilde{y}(X,c,t)\in[0,1]\) denote the post-processed probability-space local evidence supplied to LogSTOP for local property \(c\) at timestep \(t\). The proof treats these post-processed values as fixed inputs to the temporal aggregation; they may be obtained from raw, calibrated, or smoothed local predictor scores.
    The local-property base case returns the log of this post-processed evidence, with the usual extended-real convention \(\log 0=-\infty\).

    \textbf{(i)} The function
    \[
    \ell(x)=\log x, \qquad x\in(0,1],
    \]
    is monotone nondecreasing, and extends monotonically to \(x=0\) by \(\ell(0)=-\infty\).

    \textbf{(ii)} The function
    \[
    h(x)=\log(1-x), \qquad x\in[0,1),
    \]
    is monotone nonincreasing, and extends monotonically to \(x=1\) by \(h(1)=-\infty\).

    \textbf{(iii)} Addition is monotone nondecreasing in each argument.

    \textbf{(iv)} The function
    \[
    g(a,b)=\log(e^a+e^b-e^{a+b}), \qquad a,b\in[-\infty,0],
    \]
    is monotone nondecreasing in each argument.

    We now prove the proposition by induction on the structure of \(\varphi\).

    \textbf{Base cases.}

    \emph{Case 1: \(\varphi = c\).}
    Every post-processed local evidence variable \(\tilde{y}(X,c,t')\) that enters the score for \(c\) has positive effective polarity. Hence, by hypothesis,
    \[
    \tilde{y}(X',c,t')\ge \tilde{y}(X,c,t')
    \]
    for every such \(t'\). The local-property base case evaluates to the log of the post-processed evidence at the current timestep:
    \[
    \hat{y}(X',c,t_s,t_e)=\log \tilde{y}(X',c,t_s)\ge \log \tilde{y}(X,c,t_s)=\hat{y}(X,c,t_s,t_e).
    \]

    \emph{Case 2: \(\varphi = \neg c\).}
    The post-processed local evidence variable \(\tilde{y}(X,c,t_s)\) has negative effective polarity in \(\hat{y}(X,\neg c,t_s,t_e)\): increasing it decreases the negated score. Hence, by hypothesis,
    \[
    \tilde{y}(X',c,t_s)\le \tilde{y}(X,c,t_s).
    \]
    By definition,
    \[
    \hat{y}(X,\neg c,t_s,t_e)=h(\tilde{y}(X,c,t_s)).
    \]
    Since \(h\) is monotone nonincreasing on \([0,1)\),
    \[
    \hat{y}(X',\neg c,t_s,t_e)\ge \hat{y}(X,\neg c,t_s,t_e).
    \]

    \textbf{Inductive step.}
    Assume that the claim holds for all strict sub-properties of \(\varphi\).

    \emph{Case 3: \(\varphi = \varphi_1 \land \varphi_2\).}
    By the induction hypothesis,
    \[
    \hat{y}(X',\varphi_i,t_s,t_e)\ge \hat{y}(X,\varphi_i,t_s,t_e), \qquad i\in\{1,2\}.
    \]
    Since conjunction is additive,
    \[
    \hat{y}(X',\varphi_1 \land \varphi_2,t_s,t_e)
    =
    \hat{y}(X',\varphi_1,t_s,t_e)
    +
    \hat{y}(X',\varphi_2,t_s,t_e)
    \]
    \[
    \ge
    \hat{y}(X,\varphi_1,t_s,t_e)
    +
    \hat{y}(X,\varphi_2,t_s,t_e)
    =
    \hat{y}(X,\varphi_1 \land \varphi_2,t_s,t_e).
    \]

    \emph{Case 4: \(\varphi = \varphi_1 \lor \varphi_2\).}
    By the induction hypothesis,
    \[
    \hat{y}(X',\varphi_i,t_s,t_e)\ge \hat{y}(X,\varphi_i,t_s,t_e), \qquad i\in\{1,2\}.
    \]
    Since disjunction is computed using \(g(a,b)=\log(e^a+e^b-e^{a+b})\), and \(g\) is monotone nondecreasing in each argument,
    \[
    \hat{y}(X',\varphi_1 \lor \varphi_2,t_s,t_e)
    \ge
    \hat{y}(X,\varphi_1 \lor \varphi_2,t_s,t_e).
    \]

    \emph{Case 5: \(\varphi = \bigcirc\, \varphi'\).}
    If \(t_s=t_e\), then both sides are \(-\infty\), so the claim is immediate. Otherwise,
    \[
    \hat{y}(X,\bigcirc\,\varphi',t_s,t_e)
    =
    \hat{y}(X,\varphi',t_s+1,t_e).
    \]
    The claim then follows directly from the induction hypothesis applied at timestep \(t_s+1\).

    \emph{Case 6: \(\varphi = \BoxOp{I}\,\varphi'\).}
    If \(t_s=t_e\), then
    \[
    \hat{y}(X,\BoxOp{I}\,\varphi',t_s,t_e)
    =
    \hat{y}(X,\varphi',t_s,t_e),
    \]
    and the claim follows from the induction hypothesis. Otherwise, by the recursive definition,
    \[
    \hat{y}(X,\BoxOp{I}\,\varphi',t_s,t_e)
    =
    \hat{y}(X,\varphi' \land \bigcirc\,\BoxOp{I}\,\varphi',t_s,t_e).
    \]
    The claim then follows from the conjunction case, the Next case, and the induction hypothesis.

    \emph{Case 7: \(\varphi = \varphi_1 \UntilOp{I} \varphi_2\).}
    If \(t_s=t_e\), then
    \[
    \hat{y}(X,\varphi_1 \UntilOp{I} \varphi_2,t_s,t_e)
    =
    \hat{y}(X,\varphi_2,t_s,t_e),
    \]
    and the claim follows from the induction hypothesis. Otherwise, by the recursive definition,
    \[
    \hat{y}(X,\varphi_1 \UntilOp{I} \varphi_2,t_s,t_e)
    =
    \hat{y}(X,\varphi_2 \lor (\varphi_1 \land \bigcirc(\varphi_1 \UntilOp{I}\varphi_2)),t_s,t_e).
    \]
    The claim therefore follows from the disjunction, conjunction, and Next cases together with the induction hypothesis.

    Thus, in all cases,
    \[
    \hat{y}(X',\varphi,t_s,t_e)\ge \hat{y}(X,\varphi,t_s,t_e).
    \]
    \qed
\end{proof}

\subsection{Proof for the complexity of LogSTOP}
\label{proof:complexity}
\begin{proof}
    Fix an evaluation interval \([t_s,t_e]\). The recursive definition in Algorithm~\ref{alg:LogSTOP_complete} can be evaluated using dynamic programming (or memoization) over sub-properties and timesteps within this fixed interval. We define a table \(M\) of size \(|\varphi| \times (t_e-t_s+1)\), where entry \(M[\psi, t]\) stores the score for sub-property \(\psi\) evaluated at timestep \(t\) using the post-processed local evidence \(\tilde y\) precomputed once over \([t_s,t_e]\).

	    For each local property \(c\) that appears in \(\varphi\), we first precompute prefix sums
	    \[
		    P_c(t)=\sum_{u=1}^{t} \hat{y}(X,c,u).
	    \]
	    These prefix sums allow the smoothed probability-space local evidence at timestep \(t\),
	    \[
	    \frac{1}{|W_t|}\sum_{u \in W_t} \hat{y}(X,c,u),
	    \]
		    to be recovered in \(\mathcal{O}(1)\) time from the interval sum \(P_c(r_t)-P_c(\ell_t-1)\), where \(W_t=[\ell_t,r_t]\) is the truncated left-right window determined by \([t_s,t_e]\) and the local-property-specific radius \(w_c\).

    \noindent\textbf{Space Complexity:}
    The number of unique sub-properties is bounded by \(|\varphi|\), and the number of reachable timesteps is at most \(T\). Thus, the memoization table requires \(\mathcal{O}(T \cdot |\varphi|)\) space. The prefix sums for all local properties appearing in \(\varphi\) require an additional \(\mathcal{O}(T \cdot |\varphi|)\) space in the worst case, so the overall space complexity remains \(\mathcal{O}(T \cdot |\varphi|)\).

    \noindent\textbf{Time Complexity:}
    We analyze the cost to populate each entry in the table, assuming dependent sub-problems have already been computed and can be looked up in \(\mathcal{O}(1)\) time:
    \begin{itemize}
        \item \textbf{Local properties (\(\varphi = c\)):} After the smoothed local evidence values are precomputed in Line~\ref{alg:smoothing_op}, the local-property case takes the log of a table lookup in \(\mathcal{O}(1)\) time.
        \item \textbf{Boolean properties:} 
        For \(\varphi = \neg c\), the operation is constant time after the smoothed local evidence for \(c\) is available (Line~\ref{alg:not_op}).
        For \(\varphi = \varphi_1 \land \varphi_2\), we sum two looked-up values (Line~\ref{alg:and_op}), taking \(\mathcal{O}(1)\).
        For \(\varphi = \varphi_1 \lor \varphi_2\), we combine two looked-up values using the direct disjunction rule in Line~\ref{alg:or_op}, again taking \(\mathcal{O}(1)\).
        \item \textbf{Next (\(\bigcirc \varphi'\)):} Line~\ref{alg:next_op} involves a single recursive call with \(t + 1\), which corresponds to a table lookup \(M[\varphi', t+1]\). This is \(\mathcal{O}(1)\).
        \item \textbf{Other temporal properties (\(\Box, \mathcal{U}\)):}
        The algorithm expands these properties into fixed logical unrollings involving \(\land\), \(\lor\), and \(\bigcirc\).
        For \(\Box \varphi'\), the recursion expands to \(\varphi' \land \bigcirc \Box \varphi'\) (Line~\ref{alg:always_op}).
        For \(\mathcal{U}\), the final timestep reduces to \(\varphi_2\), while earlier timesteps expand to \(\varphi_2 \lor (\varphi_1 \land \bigcirc(\varphi_1 \UntilOp{I} \varphi_2))\) (Line~\ref{alg:until_op}).
        Since these expansions use a constant number of Boolean and Next operations, determining the value at \(t\) relies only on values at \(t\) and \(t+1\), and therefore takes \(\mathcal{O}(1)\) time per table entry.
    \end{itemize}
    
    The prefix sums for all local properties in \(\varphi\) can be computed in \(\mathcal{O}(T \cdot |\varphi|)\) time in the worst case. The dynamic-programming table contains \(\mathcal{O}(T \cdot |\varphi|)\) entries, and each entry can be filled in \(\mathcal{O}(1)\) time. Thus, the total time complexity is \(\mathcal{O}(T \cdot |\varphi|)\).
    \qed
\end{proof}

\begin{proposition}[Consistency with respect to the standard Boolean semantics for LTL]
\label{thm:consistency}
Let \(y(X,c,t)\in\{0,1\}\) denote the true Boolean labeling function and let \(\hat{y}(X,c,t)=y(X,c,t)\) denote the perfect local property predictor for all local properties \(c\) and timesteps \(t\). Then, for any finite sequence \(X\) and temporal property \(\varphi\),
\[
\hat{y}(X,\varphi,t,T,\{w_c=0\}) > -\infty
\iff
(X,t)\models\varphi,
\]
where \(\models\) denotes satisfaction under standard Boolean LTL semantics on finite traces.
\end{proposition}

Although LogSTOP is not intended to represent the exact probability that a sequence satisfies a temporal property, Proposition~\ref{thm:consistency} shows that it recovers the standard Boolean notion of temporal satisfaction in the noiseless binary setting, i.e., when the local property predictors are perfect and no smoothing is applied.

\subsection{Proof for Proposition~\ref{thm:consistency} (Consistency of LogSTOP with respect to the standard Boolean semantics for LTL)}
\label{proof:consistency}
\begin{proof}
    We prove this by structural induction on the formula \(\varphi\). Since this proposition concerns the noiseless binary setting with no smoothing, we use \(\hat{y}(X, \varphi, t, T)\) to denote \(\hat{y}(X, \varphi, t, T, \{w_c=0\})\). 
    For any starting time step \(t\):
    
	    \noindent\textbf{Base Case (Local property \(\varphi = c\)):}
	    Since no smoothing is applied, \(\tilde{y}(X,c,t)=\hat{y}(X,c,t)\), and the local-property case returns \(\log \tilde{y}(X,c,t)\).
    If \((X, t) \models c\), then \(y(X, c, t) = 1\). The algorithm returns \(\log(1) = 0 > -\infty\).
    If \((X, t) \not\models c\), then \(y(X, c, t) = 0\). The algorithm returns \(\log(0) = -\infty\).
	    Thus, consistency holds.

    \noindent\textbf{Base Cases (\(\varphi=\top\) and \(\varphi=\bot\)):}
    Algorithm~\ref{alg:LogSTOP_complete} returns \(0>-\infty\) for \(\top\) and \(-\infty\) for \(\bot\), matching Boolean finite-trace semantics.

    \noindent\textbf{Inductive Step:} Assume consistency holds for sub-formulas \(\varphi_1, \varphi_2\) (IH).

    \begin{enumerate}
        \item \textbf{Negation of a local property (\(\varphi = \neg c\)):}
        With no smoothing and perfect binary local predictions, \(\tilde{y}(X,c,t)=y(X,c,t)\), so \(\hat{y}(X, \varphi, t, T) = \log(1 - y(X,c,t))\).
        \begin{itemize}
            \item If \((X, t) \models \neg c\), then \((X, t) \not\models c\). By the local-property base case, \(\hat{y}(X,c,t,T) = -\infty\), so \(e^{\hat{y}(X,c,t,T)} = 0\).
            Result: \(\log(1 - 0) = 0 > -\infty\).
            \item If \((X, t) \not\models \neg c\), then \((X, t) \models c\). By the local-property base case, \(\hat{y}(X,c,t,T) = 0\), so \(e^{\hat{y}(X,c,t,T)} = 1\).
            Result: \(\log(1 - 1) = -\infty\).
        \end{itemize}
        
        \item \textbf{Conjunction (\(\varphi = \varphi_1 \land \varphi_2\)):}
        \(\hat{y}(X, \varphi, t, T) = \hat{y}_1(X, \varphi_1, t, T) + \hat{y}_2(X, \varphi_2, t, T)\).
        \begin{itemize}
            \item \((X, t) \models \varphi_1 \land \varphi_2\) iff \((X, t) \models \varphi_1\) AND \((X, t) \models \varphi_2\).
        By IH, this corresponds to \(\hat{y}_1(X, \varphi_1, t, T) = 0\) AND \(\hat{y}_2(X, \varphi_2, t, T) = 0\).
        Sum: \(0 + 0 = 0 > -\infty\).
        \item If either is False (\(-\infty\)), the sum is \(-\infty\).
        \end{itemize}

        \item \textbf{Next (\(\varphi = \bigcirc \varphi_1\)):}
        \(\hat{y}(X, \varphi, t, T) = \hat{y}(X, \varphi_1, t+1, T)\).
        \begin{itemize}
            \item If \(t + 1 > T\), it returns \(-\infty\) , matching finite trace semantics~\cite{ltlf} where \(\bigcirc\) fails at the end of a trace.
            \item Otherwise, by IH, the result matches the truth value of \(\varphi_1\) at \(t+1\).
        \end{itemize}
        
        \item \textbf{Disjunction (\(\varphi = \varphi_1 \lor \varphi_2\)):}
        Let \(p_i=e^{\hat{y}(X,\varphi_i,t,T)}\) for \(i\in\{1,2\}\). The disjunction rule returns \(\log(p_1+p_2-p_1p_2)\). By the induction hypothesis, \(p_i\in\{0,1\}\) indicates whether \((X,t)\models\varphi_i\). Therefore \(p_1+p_2-p_1p_2=1\) iff at least one of \(\varphi_1\) or \(\varphi_2\) holds at \(t\), and is \(0\) otherwise, matching Boolean disjunction.

        \item \textbf{Always (\(\varphi = \BoxOp{} \varphi_1\)):}
        If \(t=T\), Algorithm~\ref{alg:LogSTOP_complete} returns \(\hat{y}(X,\varphi_1,T,T,0)\), which is consistent by the induction hypothesis. If \(t<T\), Algorithm~\ref{alg:LogSTOP_complete} evaluates \(\varphi_1 \land \bigcirc\BoxOp{}\varphi_1\). By the induction hypothesis for \(\varphi_1\), the Next case, and induction over the remaining trace length for \(\BoxOp{}\varphi_1\), this score is finite iff \(\varphi_1\) holds at every timestep from \(t\) to \(T\), matching finite-trace Boolean semantics.

        \item \textbf{Until (\(\varphi = \varphi_1 \UntilOp{} \varphi_2\)):}
        If \(t=T\), Algorithm~\ref{alg:LogSTOP_complete} returns \(\hat{y}(X,\varphi_2,T,T,0)\), so the score is finite iff \(\varphi_2\) holds at the final timestep. This matches strong Until on finite traces. If \(t<T\), Algorithm~\ref{alg:LogSTOP_complete} evaluates \(\varphi_2 \lor (\varphi_1 \land \bigcirc(\varphi_1 \UntilOp{} \varphi_2))\). By the induction hypotheses for \(\varphi_1\) and \(\varphi_2\), the Next case, and induction over the remaining trace length for \(\varphi_1 \UntilOp{} \varphi_2\), this score is finite iff either \(\varphi_2\) holds at \(t\), or \(\varphi_1\) holds at \(t\) and \(\varphi_1 \UntilOp{} \varphi_2\) holds from \(t+1\). This is exactly the finite-trace Boolean semantics for Until.
    \end{enumerate}
    \qed
\end{proof}

\begin{proposition}[Complexity of Ranked Retrieval]
\label{thm:retrieval_complexity}
Given a database \(\mathcal{D}\) of \(N\) sequences with maximum length \(T\), a temporal property \(\varphi\), target subsequence length range \([T_{lo}, T_{hi}]\), a retrieval count \(k\), and maximum smoothing radius \(w_{\max}\), Algorithm~\ref{alg:logstop-retrieval} computes the ranked list in \(\mathcal{O}(N \cdot ((T-T_{lo}+1) \cdot T_{hi} \cdot (1+w_{\max}) \cdot |\varphi| + \log k))\) time.
\end{proposition}

\subsection{Proof for Proposition~\ref{thm:retrieval_complexity} (Complexity of Ranked Retrieval with LogSTOP)}
\label{proof:retrieval_complexity}

\begin{proof}
    We analyze the computational complexity of Algorithm~\ref{alg:logstop-retrieval} by examining the nested structure of the algorithm and the ranking mechanism:

    \begin{enumerate}
        \item \textbf{Sequence Iteration (Line~\ref{alg:retrieval_outer}):} The outer loop iterates over all \(N\) sequences in the database.
        
        \item \textbf{Sliding Window Loop (Line~\ref{alg:retrieval_inner}):} For a given sequence \(X_i\) of length \(T\), the loop iterates through endpoint times \(t_e \in [T_{lo}, T]\), executing \(T - T_{lo} + 1\) times.
        
        \item \textbf{LogSTOP Computation (Line~\ref{alg:retrieval_logstop}):} Inside the loop, the algorithm invokes LogSTOP on candidate windows ending at \(t_e\), with lengths bounded by \(T_{hi}\). The dynamic-programming recursion over temporal subproperties contributes \(\mathcal{O}(T_{hi}\cdot|\varphi|)\) time. Since smoothing is recomputed with respect to each candidate window, only the boundary timesteps within \(w_{\max}\) of the candidate start or end can differ from a globally smoothed sequence, adding at most a factor of \(w_{\max}\) across the \(O(T_{hi})\) candidate starts for this endpoint. Thus, the per-endpoint cost is \(\mathcal{O}(T_{hi}\cdot (1+w_{\max})\cdot|\varphi|)\). In our experiments, \(w_{\max}\) is a small hyperparameter, with \(w_{\max}\ll T\).
        
        \item \textbf{Maximization (Line~\ref{alg:retrieval_max}):} The algorithm scans valid start times \(t \in [t_{\min}, t_{\max}]\) to find the best subsequence ending at \(t_e\). This requires iterating over at most \(T_{hi}\) elements, taking \(\mathcal{O}(T_{hi})\) time.

        \item \textbf{Top-k Maintenance (Line~\ref{alg:retrieval_rank}):} To maintain the top-\(k\) scores efficiently, we employ a min-heap of size \(k\). For each sequence \(X_i\), we compare its final score \(s_i\) with the heap's root (minimum). If \(s_i\) is larger, we replace the root and sift down, taking \(\mathcal{O}(\log k)\) time.
    \end{enumerate}

    \noindent\textbf{Total Time Complexity:}
    The time per sequence combines the scoring loop and the heap update:
    \[
    \begin{aligned}
    &\underbrace{\mathcal{O}(T-T_{lo}+1) \times \mathcal{O}(T_{hi} \cdot (1+w_{\max}) \cdot |\varphi|)}_{\text{Scoring}}
    + \underbrace{\mathcal{O}(\log k)}_{\text{Ranking}} \\
    &\qquad =
    \mathcal{O}((T-T_{lo}+1) \cdot T_{hi} \cdot (1+w_{\max}) \cdot |\varphi| + \log k).
    \end{aligned}
    \]
    Summing over the database of \(N\) sequences:
    \[ \text{Total Time} = \mathcal{O}(N \cdot ((T-T_{lo}+1) \cdot T_{hi} \cdot (1+w_{\max}) \cdot |\varphi| + \log k)) \]
    
    \noindent\textbf{Space Complexity:}
    The memory is dominated by the dynamic programming table for the current window (\(\mathcal{O}(T_{hi} \cdot |\varphi|)\)) and the storage for the top-\(k\) heap (\(\mathcal{O}(k)\)).
    \qed
\end{proof}

%% file: appendix/algorithm_details.tex
\section{More details on Algorithm 1}
\label{appendix:alg1_details}

Algorithm~\ref{alg:LogSTOP_complete} describes how LogSTOP is computed for a sequence \(X[t:T]\) with respect to temporal property \(\varphi\), given start and end timesteps \(t\) and \(T\), and local-property-specific smoothing radii \(\{w_c\}\). The algorithm first precomputes smoothed local evidence over the target evaluation interval and then recursively aggregates this evidence in log-space according to the temporal property. Here we discuss the intuition behind the operators, along with some examples:

\noindent{\textbf{Average smoothing of confidence scores of local properties.}} Firstly, LogSTOP smooths local property scores using temporal windows whose radii may differ across local properties. For local property \(c\) at timestep \(t\), Algorithm~\ref{alg:LogSTOP_complete} averages the local property scores \(\hat{y}(X,c,t')\) over the available timesteps \(t' \in [\max\{t-w_c,t_s\}, \min\{t+w_c,t_e\}]\) (line~\ref{alg:smoothing_op}). Thus, boundary timesteps use whatever neighbors are available; for example, the first prediction only uses the current timestep and available right neighbors. When \(w_c=0\), no smoothing is applied for class \(c\). For example, given a sequence of scores for object "car",
\[\hat{y}(X,{car}, t \in [1, 3]) = [0.9, 0.1, 0.9]\] 
and smoothing radius \(w_{\mathsf{car}} = 1\), the smoothed local evidence at \(t=2\) is
\[
(0.9 + 0.1 + 0.9)/3=0.63.
\]
In this example, note that the score \(0.1\) is likely incorrect since the car cannot momentarily disappear. Average smoothing reduces the impact of this incorrect local prediction, hence capturing the assumption that confidence scores for persistent local properties should not drastically change within a small temporal neighborhood. 
The temporal operators themselves still advance one timestep at a time; smoothing only changes the local property scores that enter the recursive composition.

\noindent{\textbf{Handling of logical operators (\(\neg, \land, \lor\))}.} 
The LogSTOP for \(\neg c\) at timestep \(t\) is intuitively high when the post-processed evidence for local property \(c\) is low (line~\ref{alg:not_op}). For example, given high post-processed evidence \(\tilde{y}(X,{car}, t) = 0.9\), the score for "not car" \(\hat{y}(X,\neg {car}, t, T, \{w_c\}) = \log(1 - 0.9) = \log(0.1)\) is low.

The LogSTOP for \(\varphi_1 \land \varphi_2\) at timestep \(t\) is high only when the scores for both \(\varphi_1\) and \(\varphi_2\) are high (line~\ref{alg:and_op}). For example, given high scores \(\hat{y}(X,{car}, t, T, \{w_c\}) = \log(0.9)\) and \(\hat{y}(X,{pedestrian}, t, T, \{w_c\}) = \log(0.9)\), the score for "car and pedestrian" \(\hat{y}(X,{car} \land {pedestrian}, t, T, \{w_c\}) = \log(0.9) + \log(0.9) = \log(0.81)\). However, if either of the scores are low, e.g., if \(\hat{y}(X,{pedestrian}, t, T, \{w_c\}) = \log(0.1)\), the score drops significantly to \(\hat{y}(X,{car} \land {pedestrian}, t, T, \{w_c\}) = \log(0.9) + \log(0.1) = \log(0.09)\). The LogSTOP for \(\varphi_1 \lor \varphi_2\) directly aggregates support for either sub-property as \(\log(p_1+p_2-p_1p_2)\), where \(p_i=e^{\hat{y}(X,\varphi_i,t,T,\{w_c\})}\) (line~\ref{alg:or_op}). This is intuitively only low when both the scores are low. 

Conjunction and disjunction are defined not only over local properties \(c\) as in the examples above, but over any temporal properties \(\varphi_1, \varphi_2\). Hence, for temporal property \(\varphi\) = "(car or truck) and (not pedestrian)", the score is high only if it is high for both \(\varphi_1\) = "(car or truck)" and \(\varphi_2\) = "(not pedestrian)". The scores for \(\varphi_1, \varphi_2\) can be recursively computed.

\noindent{\textbf{Handling of temporal operators (\(\bigcirc, \BoxOp{I}, \UntilOp{I}\)).} }
As discussed above, the property Next \(\varphi\) (\(\bigcirc\, \varphi\)) evaluates whether \(\varphi\) is expressed starting at the next timestep \(t + 1\) (line~\ref{alg:next_op}). At the final timestep of a finite trace, Next returns \(-\infty\), matching the standard strong-next semantics.
The property Always \(\varphi\) (\(\BoxOp{I}\, \varphi\)) is interpreted here as a "temporal and" operator over the sequence (line~\ref{alg:always_op}). At the final timestep, \(\BoxOp{I}\, \varphi\) reduces to \(\varphi\). At earlier timesteps, \(\BoxOp{I}\, \varphi\) can be written recursively as \(\varphi \land \bigcirc\, \BoxOp{I}\, \varphi\): the property \(\varphi\) is expressed by \(x_t\) and always after by \([x_{t+1}, \ldots]\) (\(\bigcirc\, {\BoxOp{I}\,\varphi}\)). Similar to the logical \(\land\) operator, the score is high only if it is high for all timesteps of the sequence. The computation in \(\log\) space is beneficial to prevent any underflow here with fixed precision.

The property \(\varphi_1\) Until \(\varphi_2\) (\(\varphi = \varphi_1 \UntilOp{I} \varphi_2\)) reduces to \(\varphi_2\) at the final timestep. At earlier timesteps, it can be written recursively as \(\varphi = \varphi_2 \lor (\varphi_1 \land \bigcirc\,(\varphi_1 \UntilOp{I} \varphi_2))\) (line~\ref{alg:until_op}). This informally translates to evaluating whether either (1) \(\varphi_2\) is expressed by \(x_t\), or (2) \(\varphi_1\) is expressed instead and \(\varphi = \varphi_1 \UntilOp{I} \varphi_2\) is expressed  by \([x_{t+1}, \ldots]\).

%% file: appendix/retrieval_algorithm.tex
\section{Ranked Retrieval Algorithm}
\label{appendix:retrieval_algorithm}

\begin{algorithm}[H]
\caption{LogSTOP for Ranked Retrieval}
\label{alg:logstop-retrieval}
\begin{algorithmic}[1]
\normalsize
\Require Database \(\mathcal{D} = \{X_1, X_2, \ldots, X_N\}\), temporal property \(\varphi\), target subsequence length range \([T_{lo}, T_{hi}]\), number of retrievals \(k\), smoothing radii \(\{w_c\}_{c\in C}\)
\Ensure Ranked list \(\mathcal{R}\) of top-\(k\) sequences

\State \(\mathcal{R} \gets \mathrm{MinHeap}()\)
\For{\(X_i \in \mathcal{D}\)}
    \label{alg:retrieval_outer}
    \State \(s_i \leftarrow -\inf\)
    \State \(T \gets X_i{.length()}\)
    \For{\(t_e \in [T_{lo}, T]\)}
        \label{alg:retrieval_inner}
        \State \(t_{\min} \gets \max\{1, t_e - T_{hi} + 1\}\)
        \State \(t_{\max} \gets t_e - T_{lo} + 1\)
        \State Compute \(\hat{y}(X_i, \varphi, t, t_e, \{w_c\})\) for all \(t \in [t_{\min}, t_e]\) using Algorithm~\ref{alg:LogSTOP_complete}
        \label{alg:retrieval_logstop}
        \State \(s_{i, t_e} \gets \max\{\hat{y}(X_i, \varphi, t, t_e, \{w_c\}) \text{ for } t \in [t_{\min}, t_{\max}]\}\)
        \label{alg:retrieval_max}
        \State \(s_i \leftarrow \max(s_i, s_{i, t_e})\)
    \EndFor
    \State \(\mathcal{R}\text{.push\_topk}((i, s_i), k)\)
    \label{alg:retrieval_rank}
\EndFor
\Return sequences in \(\mathcal{R}\) in decreasing order of \(s_i\)
\end{algorithmic}
\end{algorithm}

%% file: appendix/llm_ablations_prompts.tex
\section{Prompts to LLMs for matching and translation}
\label{appendix:llm_ablation_prompts}

\input{graphics/charts/llm_scoring_prompts}

\input{graphics/charts/llm_parsing_prompts}

%% file: graphics/charts/llm_scoring_prompts.tex
\begin{llmprompt}[label={prompt:scoring-nl}]{Query matching with an NL query}
\noindent{\normalfont\bfseries System message}\par\smallskip
You are given a query that you need to match against a video. The query is provided as a natural-language query over objects in the video (car, person, bus, truck, bicycle, and motorcycle). You don't have access to the raw video, instead you are given frame-level detection scores for these objects from the YOLO object detector. Use this information to decide whether the video matches the temporal query. All temporal queries are evaluated from the first frame of the video; when a query mentions previous frames, interpret this as the prefix from frame 1 up to, but not including, the chosen witness/anchor frame. The provided detector thresholds are calibration reference scores r\_c(0.5): for each object class c, this is the detector score level at which the empirical probability of c being present is 50\% on a held-out calibration split. Use them as calibration context for interpreting the score scale. You may reason briefly in text, but do not write code or markdown. End your response with a final line beginning with FINAL\_JSON: followed by \{"answer": true/false, "confidence": a number from 0 to 1\}.\par
\medskip
\noindent{\normalfont\bfseries User message template}\par\smallskip
Task: Decide whether the video matches the temporal query.\par
\smallskip
Query:\par
\textless{}NL\_QUERY\textgreater{}\par
\smallskip
Detector calibration reference scores r\_c(0.5):\par
\textless{}CALIBRATION\_REFERENCE\_SCORES\textgreater{}\par
\smallskip
Detection scores:\par
\textless{}FRAME\_SCORE\_TIMELINE\textgreater{}\par
\smallskip
Instructions:\par
- Frame numbers are 1-indexed: t=1 is the first frame, t=2 is the second frame, etc.\par
- All temporal queries are evaluated from t=1. If a query mentions previous frames, those frames are the prefix from t=1 up to, but not including, the chosen witness/anchor frame.\par
- Each score is the YOLO detector confidence in [0,1] for that object in that frame.\par
- The detector thresholds are calibration reference scores r\_c(0.5): for each object class c, this is the score level at which the empirical probability of c being present is 50\% on a held-out calibration split. Use these values as context for interpreting the detector score scale.\par
- Evaluate over this finite trace only.\par
- Do not write a program or describe an algorithm.\par
- Reason briefly if useful. The last line of your response must be exactly: FINAL\_JSON: \{"answer": true, "confidence": 0.82\} or the same format with false.\par
\smallskip
Output format: \{"answer": true/false, "confidence": number from 0 to 1\}\par
\end{llmprompt}

\begin{llmprompt}[label={prompt:scoring-ltl}]{Query matching with an LTL formula}
\noindent{\normalfont\bfseries System message}\par\smallskip
You are given a query that you need to match against a video. The query is provided as a LTL formula over objects in the video (car, person, bus, truck, bicycle, and motorcycle). You don't have access to the raw video, instead you are given frame-level detection scores for these objects from the YOLO object detector. Use this information to decide whether the video matches the temporal query. All temporal queries are evaluated from the first frame of the video; when a query mentions previous frames, interpret this as the prefix from frame 1 up to, but not including, the chosen witness/anchor frame. The provided detector thresholds are calibration reference scores r\_c(0.5): for each object class c, this is the detector score level at which the empirical probability of c being present is 50\% on a held-out calibration split. Use them as calibration context for interpreting the score scale. The query is already an LTL formula; do not reinterpret it as natural language. Evaluate it using the formula syntax and finite-trace semantics below. Allowed LTL syntax: eventually(p), always(p), not(p), and(p,q), or(p,q), until(p,q). Semantics over the finite trace: an atom p holds at frame t when object p is present at t; not(phi) negates phi at t; and(phi,psi) and or(phi,psi) combine truth values at t; eventually(phi) holds at t iff phi holds at some frame u \textgreater{}= t; always(phi) holds at t iff phi holds at every frame u \textgreater{}= t; until(phi,psi) holds at t iff there exists a frame u \textgreater{}= t where psi holds, and phi holds at every frame v with t \textless{}= v \textless{} u. The first argument of until is the condition before the witness frame; the second argument is the condition at the witness frame. You may reason briefly in text, but do not write code or markdown. End your response with a final line beginning with FINAL\_JSON: followed by \{"answer": true/false, "confidence": a number from 0 to 1\}.\par
\medskip
\noindent{\normalfont\bfseries User message template}\par\smallskip
Task: Decide whether the video matches the temporal query.\par
\smallskip
LTL formula:\par
\textless{}LTL\_FORMULA\textgreater{}\par
\smallskip
LTL syntax and semantics:\par
Allowed LTL syntax: eventually(p), always(p), not(p), and(p,q), or(p,q), until(p,q). Semantics over the finite trace: an atom p holds at frame t when object p is present at t; not(phi) negates phi at t; and(phi,psi) and or(phi,psi) combine truth values at t; eventually(phi) holds at t iff phi holds at some frame u \textgreater{}= t; always(phi) holds at t iff phi holds at every frame u \textgreater{}= t; until(phi,psi) holds at t iff there exists a frame u \textgreater{}= t where psi holds, and phi holds at every frame v with t \textless{}= v \textless{} u. The first argument of until is the condition before the witness frame; the second argument is the condition at the witness frame.\par
\smallskip
Detector calibration reference scores r\_c(0.5):\par
\textless{}CALIBRATION\_REFERENCE\_SCORES\textgreater{}\par
\smallskip
Detection scores:\par
\textless{}FRAME\_SCORE\_TIMELINE\textgreater{}\par
\smallskip
Instructions:\par
- Frame numbers are 1-indexed: t=1 is the first frame, t=2 is the second frame, etc.\par
- All temporal queries are evaluated from t=1. If a query mentions previous frames, those frames are the prefix from t=1 up to, but not including, the chosen witness/anchor frame.\par
- Each score is the YOLO detector confidence in [0,1] for that object in that frame.\par
- The detector thresholds are calibration reference scores r\_c(0.5): for each object class c, this is the score level at which the empirical probability of c being present is 50\% on a held-out calibration split. Use these values as context for interpreting the detector score scale.\par
- Evaluate over this finite trace only.\par
- Do not write a program or describe an algorithm.\par
- Reason briefly if useful. The last line of your response must be exactly: FINAL\_JSON: \{"answer": true, "confidence": 0.82\} or the same format with false.\par
\smallskip
Output format: \{"answer": true/false, "confidence": number from 0 to 1\}\par
\end{llmprompt}

%% file: graphics/charts/llm_parsing_prompts.tex
\begin{llmprompt}[label={prompt:parsing-15-examples}]{NL-to-LTL parsing with 15 template examples}
\noindent{\normalfont\bfseries System message}\par\smallskip
Translate natural-language video queries into Linear Temporal Logic (LTL) formulas over object predicates. Use only these atomic predicates: car, person, bus, truck, bicycle, motorcycle. Use this exact functional syntax: eventually(p), always(p), not(p), and(p,q), or(p,q), until(p,q). Do not include detector thresholds or score comparisons. Semantics over a finite video trace: eventually(p) means p holds in some frame; always(p) means p holds in every frame; not(p) means p is absent; and/or are Boolean; until(p,q) means q holds at some frame and p holds in all previous frames before that frame. The first argument of until is always the previous-frame condition; the second argument is the event or subsequent-frame condition. All formulas are evaluated from the first frame of the video; in these queries, previous frames means the prefix from the first frame up to, but not including, the chosen witness/anchor frame. When a previous-frame condition is translated with until(...), until already chooses the anchor/witness frame; do not wrap the result in eventually just because the query says 'starting at some frame'. For 'starting at some frame ... q in all subsequent frames and p in all previous frames', use until(p, always(q)). For 'starting at some frame ... q in some subsequent frame and p in all previous frames', use until(p, eventually(q)). For 'starting at any frame ... p in some subsequent frame', use always(eventually(p)).\par
\smallskip
Pattern examples:\par
Query 'Is a car present in any frame?' -\textgreater{} eventually(car).\par
Query 'Is a person present in all frames?' -\textgreater{} always(person).\par
Query 'Is a person present in any frame and a car present in all previous frames?' -\textgreater{} until(car, person).\par
Query 'Is a person present in any frame and a car absent in all previous frames?' -\textgreater{} until(not(car), person).\par
Query 'Is a person absent in any frame and a car present in all previous frames?' -\textgreater{} until(car, not(person)).\par
Query 'Are both a bus and a truck present in all frames?' -\textgreater{} and(always(bus), always(truck)).\par
Query 'Is a motorcycle present in any frame and both a car and a bus present in all previous frames?' -\textgreater{} until(and(car, bus), motorcycle).\par
Query 'Starting at some frame, is a motorcycle present in all subsequent frames and a car present in all previous frames?' -\textgreater{} until(car, always(motorcycle)).\par
Query 'Starting at some frame, is a motorcycle present in all subsequent frames and a car absent in all previous frames?' -\textgreater{} until(not(car), always(motorcycle)).\par
Query 'Starting at some frame, is a motorcycle present in some subsequent frame and a car absent in all previous frames?' -\textgreater{} until(not(car), eventually(motorcycle)).\par
Query 'Starting at some frame, is a bicycle present in some subsequent frame and both a car and a person present in all previous frames?' -\textgreater{} until(and(car, person), eventually(bicycle)).\par
Query 'Starting at some frame, is a person present in all subsequent frames?' -\textgreater{} eventually(always(person)).\par
Query 'Starting at any frame, is a person present in some subsequent frame?' -\textgreater{} always(eventually(person)).\par
Query 'Is a car present in all frames or is a truck present in any frame?' -\textgreater{} or(always(car), eventually(truck)).\par
Query 'Is a car present in all frames and is a truck present in any frame?' -\textgreater{} and(always(car), eventually(truck)).\par
\smallskip
Always close all parentheses. Return only JSON: \{"formula": "...", "confidence": 0.82\}. Do not write reasoning, markdown, or explanations.\par
\medskip
\noindent{\normalfont\bfseries User message template}\par\smallskip
Query:\par
\textless{}NL\_QUERY\textgreater{}\par
\smallskip
Allowed atoms:\par
car, person, bus, truck, bicycle, motorcycle\par
\smallskip
Allowed operators:\par
eventually, always, not, and, or, until\par
\smallskip
Output syntax:\par
eventually(atom), always(atom), not(atom), and(expr,expr), or(expr,expr), until(expr,expr)\par
\smallskip
Remember: in until(previous\_condition, event\_condition), the previous-frame condition is first.\par
\smallskip
Your entire response must be a JSON object like \{"formula": "eventually(car)", "confidence": 0.82\}.\par
\end{llmprompt}

\begin{llmprompt}[label={prompt:parsing-3-examples}]{NL-to-LTL parsing with 3 selected examples}
\noindent{\normalfont\bfseries System message}\par\smallskip
Task: Translate natural-language video queries into Linear Temporal Logic (LTL) formulas over object predicates.\par
\smallskip
Allowed atomic predicates:\par
car, person, bus, truck, bicycle, motorcycle\par
\smallskip
Required output syntax:\par
eventually(p), always(p), not(p), and(p,q), or(p,q), until(p,q)\par
\smallskip
Do not include detector thresholds or score comparisons.\par
\smallskip
Finite-trace semantics evaluated at a current frame t:\par
- An atom p holds at t when object p is present at t.\par
- not(phi) negates phi at t.\par
- and(phi,psi) and or(phi,psi) combine truth values at t.\par
- eventually(phi) holds at t iff phi holds at some frame u \textgreater{}= t.\par
- always(phi) holds at t iff phi holds at every frame u \textgreater{}= t.\par
- until(phi,psi) holds at t iff there exists a frame u \textgreater{}= t where psi holds, and phi holds at every frame v with t \textless{}= v \textless{} u.\par
\smallskip
Anchor-frame convention:\par
- All formulas are evaluated at the first frame of the video. In these queries, "previous frames" means the prefix from the first frame up to, but not including, the chosen witness/anchor frame. Encode this prefix condition with the first argument of until(...).\par
\smallskip
Important until convention:\par
- The first argument of until is the condition before the witness frame.\par
- The second argument of until is the condition at the witness frame.\par
- Either argument may itself be a nested temporal or Boolean formula.\par
- In this syntax, eventually and always are suffix/future operators, not past-time operators.\par
\smallskip
Nested formula guidance:\par
- Queries may be nested.\par
- The target formula may use simple temporal forms: eventually(p), always(p), until(p,q).\par
- The target formula may use Boolean combinations of temporal formulas: and(always(p), eventually(q)), or(always(p), eventually(q)).\par
- The target formula may use temporal formulas over Boolean or negated predicates: until(not(p), q), until(p, not(q)), until(and(p,q), r), and(always(p), always(q)).\par
- The target formula may use nested temporal formulas: until(p, always(q)), eventually(always(p)), always(eventually(p)).\par
- The target formula may combine Boolean and temporal children: until(not(p), always(q)), until(not(p), eventually(q)), until(and(p,q), eventually(r)).\par
- Choose the structure that matches the query, then bind p, q, and r to the mentioned objects.\par
\smallskip
Scope rules:\par
- If a query has a previous-frame condition tied to an event condition, translate it as one until(previous\_condition, event\_condition) formula.\par
- When a previous-frame condition is translated with until(...), until already chooses the anchor/witness frame; do not wrap the result in eventually just because the query says "starting at some frame".\par
- Do not translate tied previous/event conditions as independent conjunctions such as and(eventually(event), always(previous\_condition)).\par
- Never use always(phi) to mean that phi held in previous frames; always(phi) only talks about the suffix from the current frame onward.\par
- Inside until(previous\_condition, event\_condition), do not wrap the previous\_condition in always just because the text says it holds in all previous frames; until already supplies the previous-frame quantification over the prefix before the witness.\par
- The previous\_condition should be an atom, not(atom), and(atom,atom), or another local Boolean condition unless the query explicitly places a temporal requirement inside the previous condition.\par
- When an event is paired with a previous-frame condition, the event is the witness condition of until. If the event only needs to hold at the witness frame, use the bare event atom as event\_condition, not eventually(event).\par
- If the event condition requires existence somewhere in the suffix beginning at the witness frame, use eventually(event) as event\_condition.\par
- If the event condition requires holding throughout the suffix beginning at the witness frame, use always(event) as event\_condition.\par
- If there is an existential anchor frame but no previous-frame condition, use eventually(phi), where phi is the suffix property at that anchor.\par
- If the suffix property is that p holds throughout the suffix, this becomes eventually(always(p)).\par
- If the requirement is checked from every anchor frame, use always(phi), where phi is the suffix property at each anchor.\par
- If the suffix property is that p occurs somewhere in the suffix, this becomes always(eventually(p)).\par
- Do not use constants such as true or false; every formula must be built only from the allowed atoms and operators.\par
\smallskip
Translation procedure:\par
- First identify whether the query contains a previous-frame condition.\par
- If it does, identify the associated witness/event condition and use until(previous\_condition, event\_condition).\par
- For previous-frame conditions, remove the temporal words that describe the prefix; keep the local predicate or local Boolean condition inside the first until argument.\par
- Only use a top-level and/or when the query combines two independent temporal requirements, not when one clause describes previous frames for the other.\par
- If there is no previous-frame condition, choose the outer temporal operator directly from the query scope.\par
- Always close all parentheses.\par
\smallskip
Targeted complex examples:\par
- Query 'Is a person present in any frame and a car present in all previous frames?' -\textgreater{} until(car, person).\par
- Query 'Starting at some frame, is a bicycle present in some subsequent frame and both a car and a person present in all previous frames?' -\textgreater{} until(and(car, person), eventually(bicycle)).\par
- Query 'Starting at some frame, is a person present in all subsequent frames?' -\textgreater{} eventually(always(person)).\par
\smallskip
Return only JSON: \{"formula": "...", "confidence": 0.82\}. Do not write reasoning, markdown, or explanations.\par
\medskip
\noindent{\normalfont\bfseries User message template}\par\smallskip
Query:\par
\textless{}NL\_QUERY\textgreater{}\par
\smallskip
Allowed atoms:\par
car, person, bus, truck, bicycle, motorcycle\par
\smallskip
Allowed operators:\par
eventually, always, not, and, or, until\par
\smallskip
Output syntax:\par
eventually(atom), always(atom), not(atom), and(expr,expr), or(expr,expr), until(expr,expr)\par
\smallskip
Remember: in until(previous\_condition, event\_condition), the previous-frame condition is first.\par
\smallskip
Your entire response must be a JSON object like \{"formula": "eventually(car)", "confidence": 0.82\}.\par
\end{llmprompt}

\begin{llmprompt}[label={prompt:parsing-3-reasoned-examples}]{NL-to-LTL parsing with 3 reasoned examples}
\noindent{\normalfont\bfseries System message}\par\smallskip
Task: Translate natural-language video queries into Linear Temporal Logic (LTL) formulas over object predicates.\par
\smallskip
Allowed atomic predicates:\par
car, person, bus, truck, bicycle, motorcycle\par
\smallskip
Required output syntax:\par
eventually(p), always(p), not(p), and(p,q), or(p,q), until(p,q)\par
\smallskip
Do not include detector thresholds or score comparisons.\par
\smallskip
Finite-trace semantics evaluated at a current frame t:\par
- An atom p holds at t when object p is present at t.\par
- not(phi) negates phi at t.\par
- and(phi,psi) and or(phi,psi) combine truth values at t.\par
- eventually(phi) holds at t iff phi holds at some frame u \textgreater{}= t.\par
- always(phi) holds at t iff phi holds at every frame u \textgreater{}= t.\par
- until(phi,psi) holds at t iff there exists a frame u \textgreater{}= t where psi holds, and phi holds at every frame v with t \textless{}= v \textless{} u.\par
\smallskip
Anchor-frame convention:\par
- All formulas are evaluated at the first frame of the video. In these queries, "previous frames" means the prefix from the first frame up to, but not including, the chosen witness/anchor frame. Encode this prefix condition with the first argument of until(...).\par
\smallskip
Important until convention:\par
- The first argument of until is the condition before the witness frame.\par
- The second argument of until is the condition at the witness frame.\par
- Either argument may itself be a nested temporal or Boolean formula.\par
- In this syntax, eventually and always are suffix/future operators, not past-time operators.\par
\smallskip
Nested formula guidance:\par
- Queries may be nested.\par
- The target formula may use simple temporal forms: eventually(p), always(p), until(p,q).\par
- The target formula may use Boolean combinations of temporal formulas: and(always(p), eventually(q)), or(always(p), eventually(q)).\par
- The target formula may use temporal formulas over Boolean or negated predicates: until(not(p), q), until(p, not(q)), until(and(p,q), r), and(always(p), always(q)).\par
- The target formula may use nested temporal formulas: until(p, always(q)), eventually(always(p)), always(eventually(p)).\par
- The target formula may combine Boolean and temporal children: until(not(p), always(q)), until(not(p), eventually(q)), until(and(p,q), eventually(r)).\par
- Choose the structure that matches the query, then bind p, q, and r to the mentioned objects.\par
\smallskip
Scope rules:\par
- If a query has a previous-frame condition tied to an event condition, translate it as one until(previous\_condition, event\_condition) formula.\par
- When a previous-frame condition is translated with until(...), until already chooses the anchor/witness frame; do not wrap the result in eventually just because the query says "starting at some frame".\par
- Do not translate tied previous/event conditions as independent conjunctions such as and(eventually(event), always(previous\_condition)).\par
- Never use always(phi) to mean that phi held in previous frames; always(phi) only talks about the suffix from the current frame onward.\par
- Inside until(previous\_condition, event\_condition), do not wrap the previous\_condition in always just because the text says it holds in all previous frames; until already supplies the previous-frame quantification over the prefix before the witness.\par
- The previous\_condition should be an atom, not(atom), and(atom,atom), or another local Boolean condition unless the query explicitly places a temporal requirement inside the previous condition.\par
- When an event is paired with a previous-frame condition, the event is the witness condition of until. If the event only needs to hold at the witness frame, use the bare event atom as event\_condition, not eventually(event).\par
- If the event condition requires existence somewhere in the suffix beginning at the witness frame, use eventually(event) as event\_condition.\par
- If the event condition requires holding throughout the suffix beginning at the witness frame, use always(event) as event\_condition.\par
- If there is an existential anchor frame but no previous-frame condition, use eventually(phi), where phi is the suffix property at that anchor.\par
- If the suffix property is that p holds throughout the suffix, this becomes eventually(always(p)).\par
- If the requirement is checked from every anchor frame, use always(phi), where phi is the suffix property at each anchor.\par
- If the suffix property is that p occurs somewhere in the suffix, this becomes always(eventually(p)).\par
- Do not use constants such as true or false; every formula must be built only from the allowed atoms and operators.\par
\smallskip
Translation procedure:\par
- First identify whether the query contains a previous-frame condition.\par
- If it does, identify the associated witness/event condition and use until(previous\_condition, event\_condition).\par
- For previous-frame conditions, remove the temporal words that describe the prefix; keep the local predicate or local Boolean condition inside the first until argument.\par
- Only use a top-level and/or when the query combines two independent temporal requirements, not when one clause describes previous frames for the other.\par
- If there is no previous-frame condition, choose the outer temporal operator directly from the query scope.\par
- Always close all parentheses.\par
\smallskip
Targeted complex examples with brief rationale:\par
- Query 'Is a person present in any frame and a car present in all previous frames?' Rationale: person is the witness event; car is the prefix condition before that witness. Formula: until(car, person).\par
- Query 'Starting at some frame, is a bicycle present in some subsequent frame and both a car and a person present in all previous frames?' Rationale: the prefix condition is the local conjunction and(car, person); the witness condition requires bicycle somewhere in the suffix, so use eventually(bicycle). Formula: until(and(car, person), eventually(bicycle)).\par
- Query 'Starting at some frame, is a person present in all subsequent frames?' Rationale: there is no previous-frame condition, so choose an anchor with eventually; from that anchor onward person must always hold. Formula: eventually(always(person)).\par
\smallskip
Return only JSON: \{"formula": "...", "confidence": 0.82\}. Do not write reasoning, markdown, or explanations.\par
\medskip
\noindent{\normalfont\bfseries User message template}\par\smallskip
Query:\par
\textless{}NL\_QUERY\textgreater{}\par
\smallskip
Allowed atoms:\par
car, person, bus, truck, bicycle, motorcycle\par
\smallskip
Allowed operators:\par
eventually, always, not, and, or, until\par
\smallskip
Output syntax:\par
eventually(atom), always(atom), not(atom), and(expr,expr), or(expr,expr), until(expr,expr)\par
\smallskip
Remember: in until(previous\_condition, event\_condition), the previous-frame condition is first.\par
\smallskip
Your entire response must be a JSON object like \{"formula": "eventually(car)", "confidence": 0.82\}.\par
\end{llmprompt}

%% file: sources.bib
@article{gmrob,
  author={Mehdipour, Noushin and Vasile, Cristian-Ioan and Belta, Calin},
  journal={IEEE Transactions on Automatic Control}, 
  title={Generalized Mean Robustness for Signal Temporal Logic}, 
  year={2024},
  volume={},
  number={},
  pages={1-8},
  doi={10.1109/TAC.2024.3482104}}

@article{FAINEKOS20094262,
title = {Robustness of temporal logic specifications for continuous-time signals},
journal = {Theoretical Computer Science},
volume = {410},
number = {42},
pages = {4262-4291},
year = {2009},
issn = {0304-3975},
doi = {https://doi.org/10.1016/j.tcs.2009.06.021},
url = {https://www.sciencedirect.com/science/article/pii/S0304397509004149},
author = {Georgios E. Fainekos and George J. Pappas}
}

@inproceedings{Akazaki2015TimeRI,
  title={Time Robustness in MTL and Expressivity in Hybrid System Falsification},
  author={Takumi Akazaki and Ichiro Hasuo},
  booktitle={International Conference on Computer Aided Verification},
  year={2015},
  url={https://api.semanticscholar.org/CorpusID:14307127}
}

@inproceedings{Choi2024TowardsNV,
  title={Towards Neuro-Symbolic Video Understanding},
  author={Minkyu Choi and Harsh Goel and Mohammad Omama and Yunhao Yang and Sahil Shah and Sandeep Chinchali},
  booktitle={European Conference on Computer Vision},
  year={2024},
  url={https://api.semanticscholar.org/CorpusID:268513042}
}

@article{videomme,
  title={Video-mme: The first-ever comprehensive evaluation benchmark of multi-modal llms in video analysis},
  author={Fu, Chaoyou and Dai, Yuhan and Luo, Yongdong and Li, Lei and Ren, Shuhuai and Zhang, Renrui and Wang, Zihan and Zhou, Chenyu and Shen, Yunhang and Zhang, Mengdan and others},
  journal={arXiv preprint arXiv:2405.21075},
  year={2024}
}

@article{videollava,
  title={Video-llava: Learning united visual representation by alignment before projection},
  author={Lin, Bin and Ye, Yang and Zhu, Bin and Cui, Jiaxi and Ning, Munan and Jin, Peng and Yuan, Li},
  journal={arXiv preprint arXiv:2311.10122},
  year={2023}
}

@article{xue2024longvila,
  title={Longvila: Scaling long-context visual language models for long videos},
  author={Xue, Fuzhao and Chen, Yukang and Li, Dacheng and Hu, Qinghao and Zhu, Ligeng and Li, Xiuyu and Fang, Yunhao and Tang, Haotian and Yang, Shang and Liu, Zhijian and others},
  journal={arXiv preprint arXiv:2408.10188},
  year={2024}
}

@inproceedings{yolo,
  title={You only look once: Unified, real-time object detection},
  author={Redmon, Joseph and Divvala, Santosh and Girshick, Ross and Farhadi, Ali},
  booktitle={Proceedings of the IEEE conference on computer vision and pattern recognition},
  pages={779--788},
  year={2016}
}

@article{storm,
  title={The probabilistic model checker Storm},
  author={Hensel, Christian and Junges, Sebastian and Katoen, Joost-Pieter and Quatmann, Tim and Volk, Matthias},
  journal={International Journal on Software Tools for Technology Transfer},
  pages={1--22},
  year={2022},
  publisher={Springer}
}

@InProceedings{waymo, author = {Sun, Pei and Kretzschmar, Henrik and Dotiwalla, Xerxes and Chouard, Aurelien and Patnaik, Vijaysai and Tsui, Paul and Guo, James and Zhou, Yin and Chai, Yuning and Caine, Benjamin and Vasudevan, Vijay and Han, Wei and Ngiam, Jiquan and Zhao, Hang and Timofeev, Aleksei and Ettinger, Scott and Krivokon, Maxim and Gao, Amy and Joshi, Aditya and Zhang, Yu and Shlens, Jonathon and Chen, Zhifeng and Anguelov, Dragomir}, title = {Scalability in Perception for Autonomous Driving: Waymo Open Dataset}, booktitle = {Proceedings of the IEEE/CVF Conference on Computer Vision and Pattern Recognition (CVPR)}, month = {June}, year = {2020} }

@InProceedings{nuscenes,
author = {Caesar, Holger and Bankiti, Varun and Lang, Alex H. and Vora, Sourabh and Liong, Venice Erin and Xu, Qiang and Krishnan, Anush and Pan, Yu and Baldan, Giancarlo and Beijbom, Oscar},
title = {nuScenes: A Multimodal Dataset for Autonomous Driving},
booktitle = {Proceedings of the IEEE/CVF Conference on Computer Vision and Pattern Recognition (CVPR)},
month = {June},
year = {2020}
}

@article{busso2008iemocap,
  title={IEMOCAP: Interactive emotional dyadic motion capture database},
  author={Busso, Carlos and Bulut, Murtaza and Lee, Chi-Chun and Kazemzadeh, Abe and Mower, Emily and Kim, Samuel and Chang, Jeannette N and Lee, Sungbok and Narayanan, Shrikanth S},
  journal={Language resources and evaluation},
  volume={42},
  pages={335--359},
  year={2008},
  publisher={Springer}
}

@article{zhang2024longva,
  title={Long Context Transfer from Language to Vision},
  author={Peiyuan Zhang and Kaichen Zhang and Bo Li and Guangtao Zeng and Jingkang Yang and Yuanhan Zhang and Ziyue Wang and Haoran Tan and Chunyuan Li and Ziwei Liu},
  journal={arXiv preprint arXiv:2406.16852},
  year={2024},
  url = {https://arxiv.org/abs/2406.16852}
}

@article{Qwen-Audio,
  title={Qwen-Audio: Advancing Universal Audio Understanding via Unified Large-Scale Audio-Language Models},
  author={Chu, Yunfei and Xu, Jin and Zhou, Xiaohuan and Yang, Qian and Zhang, Shiliang and Yan, Zhijie  and Zhou, Chang and Zhou, Jingren},
  journal={arXiv preprint arXiv:2311.07919},
  year={2023}
}

@article{Qwen2-Audio,
  title={Qwen2-Audio Technical Report},
  author={Chu, Yunfei and Xu, Jin and Yang, Qian and Wei, Haojie and Wei, Xipin and Guo,  Zhifang and Leng, Yichong and Lv, Yuanjun and He, Jinzheng and Lin, Junyang and Zhou, Chang and Zhou, Jingren},
  journal={arXiv preprint arXiv:2407.10759},
  year={2024}
}

@software{yolov8_ultralytics,
  author = {Glenn Jocher and Ayush Chaurasia and Jing Qiu},
  title = {Ultralytics YOLOv8},
  version = {8.0.0},
  year = {2023},
  url = {https://github.com/ultralytics/ultralytics},
  orcid = {0000-0001-5950-6979, 0000-0002-7603-6750, 0000-0003-3783-7069},
  license = {AGPL-3.0}
}

@inproceedings{superb,
  author={Shu-wen Yang and Po-Han Chi and Yung-Sung Chuang and Cheng-I Jeff Lai and Kushal Lakhotia and Yist Y. Lin and Andy T. Liu and Jiatong Shi and Xuankai Chang and Guan-Ting Lin and Tzu-Hsien Huang and Wei-Cheng Tseng and Ko-tik Lee and Da-Rong Liu and Zili Huang and Shuyan Dong and Shang-Wen Li and Shinji Watanabe and Abdelrahman Mohamed and Hung-yi Lee},
  title={{SUPERB: Speech Processing Universal PERformance Benchmark}},
  year=2021,
  booktitle={Proc. Interspeech 2021},
  pages={1194--1198},
  doi={10.21437/Interspeech.2021-1775}
}

@article{hsu2021hubert,
  title={Hubert: Self-supervised speech representation learning by masked prediction of hidden units},
  author={Hsu, Wei-Ning and Bolte, Benjamin and Tsai, Yao-Hung Hubert and Lakhotia, Kushal and Salakhutdinov, Ruslan and Mohamed, Abdelrahman},
  journal={IEEE/ACM transactions on audio, speech, and language processing},
  volume={29},
  pages={3451--3460},
  year={2021},
  publisher={IEEE}
}

@inproceedings{ltl,
  title={The temporal logic of programs},
  author={Pnueli, Amir},
  booktitle={18th annual symposium on foundations of computer science (sfcs 1977)},
  pages={46--57},
  year={1977},
  organization={ieee}
}

@article{cai2024temporalbench,
  title={Temporalbench: Benchmarking fine-grained temporal understanding for multimodal video models},
  author={Cai, Mu and Tan, Reuben and Zhang, Jianrui and Zou, Bocheng and Zhang, Kai and Yao, Feng and Zhu, Fangrui and Gu, Jing and Zhong, Yiwu and Shang, Yuzhang and others},
  journal={arXiv preprint arXiv:2410.10818},
  year={2024}
}

@article{liu2024tempcompass,
  title={Tempcompass: Do video llms really understand videos?},
  author={Liu, Yuanxin and Li, Shicheng and Liu, Yi and Wang, Yuxiang and Ren, Shuhuai and Li, Lei and Chen, Sishuo and Sun, Xu and Hou, Lu},
  journal={arXiv preprint arXiv:2403.00476},
  year={2024}
}

@article{chen2024rextime,
  title={Rextime: A benchmark suite for reasoning-across-time in videos},
  author={Chen, Jr-Jen and Liao, Yu-Chien and Lin, Hsi-Che and Yu, Yu-Chu and Chen, Yen-Chun and Wang, Frank},
  journal={Advances in Neural Information Processing Systems},
  volume={37},
  pages={28662--28673},
  year={2024}
}

@inproceedings{nextqa,
  title={Next-qa: Next phase of question-answering to explaining temporal actions},
  author={Xiao, Junbin and Shang, Xindi and Yao, Angela and Chua, Tat-Seng},
  booktitle={Proceedings of the IEEE/CVF conference on computer vision and pattern recognition},
  pages={9777--9786},
  year={2021}
}

@article{qvhighlights,
  title={Detecting moments and highlights in videos via natural language queries},
  author={Lei, Jie and Berg, Tamara L and Bansal, Mohit},
  journal={Advances in Neural Information Processing Systems},
  volume={34},
  pages={11846--11858},
  year={2021}
}

@article{sakshi2024mmau,
  title={Mmau: A massive multi-task audio understanding and reasoning benchmark},
  author={Sakshi, S and Tyagi, Utkarsh and Kumar, Sonal and Seth, Ashish and Selvakumar, Ramaneswaran and Nieto, Oriol and Duraiswami, Ramani and Ghosh, Sreyan and Manocha, Dinesh},
  journal={arXiv preprint arXiv:2410.19168},
  year={2024}
}

@article{ghosh2023compa,
  title={Compa: Addressing the gap in compositional reasoning in audio-language models},
  author={Ghosh, Sreyan and Seth, Ashish and Kumar, Sonal and Tyagi, Utkarsh and Evuru, Chandra Kiran and Ramaneswaran, S and Sakshi, S and Nieto, Oriol and Duraiswami, Ramani and Manocha, Dinesh},
  journal={arXiv preprint arXiv:2310.08753},
  year={2023}
}

@inproceedings{strem,
  title={Pattern matching for perception streams},
  author={Anderson, Jacob and Fainekos, Georgios and Hoxha, Bardh and Okamoto, Hideki and Prokhorov, Danil},
  booktitle={International Conference on Runtime Verification},
  pages={251--270},
  year={2023},
  organization={Springer}
}

@inproceedings{donze2010robust,
  title={Robust satisfaction of temporal logic over real-valued signals},
  author={Donz{\'e}, Alexandre and Maler, Oded},
  booktitle={International Conference on Formal Modeling and Analysis of Timed Systems},
  pages={92--106},
  year={2010},
  organization={Springer}
}

@article{yang2023specification,
  title={Specification-driven video search via foundation models and formal verification},
  author={Yang, Yunhao and Gaglione, Jean-Rapha{\"e}l and Chinchali, Sandeep and Topcu, Ufuk},
  journal={arXiv preprint arXiv:2309.10171},
  year={2023}
}

@inproceedings{liu2024grounding,
  title={Grounding dino: Marrying dino with grounded pre-training for open-set object detection},
  author={Liu, Shilong and Zeng, Zhaoyang and Ren, Tianhe and Li, Feng and Zhang, Hao and Yang, Jie and Jiang, Qing and Li, Chunyuan and Yang, Jianwei and Su, Hang and others},
  booktitle={European conference on computer vision},
  pages={38--55},
  year={2024},
  organization={Springer}
}

@article{minderer2023scaling,
  title={Scaling open-vocabulary object detection},
  author={Minderer, Matthias and Gritsenko, Alexey and Houlsby, Neil},
  journal={Advances in Neural Information Processing Systems},
  volume={36},
  pages={72983--73007},
  year={2023}
}

@inproceedings{gu2018ava,
  title={Ava: A video dataset of spatio-temporally localized atomic visual actions},
  author={Gu, Chunhui and Sun, Chen and Ross, David A and Vondrick, Carl and Pantofaru, Caroline and Li, Yeqing and Vijayanarasimhan, Sudheendra and Toderici, George and Ricco, Susanna and Sukthankar, Rahul and others},
  booktitle={Proceedings of the IEEE conference on computer vision and pattern recognition},
  pages={6047--6056},
  year={2018}
}

@inproceedings{feichtenhofer2019slowfast,
  title={Slowfast networks for video recognition},
  author={Feichtenhofer, Christoph and Fan, Haoqi and Malik, Jitendra and He, Kaiming},
  booktitle={Proceedings of the IEEE/CVF international conference on computer vision},
  pages={6202--6211},
  year={2019}
}

@article{li2022mplug,
  title={mplug: Effective and efficient vision-language learning by cross-modal skip-connections},
  author={Li, Chenliang and Xu, Haiyang and Tian, Junfeng and Wang, Wei and Yan, Ming and Bi, Bin and Ye, Jiabo and Chen, Hehong and Xu, Guohai and Cao, Zheng and others},
  journal={arXiv preprint arXiv:2205.12005},
  year={2022}
}

@inproceedings{didemo,
  title={Localizing moments in video with natural language},
  author={Anne Hendricks, Lisa and Wang, Oliver and Shechtman, Eli and Sivic, Josef and Darrell, Trevor and Russell, Bryan},
  booktitle={Proceedings of the IEEE international conference on computer vision},
  pages={5803--5812},
  year={2017}
}

@inproceedings{liu-etal-2025-eliot,
    title = "{ELIOT}: Zero-Shot Video-Text Retrieval through Relevance-Boosted Captioning and Structural Information Extraction",
    author = "Liu, Xuye  and
      Wang, Yimu  and
      Zhao, Jian",
    editor = "Ebrahimi, Abteen  and
      Haider, Samar  and
      Liu, Emmy  and
      Haider, Sammar  and
      Leonor Pacheco, Maria  and
      Wein, Shira",
    booktitle = "Proceedings of the 2025 Conference of the Nations of the Americas Chapter of the Association for Computational Linguistics: Human Language Technologies (Volume 4: Student Research Workshop)",
    month = apr,
    year = "2025",
    address = "Albuquerque, USA",
    publisher = "Association for Computational Linguistics",
    url = "https://aclanthology.org/2025.naacl-srw.37/",
    doi = "10.18653/v1/2025.naacl-srw.37",
    pages = "381--391",
    ISBN = "979-8-89176-192-6"
}

@misc{zhang2024llavanextvideo,
  title={LLaVA-NeXT: A Strong Zero-shot Video Understanding Model},
  url={https://llava-vl.github.io/blog/2024-04-30-llava-next-video/},
  author={Zhang, Yuanhan and Li, Bo and Liu, haotian and Lee, Yong jae and Gui, Liangke and Fu, Di and Feng, Jiashi and Liu, Ziwei and Li, Chunyuan},
  month={April},
  year={2024}
}

@inproceedings{reimers-2020-multilingual-sentence-bert,
    title = "Making Monolingual Sentence Embeddings Multilingual using Knowledge Distillation",
    author = "Reimers, Nils and Gurevych, Iryna",
    booktitle = "Proceedings of the 2020 Conference on Empirical Methods in Natural Language Processing",
    month = "11",
    year = "2020",
    publisher = "Association for Computational Linguistics",
    url = "https://arxiv.org/abs/2004.09813",
}

@misc{wu2019detectron2,
  author =       {Yuxin Wu and Alexander Kirillov and Francisco Massa and
                  Wan-Yen Lo and Ross Girshick},
  title =        {Detectron2},
  howpublished = {\url{https://github.com/facebookresearch/detectron2}},
  year =         {2019}
}

@article{huang2023laser,
  title={Laser: A neuro-symbolic framework for learning spatial-temporal scene graphs with weak supervision},
  author={Huang, Jiani and Li, Ziyang and Naik, Mayur and Lim, Ser-Nam},
  journal={arXiv preprint arXiv:2304.07647},
  year={2023}
}

@inproceedings{krishna2017dense,
    title={Dense-Captioning Events in Videos},
    author={Krishna, Ranjay and Hata, Kenji and Ren, Frederic and Fei-Fei, Li and Niebles, Juan Carlos},
    booktitle={International Conference on Computer Vision (ICCV)},
    year={2017}
}

@article{luo2021clip4clip,
  title={Clip4clip: An empirical study of clip for end to end video clip retrieval},
  author={Luo, Huaishao and Ji, Lei and Zhong, Ming and Chen, Yang and Lei, Wen and Duan, Nan and Li, Tianrui},
  journal={arXiv preprint arXiv:2104.08860},
  year={2021}
}

@InProceedings{Bain21,
  author       = "Max Bain and Arsha Nagrani and G{\"u}l Varol and Andrew Zisserman",
  title        = "Frozen in Time: A Joint Video and Image Encoder for End-to-End Retrieval",
  booktitle    = "IEEE International Conference on Computer Vision",
  year         = "2021",
}

@inproceedings{liu2022ts2,
  title={Ts2-net: Token shift and selection transformer for text-video retrieval},
  author={Liu, Yuqi and Xiong, Pengfei and Xu, Luhui and Cao, Shengming and Jin, Qin},
  booktitle={European conference on computer vision},
  pages={319--335},
  year={2022},
  organization={Springer}
}

@article{tewel2022zero,
  title={Zero-shot video captioning with evolving pseudo-tokens},
  author={Tewel, Yoad and Shalev, Yoav and Nadler, Roy and Schwartz, Idan and Wolf, Lior},
  journal={arXiv preprint arXiv:2207.11100},
  year={2022}
}

@inproceedings{ltlf, author = {De Giacomo, Giuseppe and Vardi, Moshe Y.}, title = {Linear temporal logic and linear dynamic logic on finite traces}, year = {2013}, isbn = {9781577356332}, publisher = {AAAI Press}, booktitle = {Proceedings of the Twenty-Third International Joint Conference on Artificial Intelligence}, pages = {854–860}, numpages = {7}, location = {Beijing, China}, series = {IJCAI '13} }

@inproceedings{nesya,
  title     = {NeSyA: Neurosymbolic Automata},
  author    = {Manginas, Nikolaos  and Paliouras, George and De Raedt, Luc},
  booktitle = {Proceedings of the Thirty-Fourth International Joint Conference on
               Artificial Intelligence, {IJCAI-25}},
  publisher = {International Joint Conferences on Artificial Intelligence Organization},
  editor    = {James Kwok},
  pages     = {5950--5958},
  year      = {2025},
  month     = {8},
  note      = {Main Track},
  doi       = {10.24963/ijcai.2025/662},
  url       = {https://doi.org/10.24963/ijcai.2025/662},
}

@inproceedings{chen2024internvl,
    title={Internvl: Scaling up vision foundation models and aligning for generic visual-linguistic tasks},
    author={Chen, Zhe and Wu, Jiannan and Wang, Wenhai and Su, Weijie and Chen, Guo and Xing, Sen and Zhong, Muyan and Zhang, Qinglong and Zhu, Xizhou and Lu, Lewei and others},
    booktitle={Proceedings of the IEEE/CVF Conference on Computer Vision and Pattern Recognition},
    pages={24185--24198},
    year={2024}
  }
